\PassOptionsToPackage{table}{xcolor}
\documentclass[11pt]{article}

\usepackage[final]{acl}

\usepackage{times}
\usepackage{latexsym}
\usepackage{inconsolata}

\usepackage[utf8]{inputenc} 
\usepackage[T1]{fontenc}    
\usepackage{hyperref}       
\usepackage{url}            
\usepackage{booktabs}       
\usepackage{amsfonts}       
\usepackage{nicefrac}       
\usepackage{microtype}      
\usepackage{longtable}
\usepackage{array}
\usepackage{arydshln}
\usepackage{graphicx}
\usepackage{geometry}
\usepackage{caption}
\usepackage{amsmath}
\usepackage{multirow}
\usepackage{ulem} 
\usepackage{listings}
\usepackage[most]{tcolorbox} 
\tcbuselibrary{skins, breakable} 
\usepackage[dvipsnames]{xcolor}   
\usepackage{enumitem}             
\usepackage{dashrule}             
\usepackage{makecell}
\usepackage{pifont}
\usepackage[table]{xcolor}

\definecolor{boxFrameGreen}{HTML}{5E9969} 
\definecolor{boxTitleBack}{HTML}{89B890} 
\definecolor{listNumberBlue}{HTML}{2E74B5}  

\definecolor{promptframecolor}{RGB}{102,102,178}
\definecolor{promptbackcolor}{RGB}{244,251,254}

\newcommand{\tautwobench}{$\tau^2$-bench}
\newtcolorbox{promptbox}[2][]{
  colback=promptbackcolor, 
  colframe=promptframecolor, 
  fonttitle=\bfseries\normalsize, 
  fontupper=\scriptsize,
  coltitle=white, 
  title=#2, 
  #1
}

\newtcolorbox{takeaway}{
  colback=white,        
  colframe=black,       
  boxrule=0.8pt,        
  sharp corners,        
  left=5pt, right=5pt,  
  top=4pt, bottom=4pt,  
  parbox=false          
}

\NewDocumentCommand{\yafu}
{ mO{} }{\textcolor{blue}{\textsuperscript{\textit{yafu}}\textsf{\textbf{\small[#1]}}}}

\title{Agent Planning Benchmark: A Diagnostic Framework for Planning Capabilities in LLM Agents}

\author{
  \textbf{Haoyu Sun\textsuperscript{1,2,*}}\quad
  \textbf{Wenxuan Wang\textsuperscript{3,2,*}}\quad
  \textbf{Mingyang Song\textsuperscript{4}}\quad
  \textbf{Jujie He\textsuperscript{5}}
  \\
  \textbf{Weinan Zhang\textsuperscript{3}}\quad
  \textbf{Yang Liu\textsuperscript{6}}\quad
  \textbf{Yang Yang\textsuperscript{7,\textdagger}}\quad
  \textbf{Yu Cheng\textsuperscript{8,2,\textdagger}}
  \\[0.6em]
  \normalfont\textsuperscript{1}Shanghai Research Institute for Intelligent Autonomous Systems,\\
  Tongji University, Shanghai, China \\[0.2em]\quad
  \textsuperscript{2}Shanghai AI Laboratory\quad
  \textsuperscript{3}Harbin Institute of Technology
  \\
  \normalfont\textsuperscript{4}Fudan University\quad
  \textsuperscript{5}Skywork AI\quad
  \textsuperscript{6}University of California, Santa Cruz
  \\
  \normalfont\textsuperscript{7}Shanghai Jiao Tong University\quad
  \textsuperscript{8}Nanyang Technological University
  \\[0.6em]
  \normalfont\small
  \textsuperscript{*}Equal contribution.\quad
  \textsuperscript{\textdagger}Corresponding authors.
  \\
  \normalfont\small
}

\begin{document}
\maketitle

\begin{abstract}
Planning is central to LLM agents: before acting, an agent must decompose goals, select tools, reason over constraints, and decide when a task is infeasible. Yet existing agent evaluations often report only end-to-end success, making it difficult to determine whether failures stem from planning or execution. We introduce \textbf{Agent Planning Benchmark (APB)}, a planning-specific diagnostic benchmark with 4,209 multimodal cases across 22 domains and five settings, covering holistic planning, feedback-conditioned step-wise planning, and robustness under extraneous tools, broken tools, and unsolvable tasks. Across 12 MLLMs, APB reveals systematic weaknesses in long-horizon planning, tool-noise robustness, calibrated refusal, and inference-time refinement. We further validate APB on 200 ToolSandbox tasks and 200 $\tau^2$-bench tasks, where APB-guided refinement consistently improves plan correctness, plan grade, and downstream execution metrics across three representative models. APB thus serves as an upstream diagnostic complement to execution benchmarks. The APB benchmark and code are available in \href{https://github.com/Mikivishy/AgentPlanningBenchmark}{this URL}.

\end{abstract}

\section{Introduction}

Large Language Models (LLMs) have rapidly evolved from passive text generators into agents that plan, call tools, and interact with digital environments. In successful agent systems, planning is not an auxiliary behavior but the organizing layer that determines how goals are decomposed, which tools are selected, how intermediate evidence is used, and when the agent should stop or refuse. This role is reflected in ReAct and Reflexion, where reasoning traces and feedback-driven correction guide action~\citep{yao2022react,shinn2023reflexion}; in open-ended agents such as Voyager~\citep{wang2023voyager}; and in workflow-level systems such as MetaGPT, LLMCompiler, and SWE-agent~\citep{DBLP:conf/iclr/HongZCZCWZWYLZR24,kim2024llm,yang2024sweagent}. These systems demonstrate that planning is a foundational capability for reliable agents, shaping both high-level task decomposition and downstream tool-use behavior.

\begin{figure}[t] 
  \centering
  \includegraphics[width=1\columnwidth]{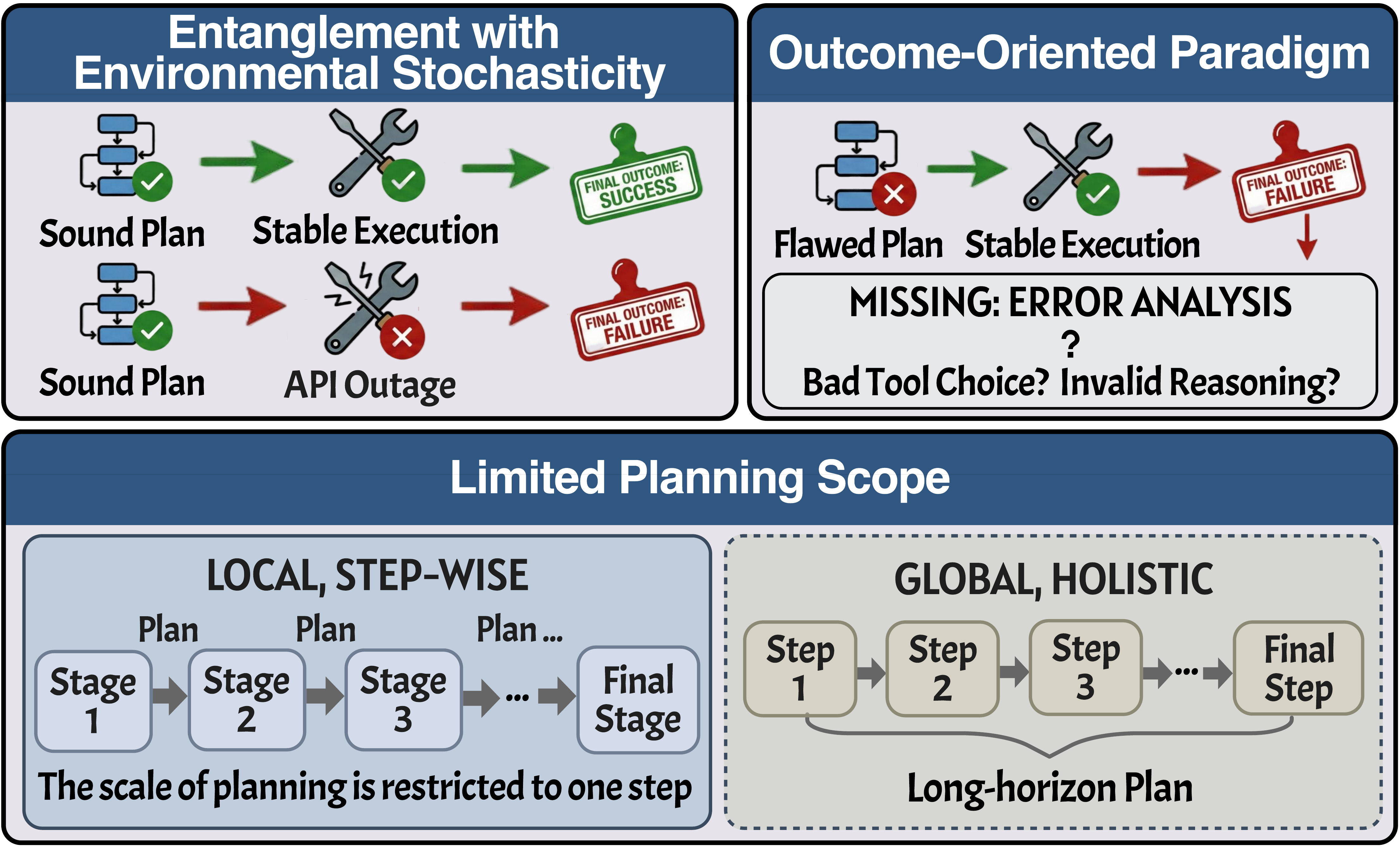}
  \caption{Systematic limitations in existing agent planning benchmarks.}
  \label{fig:head_misalignment_scenarios}
  \vspace{-17pt}
\end{figure}

\begin{figure*}[t!]
  \centering
  \includegraphics[width=1\textwidth]{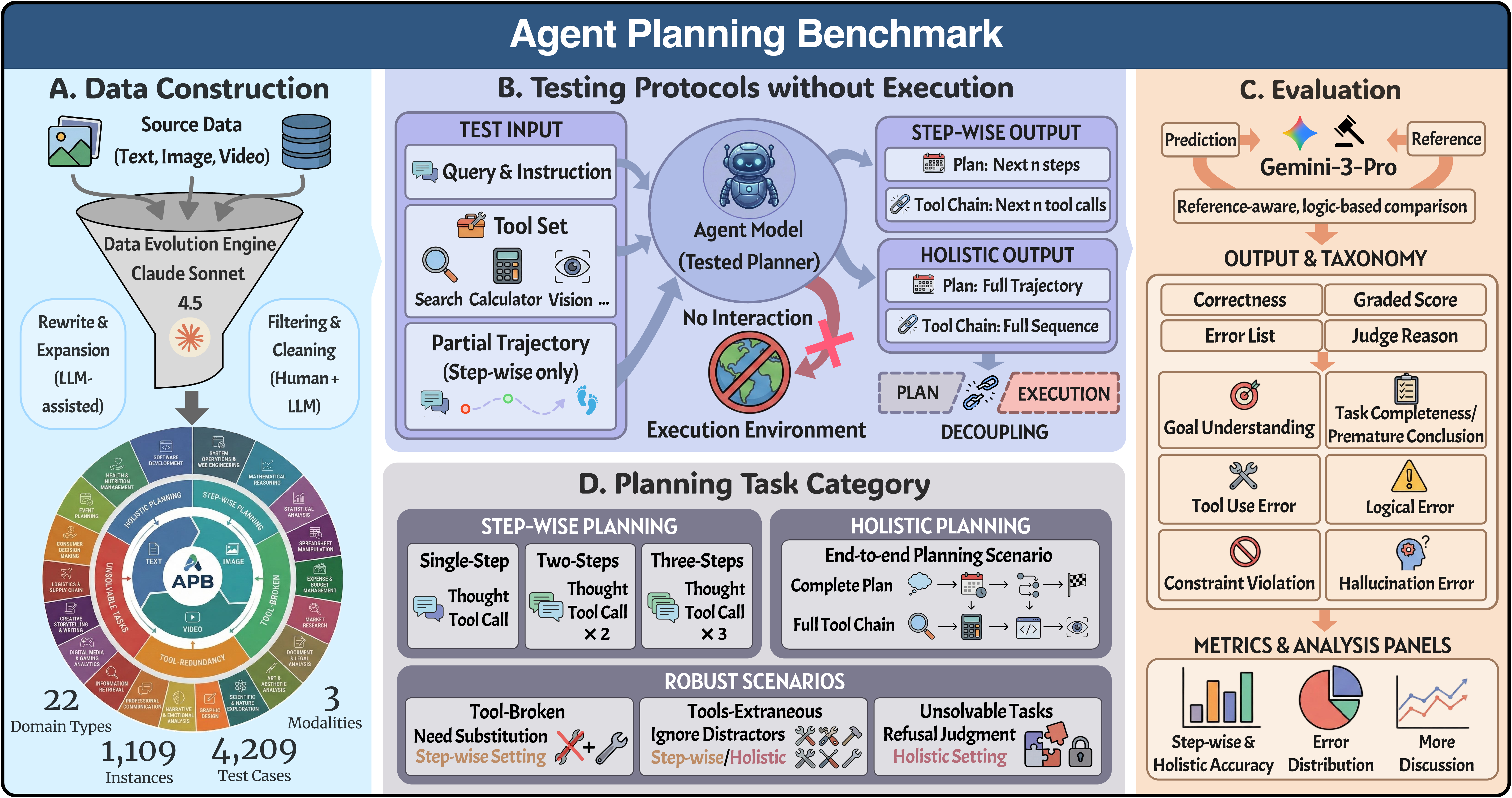}
  \caption{\textbf{Overview of APB.} The framework comprises:
    \textbf{(A) Data Construction:} A pipeline for synthesizing complex planning instances via evolution and filtering.
    \textbf{(B) Planning Protocols:} Holistic and feedback-conditioned step-wise tasks for evaluating planning logic.
    \textbf{(C) LLM-as-Judge:} Automated logic-based assessment providing a comprehensive error taxonomy.
    \textbf{(D) Task Categories:} Five complementary tasks for comprehensive assessment.}
  \label{fig:apb_framework}
  \vspace{-12pt}
\end{figure*}

Despite this importance, current evaluations do not yet provide a sufficiently diagnostic view of planning. End-to-end benchmarks are indispensable for deployed behavior, but final success entangles plan quality, tool invocation, environmental instability, and recovery~\citep{DBLP:conf/iclr/0036YZXLL0DMYZ024,deng2023mind2web}. As Figure~\ref{fig:head_misalignment_scenarios} illustrates, such outcome-oriented evaluation often leaves failure causes ambiguous. Meanwhile, planning-oriented benchmarks frequently focus on narrow or static formulations, leaving limited coverage of global long-horizon planning, feedback-conditioned local planning, noisy tool spaces, and feasibility assessment under missing information.

To address this gap, we introduce the \textbf{Agent Planning Benchmark (APB)}, a planning-focused diagnostic benchmark comprising 4,209 multimodal test cases across 22 categories. APB evaluates planning at multiple granularities: \textbf{Holistic Planning} asks models to produce complete plans and tool chains for long-horizon tasks, while \textbf{Step-wise Planning} conditions models on partial execution trajectories and tool-return feedback. APB further stresses robustness through extraneous tools, broken tools, and unsolvable tasks. Rather than only asking whether an agent succeeds, APB diagnoses \emph{why} a plan succeeds or fails through Plan Correctness, Plan Grade, and an E1--E6 error taxonomy. Our executable validation further examines APB's practical signal in controlled environments. On 200 ToolSandbox~\citep{DBLP:conf/naacl/LuHZANBMMLYWP25} and 200 \tautwobench{}~\citep{barres2025tau} tasks, we compare direct execution, plan-first execution, and APB-guided refinement across GPT-4o, Qwen3-VL-235B-A22B, and Gemini 2.5 Flash. Refined plans consistently improve downstream execution metrics, supporting APB as an upstream diagnostic signal connected to executable agent behavior.

Our evaluation across 12 Multimodal Large Language Models (MLLMs) reveals substantial variation in planning capability. Newer proprietary models dominate long-horizon holistic planning, while open-source systems remain fragile under tool noise and feasibility constraints. We further find that inference-time refinement is highly effective for holistic planning, whereas short-horizon step-wise decisions benefit less from extended reflection and may suffer from over-correction. These findings show that planning quality is not a monolithic capability: it differs across horizons, feedback conditions, and robustness settings.
%
%
To summarize, our contributions are as follows:

\begin{itemize}[leftmargin=*, label=\small\textbullet, itemsep=2.5pt, parsep=0pt, topsep=0pt]

    \item We propose APB, a planning-specific diagnostic benchmark with 4,209 multimodal cases across holistic, feedback-conditioned step-wise, and robustness-oriented planning tasks.
    \item We design a hierarchical evaluation framework with Plan Correctness, Plan Grade, and a human-informed E1--E6 error taxonomy, enabling fine-grained root-cause analysis.
    \item We evaluate 12 MLLMs and identify systematic weaknesses across horizons, feedback conditions, and robustness settings.
    \item We validate APB-guided refinement on ToolSandbox and \tautwobench{}, showing gains in both planning and execution metrics.
\end{itemize}


\section{Related Work}
\vspace{-2pt}

\paragraph{Planning in LLM-based Agents}
The paradigm of autonomous agents has evolved from simple reasoning loops to systems capable of multi-step planning and environmental interaction \citep{DBLP:journals/chinaf/XiCGHDHZWJZZFWXZWJZLYDW25, DBLP:journals/corr/abs-2503-21460, DBLP:journals/corr/abs-2401-03568, DBLP:journals/corr/abs-2401-05459}.
Foundational frameworks such as ReAct \citep{yao2022react} and Reflexion \citep{shinn2023reflexion} bind reasoning, action selection, and feedback-based revision, showing that plans and intermediate reasoning traces are central to agent behavior. Planning-centric systems further rely on task decomposition, memory, scheduling, and agent-computer interfaces to improve downstream execution, as in Voyager, MetaGPT, LLMCompiler, and SWE-agent~\citep{wang2023voyager,DBLP:conf/iclr/HongZCZCWZWYLZR24,kim2024llm,yang2024sweagent}. These lines of work motivate direct evaluation of planning: without coherent goals, constraints, and tool-use strategies, downstream execution has little chance of succeeding reliably.

\paragraph{Agent Benchmarks}
Evaluating agentic capabilities has progressed from general-purpose assessments \citep{DBLP:conf/iclr/0036YZXLL0DMYZ024, chang2024agentboard} to rigorous environment-specific testbeds.
In digital domains, benchmarks span web browsing \citep{DBLP:conf/iclr/ZhouX0ZLSCOBF0N24}, operating systems \citep{DBLP:conf/nips/XieZCLZCHCSLLXZ24, DBLP:conf/icml/BonattiZBDALLWK25}, and mobile interfaces \citep{DBLP:conf/acl/DengXSLT0L000S24}.
Concurrent efforts target specific skills, including tool utilization \citep{wang2024gta, DBLP:conf/iclr/HuangSLFWZ000G024, shen2024taskbench, DBLP:conf/coling/YeLGHWLFDJ0G025}, code generation \citep{DBLP:conf/iclr/YangJZLYWPMSNY025, si2025design2code}, and scientific reasoning \citep{wang2023scibench}, alongside embodied interaction benchmarks \citep{shridhar2020alfworld, DBLP:conf/iclr/LiuZ0ISXZLSYYQY25}.
PlanCraft evaluates long-horizon, multimodal, and deliberately unsolvable planning problems in the Minecraft crafting domain~\citep{dagan2024plancraft}.
Recently, attention has narrowed toward the cognitive core of agents, especially planning and reasoning capabilities \citep{li2025planet, wei2025plangenllms}.
APB complements these efforts through its combined coverage of 22 application domains, text/image/video inputs, holistic and feedback-conditioned step-wise planning, robustness to extraneous and broken tools in addition to unsolvable tasks, and E1--E6 diagnostic evaluation. This design provides a diagnostic view that end-to-end benchmarks cannot offer, while our executable validation examines whether APB-guided planning improvements translate to controlled execution settings.

\section{Agent Planning Benchmark}
\subsection{Task Definition}
APB defines two core tasks: \textbf{Holistic} and \textbf{Step-wise Planning}. To systematically assess planning robustness under realistic tool-use conditions, we further introduce three adversarial variants that test tool selection under noisy environments, recovery from broken tools, and principled refusal.

\paragraph{Holistic Planning (1,109 instances)}
Holistic planning evaluates whether a model can construct a complete solution plan before acting. Given a query $Q$, a system prompt $S$, and a tool set $\mathcal{T}$, the tested model is required to produce a complete solution path in a single pass:
$f_{\text{holistic}}: (Q, S, \mathcal{T}) \rightarrow (P, \mathcal{C}) .$
Specifically, the model jointly generates:
\begin{itemize}[leftmargin=*, label=\small\textbullet, itemsep=2.5pt, parsep=0pt, topsep=0pt]
    \item \textbf{Plan $P$}: A high-level decomposition translating abstract goals into logically coherent steps;
    \item \textbf{Tool Chain $\mathcal{C} = [e_1, e_2, \ldots, e_m]$}: a sequence of $m$ tool invocation nodes, with each node $e_i = (t_i, \theta_i, r_i)$ comprising tool identifier $t_i$, invocation arguments $\theta_i$, and explicit rationale $r_i$.
\end{itemize}

\paragraph{Step-wise Planning (900 instances)}

In contrast, effective agent behavior also requires localized reasoning conditioned on evolving execution contexts and prior tool feedback.

Given a system prompt $S$, a tool set $\mathcal{T}$, and a partially executed trajectory $\tau_{1:j}$ that includes the initial query, previous actions, and observed tool returns, the model is tasked with predicting the next $k$ actions:
$f_{\text{step-wise}}: (S, \mathcal{T}, \tau_{1:j}) \rightarrow (a_{j+1}, a_{j+2}, \ldots, a_{j+k}) .$
Each action $a_i = (c_i, t_i, \theta_i)$ comprises dialogue content $c_i$ and a tool invocation $(t_i, \theta_i)$. According to the prediction horizon $k$, we further distinguish:
\begin{itemize}[leftmargin=*, label=\small\textbullet, itemsep=2.5pt, parsep=0pt, topsep=0pt]
    \item \textbf{Single-step Planning} ($k=1$): predicting only the immediate next action;
    \item \textbf{Multi-step Planning} ($k \in \{2, 3\}$): predicting the subsequent two or three actions.
\end{itemize}

This design targets feedback-conditioned planning: the model uses execution history to choose its next action while avoiding premature conclusions.


\paragraph{Tool-Extraneous Planning (1,500 instances)}
To simulate open-domain noise, this task injects $n_e$ semantically similar but irrelevant tools $\mathcal{T}_r$ into $\mathcal{T}$, forming $\mathcal{T}' = \mathcal{T} \cup \mathcal{T}_r$, where $n_e \in \{2, 4, 6, 8, 10\}$:
$f_{\text{extraneous}}: (Q, S, \mathcal{T}') \rightarrow (P, \mathcal{C}) .$
The task contains 1,500 cases expanded from 150 holistic and 150 step-wise instances.

\paragraph{Tool-Broken Planning (300 instances)}
This task evaluates recovery from tool failures. Based on step-wise planning, we replace the return value of a critical tool $t_j$ with an erroneous output $o_j^{\text{error}}$, while introducing an alternative tool $t_j'$:
$f_{\text{broken}}: (S, \mathcal{T} \cup \{t_j'\}, \tau_{1:j-1}, o_j^{\text{error}}) \rightarrow (a_{j+1}) .$

\paragraph{Unsolvable Planning (400 instances)}
To evaluate principled refusal, we construct logically unsolvable instances that require $f_{\text{unsolvable}}: (Q, S, \mathcal{T}) \rightarrow \texttt{REJECT}$. We derive 400 cases from 100 holistic instances across contradictory constraints, missing information, inaccessible visual evidence, and tool removal (Appendix~\ref{app:unsolvable_synthesis}).

\begin{table*}[t]
\centering
\setlength{\aboverulesep}{0pt}
\setlength{\belowrulesep}{0pt}

\resizebox{\textwidth}{!}{
\begin{tabular}{l cccccccc | ccccc | cccc}
\toprule
\multirow{2}{*}{\textbf{Model}} & \multicolumn{8}{c|}{\textbf{Holistic Planning}} & \multicolumn{9}{c}{\textbf{Tool-Extraneous}} \\
\cmidrule(lr){2-9} \cmidrule(lr){10-18}
& \textbf{CR} & \textbf{Grade} & \textbf{E1} & \textbf{E2} & \textbf{E3} & \textbf{E4} & \textbf{E5} & \textbf{E6} & \textbf{CR} & \textbf{Grade} & \textbf{E1} & \textbf{E2-H} & \multicolumn{1}{c}{\textbf{E2-S}} & \textbf{E3} & \textbf{E4} & \textbf{E5} & \textbf{E6} \\
\midrule
\textbf{GPT-5} & \textbf{74.5} & \textbf{0.91} & 0.3 & 2.9 & 16.4 & 6.3 & 2.3 & 3.8 & 67.0 & 0.81 & 4.7 & 0.9 & \multicolumn{1}{c}{9.0} & 8.0 & 8.6 & 15.8 & 1.9 \\
\textbf{Gemini 3 Pro} & \underline{71.3} & \underline{0.89} & 1.1 & 6.5 & 16.7 & 6.9 & 4.7 & 2.7 & \textbf{76.4} & 0.87 & 3.2 & 1.1 & \multicolumn{1}{c}{7.4} & 8.0 & 7.6 & 8.8 & 1.5 \\
\textbf{Claude Sonnet 4.5} & 64.1 & 0.86 & 1.0 & 4.5 & 24.4 & 11.4 & 3.2 & 5.3 & \underline{76.4} & \textbf{0.88} & 2.4 & 1.9 & \multicolumn{1}{c}{4.8} & 10.7 & 7.8 & 6.7 & 2.7 \\
\textbf{Gemini 2.5 Pro} & 55.0 & 0.81 & 1.1 & 9.3 & 26.4 & 15.2 & 5.2 & 10.7 & 66.8 & 0.81 & 3.7 & 3.1 & \multicolumn{1}{c}{11.5} & 13.7 & 10.9 & 10.0 & 4.0 \\
\textbf{Gemini 2.5 Flash} & 36.5 & 0.69 & 1.5 & 31.9 & 28.0 & 17.5 & 6.3 & 10.2 & 59.8 & 0.79 & 3.5 & \textbf{4.4} & \multicolumn{1}{c}{28.4} & 14.5 & 13.1 & 7.3 & 4.4 \\
\textbf{GPT-4o} & 19.5 & 0.62 & 1.7 & \underline{33.5} & 43.7 & 36.4 & 15.2 & 9.3 & 45.5 & 0.70 & \underline{4.9} & 3.3 & \multicolumn{1}{c}{\underline{34.2}} & 20.3 & 24.9 & 18.4 & 4.6 \\
\hdashline
\textbf{InternVL3.5-241B-A28B} & 18.1 & 0.62 & 2.5 & 25.9 & 46.6 & 42.0 & 15.2 & 12.1 & 48.7 & 0.73 & 3.5 & 1.7 & \multicolumn{1}{c}{26.2} & 19.7 & 27.3 & 15.5 & 6.2 \\
\textbf{InternVL3.5-30B-A3B} & 8.4 & 0.49 & \textbf{7.8} & \textbf{47.9} & \underline{49.7} & \textbf{45.1} & \textbf{25.6} & 15.1 & 33.4 & 0.60 & \textbf{7.8} & 3.2 & \multicolumn{1}{c}{\textbf{49.3}} & \underline{23.5} & \textbf{37.1} & \textbf{24.5} & 8.9 \\
\textbf{InternVL3.5-38B} & 16.2 & 0.61 & 2.3 & 25.5 & 46.4 & \underline{43.2} & \underline{21.1} & 17.0 & 45.3 & 0.71 & 4.0 & 3.1 & \multicolumn{1}{c}{24.7} & 19.5 & \underline{29.1} & \underline{19.5} & 7.8 \\
\textbf{Qwen3VL-235B-A22B-In.} & 21.8 & 0.64 & 3.0 & 17.8 & 44.6 & 32.7 & 8.8 & \underline{25.5} & 52.6 & 0.74 & 3.8 & 2.8 & \multicolumn{1}{c}{15.6} & 19.6 & 21.6 & 12.5 & \underline{11.5} \\
\textbf{Qwen3VL-30B-A3B-In.} & 10.7 & 0.55 & \underline{3.1} & 26.3 & \textbf{50.0} & 38.1 & 17.6 & \textbf{32.8} & 40.2 & 0.67 & 4.7 & \textbf{4.4} & \multicolumn{1}{c}{29.4} & \textbf{23.7} & 26.1 & 18.9 & \textbf{15.9} \\
\textbf{Qwen3VL-32B-In.} & 22.5 & 0.64 & 1.5 & 17.0 & 48.2 & 28.5 & 13.5 & 24.6 & 51.9 & 0.73 & 3.6 & 2.1 & \multicolumn{1}{c}{12.5} & 17.5 & 17.6 & 9.2 & 9.2 \\
\midrule 
\multirow{2}{*}{\textbf{Model}} & \multicolumn{8}{c|}{\textbf{Step-wise Planning}} & \multicolumn{5}{c|}{\textbf{Tool-Broken}} & \multicolumn{4}{c}{\textbf{Unsolvable}} \\
\cmidrule(lr){2-9} \cmidrule(lr){10-14} \cmidrule(lr){15-18}
& \textbf{CR} & \textbf{Grade} & \textbf{E1} & \textbf{E2} & \textbf{E3} & \textbf{E4} & \textbf{E5} & \textbf{E6} & \textbf{Rp} & \textbf{Al} & \textbf{Rt} & \textbf{Rf} & \textbf{Ot} & \textbf{Conf} & \textbf{Info} & \textbf{Tool} & \textbf{Vis} \\
\midrule
\textbf{GPT-5} & \underline{77.4} & 0.85 & 3.9 & 0.8 & 5.9 & 16.3 & 8.2 & \textbf{0.0} & 76.3 & 17.3 & 5.7 & 0.0 & 0.7 & \textbf{80} & 40 & \textbf{75} & \textbf{89} \\
\textbf{Gemini 3 Pro} & \textbf{80.1} & \textbf{0.87} & 4.2 & 0.9 & 4.3 & 12.7 & 7.8 & 0.3 & \textbf{78.0} & 18.3 & 0.7 & 0.0 & 3.0 & 65 & 43 & 65 & 16 \\
\textbf{Claude Sonnet 4.5} & 77.3 & \underline{0.86} & 3.0 & 1.6 & 3.9 & 16.2 & 7.0 & 1.7 & 76.7 & 14.7 & 4.3 & 0.3 & 4.0 & 72 & 19 & 16 & 53 \\
\textbf{Gemini 2.5 Pro} & 69.2 & 0.79 & 4.4 & 2.0 & 9.3 & 18.0 & 10.6 & \textbf{2.6} & 74.7 & 12.3 & 10.0 & \underline{1.0} & 2.0 & 77 & \textbf{65} & \textbf{75} & 11 \\
\textbf{Gemini 2.5 Flash} & 66.6 & 0.78 & 5.6 & 4.4 & 9.0 & 21.4 & 12.7 & 2.2 & 44.3 & 28.3 & \textbf{24.3} & \textbf{2.0} & 1.0 & 74 & \underline{56} & 72 & 36 \\
\textbf{GPT-4o} & 57.6 & 0.73 & 5.1 & 2.7 & 7.4 & 31.3 & 16.7 & 2.1 & 42.3 & \textbf{46.3} & 6.3 & 0.3 & 4.7 & 78 & 39 & 30 & 64 \\
\hdashline
\textbf{InternVL3.5-241B-A28B} & 59.2 & 0.74 & 6.6 & 2.0 & 7.0 & 30.3 & 16.1 & 1.4 & 74.7 & 19.0 & 4.0 & 0.0 & 2.3 & 75 & 39 & 38 & 38 \\
\textbf{InternVL3.5-30B-A3B} & 38.8 & 0.58 & \textbf{9.0} & 2.0 & \textbf{20.3} & \textbf{44.3} & \textbf{27.0} & 1.3 & 37.7 & \underline{38.3} & \underline{15.7} & 0.0 & \textbf{8.3} & 69 & 35 & 18 & 50 \\
\textbf{InternVL3.5-38B} & 49.7 & 0.68 & 7.1 & 2.1 & 7.4 & \underline{38.9} & \underline{21.6} & 1.1 & 57.0 & 29.0 & 9.0 & 0.7 & 4.3 & 60 & 6 & 10 & 20 \\
\textbf{Qwen3VL-235B-A22B-In.} & 60.2 & 0.74 & 6.4 & 1.4 & 7.8 & 30.1 & 13.3 & \underline{2.4} & \underline{77.7} & 16.7 & 4.3 & 0.0 & 1.3 & 72 & 11 & 7 & 64 \\
\textbf{Qwen3VL-30B-A3B-In.} & 49.6 & 0.68 & \underline{8.0} & \underline{3.2} & \underline{9.9} & 38.4 & 19.1 & 1.3 & 55.7 & 33.3 & 7.0 & 0.0 & 4.0 & 61 & 10 & 2 & 44 \\
\textbf{Qwen3VL-32B-In.} & 59.9 & 0.75 & 5.6 & 1.2 & 8.4 & 29.0 & 15.1 & 0.8 & 68.7 & 22.0 & 7.3 & 0.0 & 2.0 & 76 & 31 & 29 & \underline{84} \\
\bottomrule
\end{tabular}
}
\caption{\textbf{Overall APB results.} CR denotes Correctness Rate; Grade is average Plan Grade. E1--E6 report error incidence rates. Tool-Broken columns show behavior distributions, and Unsolvable columns show refusal rates across four infeasibility types.}
\label{tab:main_table_results}
\vspace{-16pt}
\end{table*}

\subsection{Data Construction}

We aggregate data from OpenCUA, GTA, GAIA, ToolBench, FrameThinker, and real traffic from a public online AI agent platform~\citep{DBLP:conf/nips/WangWLYXWDGXWSL25,wang2024gta,mialon2024gaia,DBLP:conf/iclr/QinLYZYLLCTQZHT24,he2025framethinker}. Each instance contains a system prompt, user query, tool set, and execution trajectory. We expand limited seed tasks by enriching queries, expanding tools, and simulating longer trajectories with Claude Sonnet~4.5~\citep{claude-sonnet-4.5}. Rule-based checks, two-stage LLM validation, and human verification ensure quality; holistic ground truth is then synthesized from validated trajectories. Details are in Appendix~\ref{app:data_construction}.

\subsection{Metrics}
We evaluate performance across three dimensions: correctness, severity, and error typology.

\textbf{(1) Plan Correctness} This binary metric serves as a strict indicator of planning success, determining whether the generated plan is logically sound, executable, and fully satisfies the task objectives.

\textbf{(2) Plan Grade} Plan Grade measures the severity of planning deficiencies against the task requirements rather than distance from a unique or optimal reference path. By evaluating goal attainment, completeness, constraint satisfaction, and logical consistency, this metric differentiates between near-misses and fundamental failures, utilizing a discrete scale of $\{0, 0.2, 0.4, 0.6, 0.8, 1\}$.

\textbf{(3) Error Taxonomy}

To systematically diagnose failure mechanisms in agent planning, we define a unified taxonomy with \textbf{six} semantically independent, annotatable error categories, where a single sample may exhibit multiple errors simultaneously. The taxonomy was developed through iterative human inspection of planning failures, assisted by LLM-based clustering and summarization, and then finalized through additional rounds of human checks to ensure that each category is both interpretable and consistently annotatable. Notably, E2 instantiates two protocol-specific subtypes that capture a key behavioral distinction between step-wise and holistic planning.

\begin{itemize}[leftmargin=*, label=\small\textbullet, itemsep=1.5pt, parsep=0pt, topsep=2pt]
\item \textbf{E1: Goal Understanding Error} Misinterpreting the fundamental user intent or task objective.
\item \textbf{E2: Premature Conclusion / Task Incompleteness} Critical failures in plan termination. The former denotes stopping before the task is fully solved (step-wise), while the latter indicates a failure to cover all required sub-tasks in the generated trajectory (holistic).
\item \textbf{E3: Constraint Violation} Violating explicit system prompt instructions or task constraints.
\item \textbf{E4: Logic Error} Invalid causal dependencies, illogical ordering, missing prerequisites, or circular reasoning. Deviation from the reference path alone is not an E4 error when the alternative remains feasible and satisfies the task constraints (Appendix~\ref{app:solution_diversity}).
\item \textbf{E5: Tool Use Error} Misunderstanding intended tool functionality or operational semantics.
\item \textbf{E6: Hallucination} Generating unsupported fabricated facts or hallucinated intermediate results.
\end{itemize}

This hierarchical framework enables both quantitative benchmarking and root-cause diagnosis: Correctness signals planning success, Grade quantifies severity, and Taxonomy identifies failure mechanisms (see Appendix~\ref{app:evaluation_metrics}). In practice, low APB scores indicate intrinsic planning deficits such as missing constraints or invalid tool choices, while high APB scores paired with low execution success can help localize bottlenecks to execution-side engineering. We operationalize the metrics via an LLM-as-Judge suite to validate logical coherence against reference trajectories (see Appendix~\ref{app:llm_judge_analysis}).

\section{Experiments}
\subsection{Evaluation Settings}
We evaluate 12 SOTA MLLMs on APB, including six open-source models (the Qwen3VL series: Qwen3VL-30B-A3B-Instruct, Qwen3VL-32B-Instruct, and Qwen3VL-235B-A22B-Instruct \citep{DBLP:journals/corr/abs-2511-21631}; and the InternVL3.5 series: InternVL3.5-30B-A3B, InternVL3.5-38B, and InternVL3.5-241B-A28B \citep{DBLP:journals/corr/abs-2508-18265}) and six proprietary models (Claude Sonnet 4.5 \citep{claude-sonnet-4.5}, GPT-5 \citep{gpt-5}, GPT-4o \citep{4o}, Gemini 3 Pro \citep{gemini-3-pro}, and Gemini 2.5 series (Pro/Flash) \citep{DBLP:journals/corr/abs-2507-06261}).

\begin{table}[htbp]
\centering
\setlength{\tabcolsep}{2.0pt}
\renewcommand{\arraystretch}{1.08}
\definecolor{dropLvl1}{RGB}{255, 235, 235} 
\definecolor{dropLvl2}{RGB}{255, 200, 200} 
\definecolor{dropLvl3}{RGB}{255, 150, 150} 
\newcommand{\gDropLight}[1]{\cellcolor{dropLvl1}#1}     
\newcommand{\gDropMed}[1]{\cellcolor{dropLvl2}\textbf{#1}} 
\newcommand{\gDropHeavy}[1]{\cellcolor{dropLvl3}\textbf{#1}} 
\newcommand{\cbox}[1]{\colorbox{#1}{\phantom{m}}}
\resizebox{\linewidth}{!}{
\begin{tabular}{l cccc cccc}
\toprule
\multirow{2}{*}{\textbf{Model}} &
\multicolumn{4}{c}{\textbf{Holistic}} &
\multicolumn{4}{c}{\textbf{Step-wise}} \\
\cmidrule(lr){2-5} \cmidrule(lr){6-9}
& \textbf{TC} & \textbf{CR\_B} & \textbf{CR\_E} & \textbf{$\Delta$} &
  \textbf{TC} & \textbf{CR\_B} & \textbf{CR\_E} & \textbf{$\Delta$} \\
\midrule
\textbf{GPT-5} & \textbf{115} & 77.3 & 55.4 & \gDropHeavy{-22.0} & 23 & 85.3 & 78.7 & \gDropMed{-6.7} \\
\textbf{Gemini 3 Pro} & 46 & 77.2 & 69.9 & \gDropMed{-7.3} & 12 & 80.0 & 82.9 & +2.9 \\
\textbf{Claude Sonnet 4.5} & \textbf{50} & 79.3 & 67.0 & \gDropMed{-12.3} & 6 & 86.7 & 85.7 & \gDropLight{-0.9} \\
\textbf{Gemini 2.5 Pro} & \textbf{60} & 61.3 & 54.7 & \gDropMed{-6.6} & 7 & 80.0 & 78.9 & \gDropLight{-1.1} \\
\textbf{Gemini 2.5 Flash} & 27 & 53.3 & 40.2 & \gDropMed{-13.1} & 13 & 78.0 & 79.3 & +1.3 \\
\textbf{GPT-4o} & \textbf{77} & 26.7 & 21.4 & \gDropMed{-5.3} & 15 & 73.3 & 69.6 & \gDropLight{-3.7} \\
\textbf{InternVL3.5-241B} & \textbf{80} & 22.3 & 23.4 & +1.1 & 15 & 77.3 & 74.0 & \gDropLight{-3.3} \\
\textbf{InternVL3.5-30B} & \textbf{78} & 12.0 & 8.9 & \gDropLight{-3.1} & 6 & 60.0 & 57.9 & \gDropLight{-2.1} \\
\textbf{InternVL3.5-38B} & \textbf{98} & 24.0 & 25.2 & +1.2 & 9 & 67.3 & 65.3 & \gDropLight{-2.0} \\
\textbf{Qwen3VL-235B} & \textbf{73} & 30.7 & 29.7 & \gDropLight{-1.0} & 5 & 74.0 & 75.5 & +1.5 \\
\textbf{Qwen3VL-30B} & \textbf{97} & 16.0 & 12.7 & \gDropLight{-3.3} & 13 & 64.0 & 67.7 & +3.7 \\
\textbf{Qwen3VL-32B} & \textbf{108} & 32.7 & 31.2 & \gDropLight{-1.5} & 21 & 72.0 & 72.5 & +0.5 \\
\bottomrule
\end{tabular}
}
\caption{\textbf{Impact of extraneous tools.} TC is extraneous-tool usage frequency; CR\_B/CR\_E are correctness rates in base and extraneous settings.}
\label{tab:redundancy_effect}
\vspace{-15pt}
\end{table}

\subsection{Main Results}

\subsubsection{Overall Performance}
\begin{figure}[htbp]
    \centering
    \includegraphics[width=1.0\columnwidth, keepaspectratio]{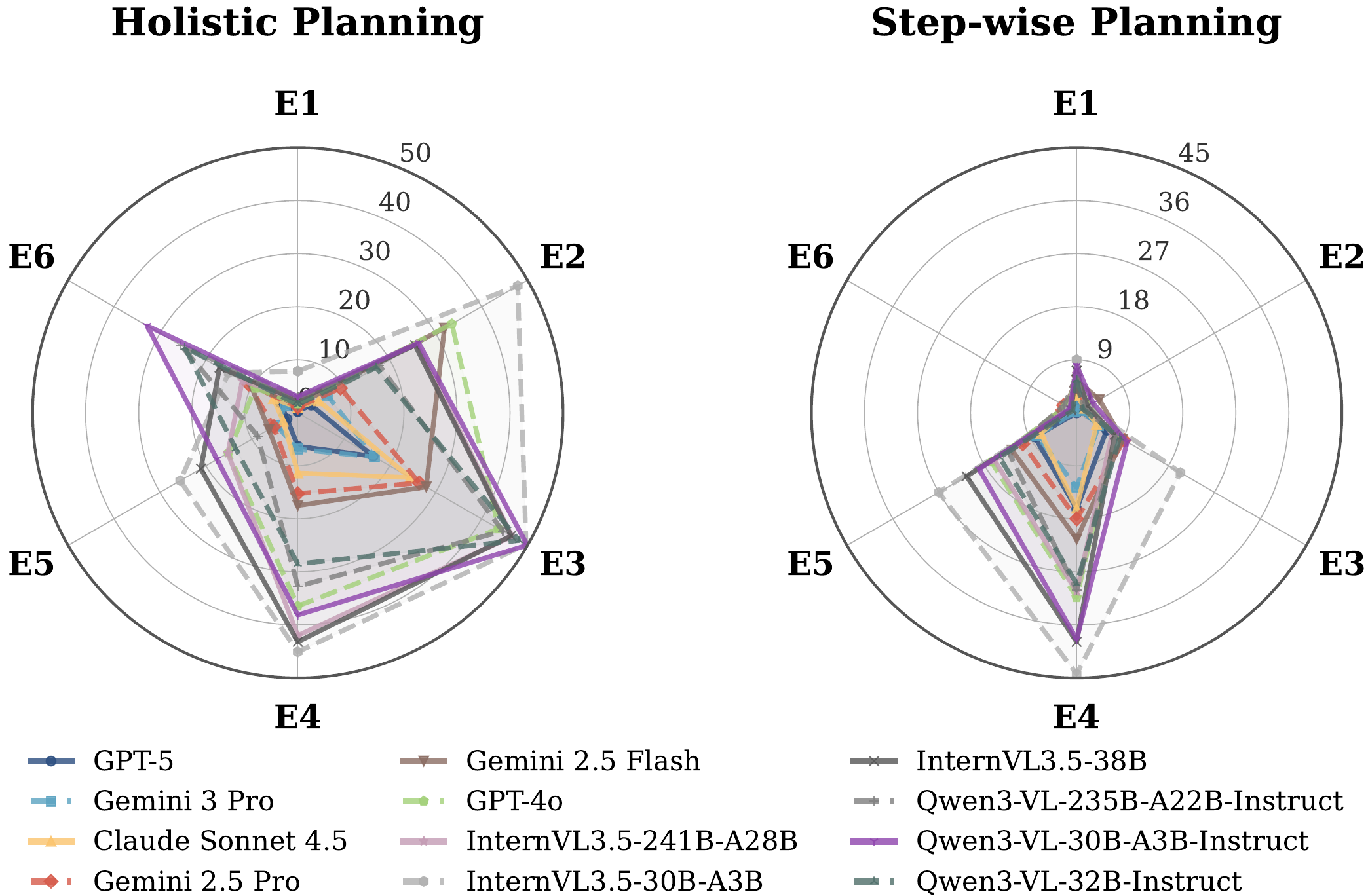}
    \caption{Comparative error distributions across planning modes.}
    \label{fig:error_distribution_2}
    \vspace{-16pt}
\end{figure}

Table~\ref{tab:main_table_results} shows clear divergence across planning modes. All models drop from feedback-conditioned step-wise planning to holistic planning, with open-source models particularly vulnerable to long-horizon global reasoning. This gap indicates that complete tool-use strategies require more than locally reasonable next-step decisions. Notably, the dense Qwen3-VL-32B-Instruct matches or outperforms much larger MoE models in both holistic and step-wise tasks. Figure~\ref{fig:error_distribution_2} further shows that Logic Error (E4) is a shared bottleneck, while holistic planning also suffers from incompleteness, constraint violations, and hallucination. Step-wise planning instead concentrates failures around local tool use and logic, reflecting its narrower decision scope and access to prior feedback.

This contrast is central to APB's design. Holistic planning exposes whether a model can maintain global consistency across a complete plan, while step-wise planning asks whether it can use the current trajectory state to choose the next immediate action. The aggregate model rankings are strongly correlated, yet a matched evaluation on the same 300 underlying samples reveals a systematic within-model difficulty gap between the two settings; full paired results and significance tests are reported in Appendix~\ref{app:matched_planning_analysis}.

\begin{figure}[t]
    \centering
    \includegraphics[width=1.0\columnwidth, keepaspectratio]{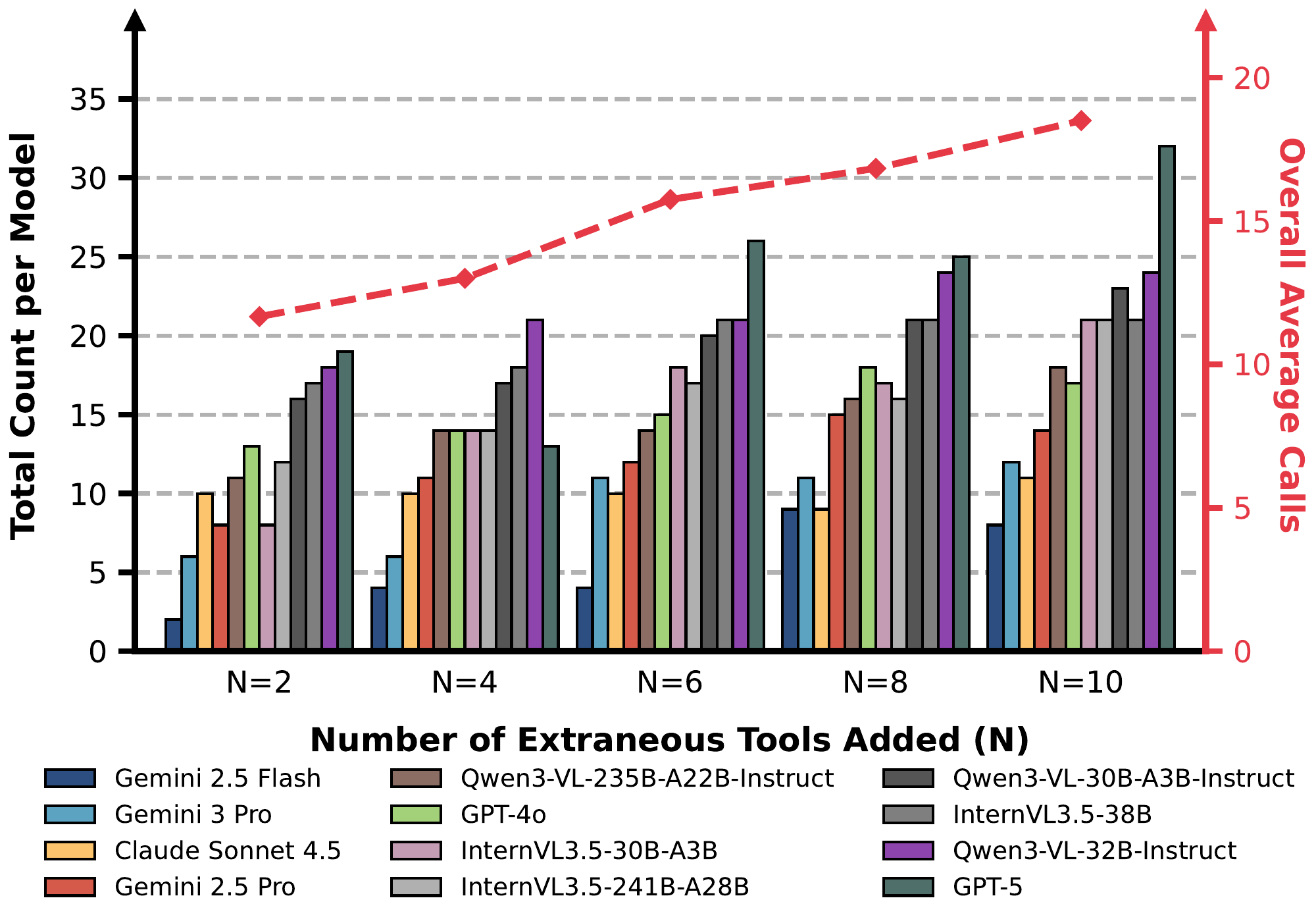}
    \caption{Trends in extraneous tool usage. The red dashed line indicates the overall average count.}
    \label{fig:Redundant_tool}
    \vspace{-8pt}
\end{figure}

\subsubsection{Analysis of Extraneous Tools}
Introducing extraneous tools significantly degrades holistic planning performance, whereas step-wise planning remains resilient. As detailed in Table~\ref{tab:redundancy_effect}, extraneous tools cause broad performance degradation: top-tier models such as GPT-5 and Claude Sonnet 4.5 exhibit substantial holistic accuracy declines of 22.0\% and 12.3\%, in contrast to the much smaller 0.9\%--6.7\% drops in step-wise mode. This vulnerability is further corroborated by Figure~\ref{fig:Redundant_tool}, which illustrates a positive correlation between extraneous-tool availability and erroneous utilization within holistic planning trajectories.

This disparity highlights distinct paradigmatic vulnerabilities: holistic planning requires one-shot trajectory generation, where expanding the tool library amplifies the cumulative probability of selection errors, leading to cascading downstream plan failures. Conversely, step-wise planning relies on isolated, locally optimal decisions, rendering it comparatively more resilient to noise interference.

\subsubsection{Adaptability to Broken Tool Setting}
In the Tool-Broken task, we classify model behaviors into five distinct categories: \textbf{Replace} (switching to the designated substitute tool), \textbf{Alter} (using a non-substitute tool for a different action), \textbf{Retry} (reattempting the same tool), \textbf{Refuse} (declining to act), and \textbf{Other}. Among these, \textbf{Replace} is regarded as the optimal behavior, indicating successful recovery by switching tools to continue the task.

Our analysis reveals different behavioral patterns. Gemini 3 Pro and Qwen3VL-235B-A22B-Instruct demonstrate superior adaptability with high Replace rates ($>$77\%), underscoring their proficiency in identifying valid substitutes. Conversely, GPT-4o and Gemini 2.5 Flash record lower Replace proportions ($<$45\%), diverging significantly into Alter or Retry actions, respectively. This indicates that robust planners favor adaptive replacement, whereas other models struggle with uncertainty, resorting to divergent or repetitive behaviors.

\vspace{4pt}
\subsubsection{Awareness of Unsolvable Tasks}
As shown in Table~\ref{tab:main_table_results}, GPT-5 demonstrates superior robustness, yet reveals a general gap where models handle explicit Constraint Conflict far better than Information Missing. This disparity suggests that current models can readily detect explicit logical contradictions stated in system prompts or instructions, but struggle with more implicit and non-explicitly stated missing information, often defaulting to hallucination rather than appropriate refusal. Moreover, open-source models notably underperform on Tool Removal, exposing deficiencies in feasibility assessment where they attempt tasks blindly despite insufficient toolsets. Even proprietary models remain vulnerable: Claude Sonnet 4.5 performs poorly on Tool Removal, while the Gemini series is notably weaker on Visual Missing scenarios. These failures highlight a persistent tendency to rely on implicit assumptions and over-completion when critical evidence is absent, rather than exercising calibrated refusal.

\begin{figure}[t]
    \centering
    \includegraphics[width=1.0\columnwidth, keepaspectratio]{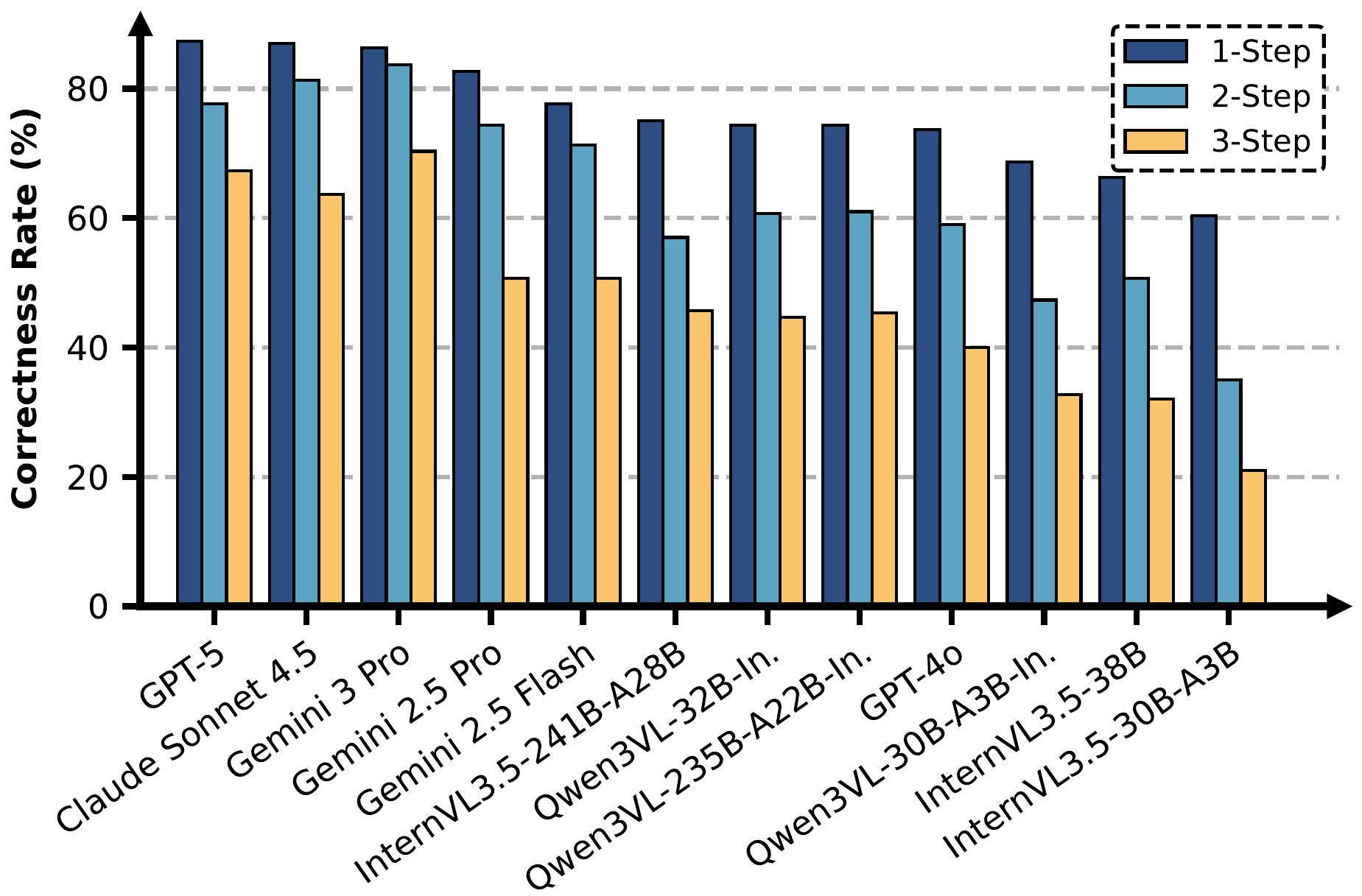}
    \caption{Step-wise Planning performance across varying prediction horizons.}
    \label{fig:step_wise_different_steps}
    \vspace{-16pt}
\end{figure}

\section{Discussion}
\subsection{Executable Validation of APB-Guided Planning}

To test whether APB's diagnostic signals translate into executable agent behavior, we validate on ToolSandbox and \tautwobench{}. For each environment, we evaluate 200 tasks under three controlled execution modes. \textbf{Direct} executes the task without planning. \textbf{Plan-first} generates an initial holistic plan $p_0$ and then executes it. \textbf{Refine} starts from the same $p_0$, applies one round of APB-guided critique, and executes the revised plan $p_1$. Gemini 3 Pro produces only Plan Grade, E1--E6 indicators, and a diagnostic rationale under an anti-leak instruction; the tested model then revises its own plan without access to hidden supervision. Plan-first and Refine otherwise use identical executors, sampling configurations, and environments, so their comparison isolates the revision effect. ToolSandbox reports average trajectory similarity (higher is better), while \tautwobench{} reports average task reward (higher is better). For APB metrics, ``$a \rightarrow b$'' denotes the score before and after APB-guided refinement. Appendix~\ref{app:executable_protocol} specifies the environment-specific supervision and complete single-round protocol.

\begin{table*}[t]
\centering
\small
\setlength{\tabcolsep}{4pt}
\resizebox{\textwidth}{!}{
\begin{tabular}{llccc cc}
\toprule
\textbf{Benchmark} & \textbf{Model} & \textbf{Direct Exec.} & \textbf{Plan-first Exec.} & \textbf{Refine Exec.} & \textbf{APB CR} & \textbf{APB Grade} \\
\midrule
ToolSandbox & GPT-4o & 76.3 & 79.2 & \textbf{83.3} & 61.5 $\rightarrow$ 84.5 & 0.66 $\rightarrow$ 0.87 \\
ToolSandbox & Qwen3-VL-235B & 80.7 & 79.1 & \textbf{84.6} & 75.5 $\rightarrow$ 88.0 & 0.79 $\rightarrow$ 0.92 \\
ToolSandbox & Gemini 2.5 Flash & 85.5 & 86.0 & \textbf{89.4} & 81.5 $\rightarrow$ 92.0 & 0.83 $\rightarrow$ 0.95 \\
\midrule
\tautwobench{} & GPT-4o & 0.54 & 0.56 & \textbf{0.63} & 35.0 $\rightarrow$ 65.5 & 0.54 $\rightarrow$ 0.78 \\
\tautwobench{} & Qwen3-VL-235B & 0.47 & 0.58 & \textbf{0.61} & 32.5 $\rightarrow$ 62.5 & 0.45 $\rightarrow$ 0.79 \\
\tautwobench{} & Gemini 2.5 Flash & 0.57 & 0.62 & \textbf{0.69} & 43.0 $\rightarrow$ 83.5 & 0.57 $\rightarrow$ 0.86 \\
\bottomrule
\end{tabular}}
\caption{\textbf{Executable validation.} Direct, Plan-first, and Refine report executable performance. APB CR and APB Grade report planning quality before and after APB-guided refinement ($a \rightarrow b$). ToolSandbox uses average trajectory similarity (\%), and \tautwobench{} uses average reward.}
\label{tab:executable_validation}
\end{table*}

Table~\ref{tab:executable_validation} shows consistent gains from APB-guided refinement. Relative to Plan-first execution of $p_0$, executing $p_1$ improves ToolSandbox similarity by 3.4--5.5 points and \tautwobench{} reward by 0.03--0.07 across the three models. Since Plan-first and Refine use the same initial draft plan and differ only in whether APB-guided critique is applied, these results demonstrate that APB provides actionable diagnostic signals for identifying and correcting planning defects that affect executable behavior.

\begin{table*}[h!]
\centering
\small
\setlength{\tabcolsep}{3.5pt}
\renewcommand{\arraystretch}{1.1}

\definecolor{riseLvl1}{RGB}{235, 241, 255}
\definecolor{riseLvl2}{RGB}{214, 228, 255}
\definecolor{riseLvl3}{RGB}{175, 202, 255}

\definecolor{dropLvl1}{RGB}{255, 235, 235}
\definecolor{dropLvl2}{RGB}{255, 200, 200}
\definecolor{dropLvl3}{RGB}{255, 150, 150}

\newcommand{\gRiseL}[1]{\cellcolor{riseLvl1}#1}
\newcommand{\gRiseM}[1]{\cellcolor{riseLvl2}\textbf{#1}}
\newcommand{\gRiseH}[1]{\cellcolor{riseLvl3}\textbf{#1}}

\newcommand{\gDropL}[1]{\cellcolor{dropLvl1}#1}
\newcommand{\gDropM}[1]{\cellcolor{dropLvl2}\textbf{#1}}
\newcommand{\gDropH}[1]{\cellcolor{dropLvl3}\textbf{#1}}

\resizebox{.9\textwidth}{!}{%
\begin{tabular}{l ccc ccc | ccc ccc}
\toprule
\multirow{2}{*}{\textbf{Method}}
& \multicolumn{3}{c}{\textbf{Holistic CR (\%)}} 
& \multicolumn{3}{c|}{\textbf{Step-wise CR (\%)}}
& \multicolumn{3}{c}{\textbf{Holistic CR (\%)}} 
& \multicolumn{3}{c}{\textbf{Step-wise CR (\%)}} \\
\cmidrule(lr){2-4} \cmidrule(lr){5-7} \cmidrule(lr){8-10} \cmidrule(lr){11-13}
& \textbf{Base} & \textbf{Refined} & \textbf{$\Delta$}
& \textbf{Base} & \textbf{Refined} & \textbf{$\Delta$}
& \textbf{Base} & \textbf{Refined} & \textbf{$\Delta$}
& \textbf{Base} & \textbf{Refined} & \textbf{$\Delta$} \\
\midrule

\multicolumn{1}{l}{} 
& \multicolumn{6}{c|}{\textbf{InternVL3.5-241B-A28B}}
& \multicolumn{6}{c}{\textbf{Qwen3VL-235B-A22B-Instruct}} \\
\cmidrule(lr){2-7} \cmidrule(lr){8-13}
\textbf{Self\_R+}   & 22.00 & 51.33 & \gRiseH{+29.33} & 79.33 & 72.00 & \gDropM{-7.33} & 29.33 & 58.00 & \gRiseH{+28.67} & 78.00 & 74.00 & \gDropL{-4.00} \\
\textbf{Critic\_R+} & 22.00 & 60.00 & \gRiseH{+38.00} & 79.33 & 81.33 & \gRiseL{+2.00}  & 29.33 & 62.00 & \gRiseH{+32.67} & 78.00 & 83.33 & \gRiseM{+5.33} \\
\textbf{Self\_R-}   & 22.00 & 28.67 & \gRiseM{+6.67}  & 79.33 & 72.67 & \gDropM{-6.67} & 29.33 & 35.33 & \gRiseM{+6.00}  & 78.00 & 67.33 & \gDropM{-10.67} \\
\textbf{Critic\_R-} & 22.00 & 42.67 & \gRiseH{+20.67} & 79.33 & 82.00 & \gRiseL{+2.67}  & 29.33 & 43.33 & \gRiseM{+14.00} & 78.00 & 82.67 & \gRiseL{+4.67} \\
\midrule

\multicolumn{1}{l}{} 
& \multicolumn{6}{c|}{\textbf{InternVL3.5-30B-A3B}}
& \multicolumn{6}{c}{\textbf{Qwen3VL-30B-A3B-Instruct}} \\
\cmidrule(lr){2-7} \cmidrule(lr){8-13}
\textbf{Self\_R+}   & 9.33 & 22.00 & \gRiseM{+12.67} & 60.00 & 43.33 & \gDropH{-16.67} & 17.33 & 38.00 & \gRiseH{+20.67} & 68.67 & 67.33 & \gDropL{-1.33} \\
\textbf{Critic\_R+} & 9.33 & 42.00 & \gRiseH{+32.67} & 60.00 & 68.00 & \gRiseM{+8.00}   & 17.33 & 48.00 & \gRiseH{+30.67} & 68.67 & 78.00 & \gRiseM{+9.33} \\
\textbf{Self\_R-}   & 9.33 & 15.33 & \gRiseM{+6.00}  & 60.00 & 51.33 & \gDropM{-8.67}  & 17.33 & 21.33 & \gRiseL{+4.00}  & 68.67 & 57.33 & \gDropM{-11.33} \\
\textbf{Critic\_R-} & 9.33 & 36.00 & \gRiseH{+26.67} & 60.00 & 68.00 & \gRiseM{+8.00}   & 17.33 & 34.67 & \gRiseH{+17.33} & 68.67 & 72.00 & \gRiseL{+3.33} \\
\midrule

\multicolumn{1}{l}{} 
& \multicolumn{6}{c|}{\textbf{InternVL3.5-38B}}
& \multicolumn{6}{c}{\textbf{Qwen3VL-32B-Instruct}} \\
\cmidrule(lr){2-7} \cmidrule(lr){8-13}
\textbf{Self\_R+}   & 28.00 & 46.67 & \gRiseH{+18.67} & 67.33 & 66.67 & \gDropL{-0.67} & 31.33 & 51.33 & \gRiseH{+20.00} & 74.00 & 74.00 & +0.00 \\
\textbf{Critic\_R+} & 28.00 & 53.33 & \gRiseH{+25.33} & 67.33 & 70.00 & \gRiseL{+2.67}  & 31.33 & 56.67 & \gRiseH{+25.33} & 74.00 & 74.67 & \gRiseL{+0.67} \\
\textbf{Self\_R-}   & 28.00 & 30.67 & \gRiseL{+2.67}  & 67.33 & 68.00 & \gRiseL{+0.67}  & 31.33 & 26.67 & \gDropL{-4.67} & 74.00 & 63.33 & \gDropM{-10.67} \\
\textbf{Critic\_R-} & 28.00 & 31.33 & \gRiseL{+3.33}  & 67.33 & 70.67 & \gRiseL{+3.33}  & 31.33 & 38.00 & \gRiseM{+6.67}  & 74.00 & 74.00 & +0.00 \\
\bottomrule
\end{tabular}%
}
\caption{\textbf{Impact of inference-time refinement.} Self\_R and Critic\_R compare Base and Refined CR; (+) uses error-taxonomy guidance.}
\label{tab:refinement_effect}
\vspace{-10pt}
\end{table*}

\begin{table}[htbp]
    \centering
    \setlength{\tabcolsep}{3.5pt}
    \resizebox{\columnwidth}{!}{
    \begin{tabular}{l cc ccc}
    \toprule
    \multirow{2}{*}{\textbf{Model}} & \multirow{2}{*}{\textbf{CR}} & \multirow{2}{*}{\textbf{Grade}} & \multicolumn{3}{c}{\textbf{Average Cost}} \\
    \cmidrule(lr){4-6}
     & & & \textbf{$\mathcal{C}_{all}$} & \textbf{$\mathcal{C}_{pass}$} & \textbf{$\mathcal{C}_{com}$} \\
    \midrule
    \textbf{GPT-5} & \underline{89.5} & \textbf{0.965} & \underline{13.21} & \underline{13.35} & \underline{13.55}  \\
    \textbf{Gemini 3 Pro} & \textbf{91.2} & 0.930 & \textbf{12.54} & \textbf{12.65} & \textbf{12.94} \\
    \textbf{Gemini 2.5 Pro} & \underline{89.5} & \underline{0.958} & 13.70 & 13.86 & 13.73\\
    \textbf{Gemini 2.5 Flash} & 82.5 & 0.930 & 14.40 & 13.98 & 14.00\\
    \textbf{Claude Sonnet 4.5} & 82.5 & 0.937 & 15.40 & 16.02 & 16.15\\
    \bottomrule
    \end{tabular}
    }

    \caption{\textbf{Planning efficiency.} CR, Plan Grade, and average cost on all, solved, and commonly solved instances.}
    \label{tab:merged_plan_efficiency}
    \vspace{-8pt}
\end{table}

\subsection{Dynamics of Step-wise Planning: Horizon and Temporal Profiles}
\begin{figure}[t]
    \centering
    \includegraphics[width=1.0\columnwidth, keepaspectratio]{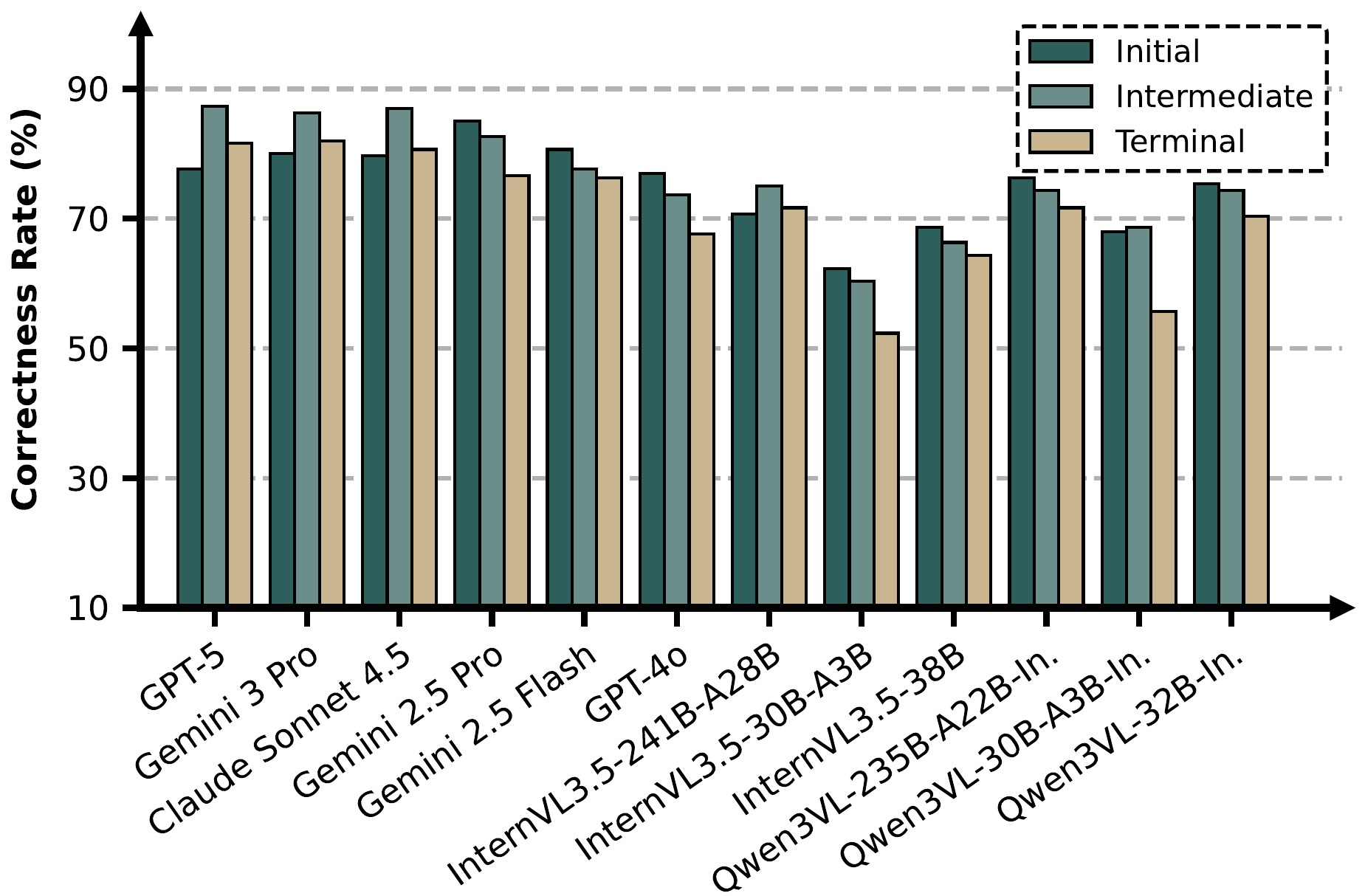}
    \caption{Step-wise Planning performance across different trajectory stages.}
    \label{fig:step_wise_different_stages}
    \vspace{-16pt}
\end{figure}

\begin{takeaway}
\textbf{\textit{Takeaway 1.}} Extended planning horizons accumulate errors; distinct models exhibit diverging performance peaks across temporal stages.
\end{takeaway}
To investigate the impact of planning horizons and historical context, we analyze performance disparities under varying prediction scopes and temporal truncation points. As illustrated in Figure~\ref{fig:step_wise_different_steps}, extending the prediction scope from 1-Step to 3-Step causes a monotonic decline in correctness across all models, highlighting the challenge of predicting multiple future actions even when prior tool feedback is available. Further analysis of interaction trajectories in Figure~\ref{fig:step_wise_different_stages} reveals divergent patterns: models such as Gemini 2.5 series and most open-source variants peak early and degrade as context increases, suggesting limitations in leveraging extended feedback histories. Conversely, models like GPT-5 reach peak performance at intermediate stages, demonstrating the ability to use accumulated context for decision optimization. These findings show why APB includes feedback-conditioned step-wise tasks in addition to holistic planning.

This analysis distinguishes two feedback-related skills. The first is reactive adaptation: using the latest observation to choose the immediate next action. The second is look-ahead consistency: projecting several future actions while preserving constraints that may only become relevant later. Current models are noticeably stronger at the former than the latter, which explains why a single success rate from an execution benchmark can hide qualitatively different planning weaknesses.

\subsection{Impact of Inference-Time Refinement}
\begin{takeaway}
\textbf{\textit{Takeaway 2.}} Refinement substantially improves holistic planning, while its gains are smaller and less stable for feedback-conditioned step-wise decisions.
\end{takeaway}
Recent advancements in test-time scaling have demonstrated that increasing inference-time compute can significantly bolster agentic performance~\citep{DBLP:journals/corr/abs-2507-23370, DBLP:journals/corr/abs-2507-05791}. To investigate the transferability of these benefits to planning, we evaluated self-refine and critic-guided strategies across four experimental settings (see Appendix~\ref{app:refinement_setting}); the results are detailed in Table~\ref{tab:refinement_effect}. In holistic planning, these methods deliver substantial gains, and critic-guided refinement with our error taxonomy is consistently strongest. For example, InternVL3.5-241B-A28B improves from 22.00\% to 60.00\% under taxonomy-guided critic refinement. This supports our central claim that holistic planning benefits from explicit diagnosis and revision. By contrast, step-wise planning already conditions on execution feedback and often concerns short-horizon decisions; adding extra reflection can induce over-correction or hallucinated alternatives (Appendix~\ref{app:refinement_cases}). The executable validation in Table~\ref{tab:executable_validation} further shows that improving holistic plans through APB-guided refinement can translate into better downstream execution metrics in controlled environments.

\subsection{Efficiency-Aware Planning}
\begin{takeaway}
\textbf{\textit{Takeaway 3.}} Top-tier models exhibit emerging economic rationality, simultaneously optimizing solution correctness and execution cost.
\end{takeaway}

Operational efficiency is paramount for the scalable deployment of agents, a concern that has increasingly become the focus of recent research~\citep{DBLP:journals/corr/abs-2503-21614}.
To evaluate this capability, we curated a subset of 106 holistic planning instances where all five evaluated models originally achieved success. We then transformed these instances into a multi-solution environment by introducing alternative tools, assigning distinct execution costs to each. This setup shifts the objective from merely constructing a valid plan to identifying the optimal path that minimizes total cost while maintaining correctness.

As detailed in Table~\ref{tab:merged_plan_efficiency}, results indicate a divergence in capability. Gemini 3 Pro demonstrates superior competence, maintaining the highest correctness rate (>91\%) while achieving the lowest average cost. Conversely, Claude Sonnet 4.5 suffers significant performance degradation with markedly higher costs. These findings indicate that advanced models have evolved beyond merely identifying feasible solutions to exhibiting preliminary economic rationality, a critical step towards globally optimizing cost-effective agent systems.

\section{Conclusion}


We introduce APB, a planning-specific diagnostic benchmark for LLM agents. APB systematically evaluates holistic planning, feedback-conditioned step-wise planning, and robustness-oriented planning across 4,209 multimodal cases. Evaluations on 12 MLLMs confirm proprietary model superiority, yet reveal holistic planning's marked decline under extraneous tools. Regarding unsolvable scenarios, evaluations reveal that while top-tier models effectively detect explicit contradictions, all models remain vulnerable to implicit information gaps. We observe performance degradation in extended horizons and short-term reflection, alongside emerging economic rationality in proprietary models optimizing execution costs. We further validate APB's practical relevance through executable experiments on ToolSandbox and \tautwobench{}. Together, these findings position APB as a foundational diagnostic testbed for diagnosing planning failures and advancing logically robust agentic systems.


\section*{Limitations}
Despite our efforts to build a rigorous and scalable benchmark, we acknowledge that our work still has several limitations, summarized below:

\begin{itemize}[leftmargin=*, label=\small\textbullet, itemsep=2.5pt, parsep=0pt, topsep=0pt]
    \item Although APB includes 4,209 multimodal test cases across 22 domains, it cannot exhaust the open-ended diversity of real-world agent tasks, and may under-represent highly specialized vertical areas (e.g., biomedicine or enterprise software).

    \item APB focuses on the planning component of agent behavior. Although our step-wise tasks include trajectory feedback and our executable validation demonstrates downstream utility in two controlled environments, APB should be used as a diagnostic complement to full end-to-end deployment benchmarks rather than as a replacement for them.

    \item APB is not a symbolic PDDL benchmark or a probabilistic MDP/POMDP benchmark. It evaluates plan synthesis from natural-language or multimodal inputs and fixed tool specifications, and therefore does not model stochastic transitions, belief-state updates, or action-outcome likelihoods.

    \item Parts of our pipeline rely on proprietary foundation models for judging and data synthesis. Although our calibration identifies Qwen3.5-397B-A17B as a strong self-hosted judge, deploying such a large model still requires substantial local compute, and the primary proprietary evaluator remains subject to provider-side updates.
\end{itemize}

\section*{Ethical considerations}
\paragraph{Potential Risks}
This work proposes a benchmark for evaluating planning capabilities of LLM-based agents. A potential risk is that the benchmark may be over-optimized as a leaderboard target, leading to limited generalization beyond the benchmark. In addition, automatic evaluation using LLM judges may introduce biases or occasional misjudgments.

\paragraph{Personally Identifying Information and Offensive Content}
The dataset used in this work consists of synthetic tasks and environments, as well as annotations generated or curated by the authors and large language models. In addition, the benchmark incorporates selected data from existing datasets and a set of collected interaction records. As these sources may potentially contain personally identifying information, we conducted careful filtering and manual inspection to remove any information that could name or uniquely identify individuals, as well as offensive or inappropriate content.

\paragraph{Instructions Given to Participants}
This work did not involve external human participants, crowdworkers, or recruited annotators. All manual annotations were performed by the authors. Therefore, no participant instructions, consent forms, or risk disclosures were required.

\bibliography{new}

\begin{thebibliography}{47}
\providecommand{\natexlab}[1]{#1}

\bibitem[{Anthropic(2025)}]{claude-sonnet-4.5}
Anthropic. 2025.
\newblock Introducing claude sonnet 4.5.
\newblock \url{https://www.anthropic.com/news/claude-sonnet-4-5}.

\bibitem[{Barres et~al.(2025)Barres, Dong, Ray, Si, and Narasimhan}]{barres2025tau}
Victor Barres, Honghua Dong, Soham Ray, Xujie Si, and Karthik Narasimhan. 2025.
\newblock Tau2-bench: Evaluating conversational agents in a dual-control environment.
\newblock \emph{arXiv preprint arXiv:2506.07982}.

\bibitem[{Bonatti et~al.(2025)Bonatti, Zhao, Bonacci, Dupont, Abdali, Li, Lu, Wagle, Koishida, Bucker, Jang, and Hui}]{DBLP:conf/icml/BonattiZBDALLWK25}
Rogerio Bonatti, Dan Zhao, Francesco Bonacci, Dillon Dupont, Sara Abdali, Yinheng Li, Yadong Lu, Justin Wagle, Kazuhito Koishida, Arthur Bucker, Lawrence~Keunho Jang, and Zheng Hui. 2025.
\newblock \href {https://proceedings.mlr.press/v267/bonatti25a.html} {Windows agent arena: Evaluating multi-modal {OS} agents at scale}.
\newblock In \emph{Forty-second International Conference on Machine Learning, {ICML} 2025, Vancouver, BC, Canada, July 13-19, 2025}, volume 267 of \emph{Proceedings of Machine Learning Research}. {PMLR} / OpenReview.net.

\bibitem[{Chang et~al.(2024)Chang, Zhang, Zhu, Yang, Yang, Jin, Lan, Kong, and He}]{chang2024agentboard}
Ma~Chang, Junlei Zhang, Zhihao Zhu, Cheng Yang, Yujiu Yang, Yaohui Jin, Zhenzhong Lan, Lingpeng Kong, and Junxian He. 2024.
\newblock Agentboard: An analytical evaluation board of multi-turn llm agents.
\newblock \emph{Advances in neural information processing systems}, 37:74325--74362.

\bibitem[{Dagan et~al.(2024)Dagan, Keller, and Lascarides}]{dagan2024plancraft}
Gautier Dagan, Frank Keller, and Alex Lascarides. 2024.
\newblock Plancraft: an evaluation dataset for planning with llm agents.
\newblock \emph{arXiv preprint arXiv:2412.21033}.

\bibitem[{DeepMind(2025)}]{gemini-3-pro}
Google DeepMind. 2025.
\newblock Gemini 3 pro best for complex tasks and bringing creative concepts to life.
\newblock \url{https://deepmind.google/models/gemini/pro/}.

\bibitem[{Deng et~al.(2024)Deng, Xu, Sun, Liu, Tan, Liu, Li, Luan, Wang, Yan, and Shang}]{DBLP:conf/acl/DengXSLT0L000S24}
Shihan Deng, Weikai Xu, Hongda Sun, Wei Liu, Tao Tan, Jianfeng Liu, Ang Li, Jian Luan, Bin Wang, Rui Yan, and Shuo Shang. 2024.
\newblock \href {https://doi.org/10.18653/V1/2024.ACL-LONG.478} {Mobile-bench: An evaluation benchmark for llm-based mobile agents}.
\newblock In \emph{Proceedings of the 62nd Annual Meeting of the Association for Computational Linguistics (Volume 1: Long Papers), {ACL} 2024, Bangkok, Thailand, August 11-16, 2024}, pages 8813--8831. Association for Computational Linguistics.

\bibitem[{Deng et~al.(2023)Deng, Gu, Zheng, Chen, Stevens, Wang, Sun, and Su}]{deng2023mind2web}
Xiang Deng, Yu~Gu, Boyuan Zheng, Shijie Chen, Sam Stevens, Boshi Wang, Huan Sun, and Yu~Su. 2023.
\newblock Mind2web: Towards a generalist agent for the web.
\newblock \emph{Advances in Neural Information Processing Systems}, 36:28091--28114.

\bibitem[{Durante et~al.(2024)Durante, Huang, Wake, Gong, Park, Sarkar, Taori, Noda, Terzopoulos, Choi, Ikeuchi, Vo, Fei{-}Fei, and Gao}]{DBLP:journals/corr/abs-2401-03568}
Zane Durante, Qiuyuan Huang, Naoki Wake, Ran Gong, Jae~Sung Park, Bidipta Sarkar, Rohan Taori, Yusuke Noda, Demetri Terzopoulos, Yejin Choi, Katsushi Ikeuchi, Hoi Vo, Li~Fei{-}Fei, and Jianfeng Gao. 2024.
\newblock \href {https://doi.org/10.48550/ARXIV.2401.03568} {Agent {AI:} surveying the horizons of multimodal interaction}.
\newblock \emph{CoRR}, abs/2401.03568.

\bibitem[{Gao et~al.(2025)Gao, Tian, Meng, Wang, Hu, Xiao, Liu, Zhang, Chen, Gao, Lin, Xiong, Peng, and Liu}]{DBLP:journals/corr/abs-2507-23370}
Pengfei Gao, Zhao Tian, Xiangxin Meng, Xinchen Wang, Ruida Hu, Yuanan Xiao, Yizhou Liu, Zhao Zhang, Junjie Chen, Cuiyun Gao, Yun Lin, Yingfei Xiong, Chao Peng, and Xia Liu. 2025.
\newblock \href {https://doi.org/10.48550/ARXIV.2507.23370} {Trae agent: An llm-based agent for software engineering with test-time scaling}.
\newblock \emph{CoRR}, abs/2507.23370.

\bibitem[{He et~al.(2025)He, Qu, Li, Huang, Liu, and Cheng}]{he2025framethinker}
Zefeng He, Xiaoye Qu, Yafu Li, Siyuan Huang, Daizong Liu, and Yu~Cheng. 2025.
\newblock Framethinker: Learning to think with long videos via multi-turn frame spotlighting.
\newblock \emph{arXiv preprint arXiv:2509.24304}.

\bibitem[{Hong et~al.(2024)Hong, Zhuge, Chen, Zheng, Cheng, Wang, Zhang, Wang, Yau, Lin, Zhou, Ran, Xiao, Wu, and Schmidhuber}]{DBLP:conf/iclr/HongZCZCWZWYLZR24}
Sirui Hong, Mingchen Zhuge, Jonathan Chen, Xiawu Zheng, Yuheng Cheng, Jinlin Wang, Ceyao Zhang, Zili Wang, Steven Ka~Shing Yau, Zijuan Lin, Liyang Zhou, Chenyu Ran, Lingfeng Xiao, Chenglin Wu, and J{\"{u}}rgen Schmidhuber. 2024.
\newblock \href {https://openreview.net/forum?id=VtmBAGCN7o} {Metagpt: Meta programming for {A} multi-agent collaborative framework}.
\newblock In \emph{The Twelfth International Conference on Learning Representations, {ICLR} 2024, Vienna, Austria, May 7-11, 2024}. OpenReview.net.

\bibitem[{Huang et~al.(2024)Huang, Shi, Li, Fan, Wu, Zhang, Liu, Zhou, Wan, Gong, and Sun}]{DBLP:conf/iclr/HuangSLFWZ000G024}
Yue Huang, Jiawen Shi, Yuan Li, Chenrui Fan, Siyuan Wu, Qihui Zhang, Yixin Liu, Pan Zhou, Yao Wan, Neil~Zhenqiang Gong, and Lichao Sun. 2024.
\newblock \href {https://openreview.net/forum?id=R0c2qtalgG} {Metatool benchmark for large language models: Deciding whether to use tools and which to use}.
\newblock In \emph{The Twelfth International Conference on Learning Representations, {ICLR} 2024, Vienna, Austria, May 7-11, 2024}. OpenReview.net.

\bibitem[{Kim et~al.(2024)Kim, Moon, Tabrizi, Lee, Mahoney, Keutzer, and Gholami}]{kim2024llm}
Sehoon Kim, Suhong Moon, Ryan Tabrizi, Nicholas Lee, Michael~W Mahoney, Kurt Keutzer, and Amir Gholami. 2024.
\newblock An llm compiler for parallel function calling.
\newblock In \emph{Forty-first International Conference on Machine Learning}.

\bibitem[{Li et~al.(2025)Li, Chen, Zhang, and Liu}]{li2025planet}
Haoming Li, Zhaoliang Chen, Jonathan Zhang, and Fei Liu. 2025.
\newblock Planet: A collection of benchmarks for evaluating llms' planning capabilities.
\newblock \emph{arXiv preprint arXiv:2504.14773}.

\bibitem[{Li et~al.(2024)Li, Wen, Wang, Li, Yuan, Liu, Liu, Xu, Wang, Sun, Kong, Wang, Geng, Luan, Jin, Ye, Xiong, Zhang, Li, Xu, Li, Li, Liu, Zhang, and Liu}]{DBLP:journals/corr/abs-2401-05459}
Yuanchun Li, Hao Wen, Weijun Wang, Xiangyu Li, Yizhen Yuan, Guohong Liu, Jiacheng Liu, Wenxing Xu, Xiang Wang, Yi~Sun, Rui Kong, Yile Wang, Hanfei Geng, Jian Luan, Xuefeng Jin, Zilong Ye, Guanjing Xiong, Fan Zhang, Xiang Li, and 6 others. 2024.
\newblock \href {https://doi.org/10.48550/ARXIV.2401.05459} {Personal {LLM} agents: Insights and survey about the capability, efficiency and security}.
\newblock \emph{CoRR}, abs/2401.05459.

\bibitem[{Liu et~al.(2024)Liu, Yu, Zhang, Xu, Lei, Lai, Gu, Ding, Men, Yang, Zhang, Deng, Zeng, Du, Zhang, Shen, Zhang, Su, Sun, Huang, Dong, and Tang}]{DBLP:conf/iclr/0036YZXLL0DMYZ024}
Xiao Liu, Hao Yu, Hanchen Zhang, Yifan Xu, Xuanyu Lei, Hanyu Lai, Yu~Gu, Hangliang Ding, Kaiwen Men, Kejuan Yang, Shudan Zhang, Xiang Deng, Aohan Zeng, Zhengxiao Du, Chenhui Zhang, Sheng Shen, Tianjun Zhang, Yu~Su, Huan Sun, and 3 others. 2024.
\newblock \href {https://openreview.net/forum?id=zAdUB0aCTQ} {Agentbench: Evaluating llms as agents}.
\newblock In \emph{The Twelfth International Conference on Learning Representations, {ICLR} 2024, Vienna, Austria, May 7-11, 2024}. OpenReview.net.

\bibitem[{Liu et~al.(2025)Liu, Zhang, Gu, Iong, Song, Xu, Zhang, Lai, Sun, Yang, Yang, Qi, Yao, Sun, Cheng, Zheng, Yu, Zhang, Hong, Ding, Pan, Gu, Zeng, Du, Song, Su, Dong, and Tang}]{DBLP:conf/iclr/LiuZ0ISXZLSYYQY25}
Xiao Liu, Tianjie Zhang, Yu~Gu, Iat~Long Iong, Xixuan Song, Yifan Xu, Shudan Zhang, Hanyu Lai, Jiadai Sun, Xinyue Yang, Yu~Yang, Zehan Qi, Shuntian Yao, Xueqiao Sun, Siyi Cheng, Qinkai Zheng, Hao Yu, Hanchen Zhang, Wenyi Hong, and 9 others. 2025.
\newblock \href {https://openreview.net/forum?id=2snKOc7TVp} {Visualagentbench: Towards large multimodal models as visual foundation agents}.
\newblock In \emph{The Thirteenth International Conference on Learning Representations, {ICLR} 2025, Singapore, April 24-28, 2025}. OpenReview.net.

\bibitem[{Lu et~al.(2025)Lu, Holleis, Zhang, Aumayer, Nan, Bai, Ma, Ma, Li, Yin, Wang, and Pang}]{DBLP:conf/naacl/LuHZANBMMLYWP25}
Jiarui Lu, Thomas Holleis, Yizhe Zhang, Bernhard Aumayer, Feng Nan, Haoping Bai, Shuang Ma, Shen Ma, Mengyu Li, Guoli Yin, Zirui Wang, and Ruoming Pang. 2025.
\newblock \href {https://doi.org/10.18653/V1/2025.FINDINGS-NAACL.65} {Toolsandbox: {A} stateful, conversational, interactive evaluation benchmark for {LLM} tool use capabilities}.
\newblock In \emph{Findings of the Association for Computational Linguistics: {NAACL} 2025, Albuquerque, New Mexico, USA, April 29 - May 4, 2025}, volume {NAACL} 2025 of \emph{Findings of {ACL}}, pages 1160--1183. Association for Computational Linguistics.

\bibitem[{Luo et~al.(2025)Luo, Zhang, Yuan, Zhao, Yang, Gu, Wu, Chen, Qiao, Long, Tu, Luo, Ju, Xiao, Wang, Xiao, Liu, Yuan, Zhang, Jin, Zhang, Wu, Zhao, Tao, Yu, and Zhang}]{DBLP:journals/corr/abs-2503-21460}
Junyu Luo, Weizhi Zhang, Ye~Yuan, Yusheng Zhao, Junwei Yang, Yiyang Gu, Bohan Wu, Binqi Chen, Ziyue Qiao, Qingqing Long, Rongcheng Tu, Xiao Luo, Wei Ju, Zhiping Xiao, Yifan Wang, Meng Xiao, Chenwu Liu, Jingyang Yuan, Shichang Zhang, and 7 others. 2025.
\newblock \href {https://doi.org/10.48550/ARXIV.2503.21460} {Large language model agent: {A} survey on methodology, applications and challenges}.
\newblock \emph{CoRR}, abs/2503.21460.

\bibitem[{Mialon et~al.(2024)Mialon, Fourrier, Wolf, LeCun, and Scialom}]{mialon2024gaia}
Gr{\'e}goire Mialon, Cl{\'e}mentine Fourrier, Thomas Wolf, Yann LeCun, and Thomas Scialom. 2024.
\newblock Gaia: a benchmark for general ai assistants.
\newblock In \emph{International Conference on Learning Representations}, volume 2024, pages 9025--9049.

\bibitem[{OpenAI(2024)}]{4o}
OpenAI. 2024.
\newblock Hello gpt-4o.
\newblock \url{https://openai.com/index/hello-gpt-4o/}.

\bibitem[{OpenAI(2025)}]{gpt-5}
OpenAI. 2025.
\newblock Introducing gpt-5.
\newblock \url{https://openai.com/index/introducing-gpt-5/}.

\bibitem[{Qin et~al.(2024)Qin, Liang, Ye, Zhu, Yan, Lu, Lin, Cong, Tang, Qian, Zhao, Hong, Tian, Xie, Zhou, Gerstein, Li, Liu, and Sun}]{DBLP:conf/iclr/QinLYZYLLCTQZHT24}
Yujia Qin, Shihao Liang, Yining Ye, Kunlun Zhu, Lan Yan, Yaxi Lu, Yankai Lin, Xin Cong, Xiangru Tang, Bill Qian, Sihan Zhao, Lauren Hong, Runchu Tian, Ruobing Xie, Jie Zhou, Mark Gerstein, Dahai Li, Zhiyuan Liu, and Maosong Sun. 2024.
\newblock \href {https://openreview.net/forum?id=dHng2O0Jjr} {Toolllm: Facilitating large language models to master 16000+ real-world apis}.
\newblock In \emph{The Twelfth International Conference on Learning Representations, {ICLR} 2024, Vienna, Austria, May 7-11, 2024}. OpenReview.net.

\bibitem[{Qu et~al.(2025)Qu, Li, Su, Sun, Yan, Liu, Cui, Liu, Liang, He, Li, Wei, Shao, Lu, Zhang, Hua, Zhou, and Cheng}]{DBLP:journals/corr/abs-2503-21614}
Xiaoye Qu, Yafu Li, Zhaochen Su, Weigao Sun, Jianhao Yan, Dongrui Liu, Ganqu Cui, Daizong Liu, Shuxian Liang, Junxian He, Peng Li, Wei Wei, Jing Shao, Chaochao Lu, Yue Zhang, Xian{-}Sheng Hua, Bowen Zhou, and Yu~Cheng. 2025.
\newblock \href {https://doi.org/10.48550/ARXIV.2503.21614} {A survey of efficient reasoning for large reasoning models: Language, multimodality, and beyond}.
\newblock \emph{CoRR}, abs/2503.21614.

\bibitem[{{Qwen Team}(2025)}]{DBLP:journals/corr/abs-2511-21631}
{Qwen Team}. 2025.
\newblock \href {https://doi.org/10.48550/ARXIV.2511.21631} {Qwen3-vl technical report}.
\newblock \emph{CoRR}, abs/2511.21631.

\bibitem[{Shen et~al.(2024)Shen, Song, Tan, Zhang, Ren, Yuan, Lu, Li, and Zhuang}]{shen2024taskbench}
Yongliang Shen, Kaitao Song, Xu~Tan, Wenqi Zhang, Kan Ren, Siyu Yuan, Weiming Lu, Dongsheng Li, and Yueting Zhuang. 2024.
\newblock Taskbench: Benchmarking large language models for task automation.
\newblock \emph{Advances in Neural Information Processing Systems}, 37:4540--4574.

\bibitem[{Shinn et~al.(2023)Shinn, Cassano, Gopinath, Narasimhan, and Yao}]{shinn2023reflexion}
Noah Shinn, Federico Cassano, Ashwin Gopinath, Karthik Narasimhan, and Shunyu Yao. 2023.
\newblock Reflexion: Language agents with verbal reinforcement learning.
\newblock \emph{Advances in Neural Information Processing Systems}, 36:8634--8652.

\bibitem[{Shridhar et~al.(2020)Shridhar, Yuan, C{\^o}t{\'e}, Bisk, Trischler, and Hausknecht}]{shridhar2020alfworld}
Mohit Shridhar, Xingdi Yuan, Marc-Alexandre C{\^o}t{\'e}, Yonatan Bisk, Adam Trischler, and Matthew Hausknecht. 2020.
\newblock Alfworld: Aligning text and embodied environments for interactive learning.
\newblock \emph{arXiv preprint arXiv:2010.03768}.

\bibitem[{Si et~al.(2025)Si, Zhang, Li, Yang, Liu, and Yang}]{si2025design2code}
Chenglei Si, Yanzhe Zhang, Ryan Li, Zhengyuan Yang, Ruibo Liu, and Diyi Yang. 2025.
\newblock Design2code: Benchmarking multimodal code generation for automated front-end engineering.
\newblock In \emph{Proceedings of the 2025 Conference of the Nations of the Americas Chapter of the Association for Computational Linguistics: Human Language Technologies (Volume 1: Long Papers)}, pages 3956--3974.

\bibitem[{Team(2025)}]{DBLP:journals/corr/abs-2507-06261}
Gemini Team. 2025.
\newblock \href {https://doi.org/10.48550/ARXIV.2507.06261} {Gemini 2.5: Pushing the frontier with advanced reasoning, multimodality, long context, and next generation agentic capabilities}.
\newblock \emph{CoRR}, abs/2507.06261.

\bibitem[{Valmeekam et~al.(2023)Valmeekam, Marquez, Olmo, Sreedharan, and Kambhampati}]{valmeekam2023planbench}
Karthik Valmeekam, Matthew Marquez, Alberto Olmo, Sarath Sreedharan, and Subbarao Kambhampati. 2023.
\newblock Planbench: An extensible benchmark for evaluating large language models on planning and reasoning about change.
\newblock \emph{Advances in Neural Information Processing Systems}, 36:38975--38987.

\bibitem[{Wang et~al.(2023{\natexlab{a}})Wang, Xie, Jiang, Mandlekar, Xiao, Zhu, Fan, and Anandkumar}]{wang2023voyager}
Guanzhi Wang, Yuqi Xie, Yunfan Jiang, Ajay Mandlekar, Chaowei Xiao, Yuke Zhu, Linxi Fan, and Anima Anandkumar. 2023{\natexlab{a}}.
\newblock Voyager: An open-ended embodied agent with large language models.
\newblock \emph{arXiv preprint arXiv:2305.16291}.

\bibitem[{Wang et~al.(2024)Wang, Zerun, Li, Zhang, Chen, Chen, and Le}]{wang2024gta}
Jize Wang, Ma~Zerun, Yining Li, Songyang Zhang, Cailian Chen, Kai Chen, and Xinyi Le. 2024.
\newblock Gta: a benchmark for general tool agents.
\newblock \emph{Advances in Neural Information Processing Systems}, 37:75749--75790.

\bibitem[{Wang et~al.(2025{\natexlab{a}})Wang, Gao, Gu, Pu, Cui, Wei, Liu, Jing, Ye, Shao, Wang, Chen, Zhang, Yang, Wang, Wei, Yin, Li, Cui, Chen, Ding, Tian, Wu, Xie, Li, Yang, Duan, Wang, Hou, Hao, Zhang, Li, Zhao, Duan, Deng, Fu, He, Wang, He, Shi, He, Xiong, Lv, Wu, Shao, Zhang, Deng, Qi, Ge, Guo, Zhang, Zhang, Cao, Lin, Tang, Gao, Huang, Gu, Lyu, Tang, Wang, Lv, Ouyang, Wang, Dou, Zhu, Lu, Lin, Dai, Su, Zhou, Chen, Qiao, Wang, and Luo}]{DBLP:journals/corr/abs-2508-18265}
Weiyun Wang, Zhangwei Gao, Lixin Gu, Hengjun Pu, Long Cui, Xingguang Wei, Zhaoyang Liu, Linglin Jing, Shenglong Ye, Jie Shao, Zhaokai Wang, Zhe Chen, Hongjie Zhang, Ganlin Yang, Haomin Wang, Qi~Wei, Jinhui Yin, Wenhao Li, Erfei Cui, and 56 others. 2025{\natexlab{a}}.
\newblock \href {https://doi.org/10.48550/ARXIV.2508.18265} {Internvl3.5: Advancing open-source multimodal models in versatility, reasoning, and efficiency}.
\newblock \emph{CoRR}, abs/2508.18265.

\bibitem[{Wang et~al.(2023{\natexlab{b}})Wang, Hu, Lu, Zhu, Zhang, Subramaniam, Loomba, Zhang, Sun, and Wang}]{wang2023scibench}
Xiaoxuan Wang, Ziniu Hu, Pan Lu, Yanqiao Zhu, Jieyu Zhang, Satyen Subramaniam, Arjun~R Loomba, Shichang Zhang, Yizhou Sun, and Wei Wang. 2023{\natexlab{b}}.
\newblock Scibench: Evaluating college-level scientific problem-solving abilities of large language models.
\newblock \emph{arXiv preprint arXiv:2307.10635}.

\bibitem[{Wang et~al.(2025{\natexlab{b}})Wang, Wang, Lu, Yang, Xie, Wang, Deng, Guo, Xu, Wu, Shen, Li, Li, Li, Chen, Zheng, Li, Lei, Cao, Fu, Shin, Shin, Hu, Wang, Chen, Ye, Zhang, Wang, Wang, Yang, Zhong, Charles, Yang, and Yu}]{DBLP:conf/nips/WangWLYXWDGXWSL25}
Xinyuan Wang, Bowen Wang, Dunjie Lu, Junlin Yang, Tianbao Xie, Junli Wang, Jiaqi Deng, Xiaole Guo, Yiheng Xu, Chen~Henry Wu, Zhennan Shen, Zhuokai Li, Ryan Li, Xiaochuan Li, Junda Chen, Boyuan Zheng, Peihang Li, Fangyu Lei, Ruisheng Cao, and 15 others. 2025{\natexlab{b}}.
\newblock \href {http://papers.nips.cc/paper\_files/paper/2025/hash/cc7ae529e945226b0d52ea4ac478c4f3-Abstract-Conference.html} {Opencua: Open foundations for computer-use agents}.
\newblock In \emph{Advances in Neural Information Processing Systems 38: Annual Conference on Neural Information Processing Systems 2025, NeurIPS 2025, San Diego, CA, USA, December 2-7, 2025 / Mexico City, Mexico, November 30 - December 5, 2025}.

\bibitem[{Wei et~al.(2025)Wei, Zhang, He, Xia, Pan, and Liu}]{wei2025plangenllms}
Hui Wei, Zihao Zhang, Shenghua He, Tian Xia, Shijia Pan, and Fei Liu. 2025.
\newblock Plangenllms: A modern survey of llm planning capabilities.
\newblock \emph{arXiv preprint arXiv:2502.11221}.

\bibitem[{Xi et~al.(2025)Xi, Chen, Guo, He, Ding, Hong, Zhang, Wang, Jin, Zhou, Zheng, Fan, Wang, Xiong, Zhou, Wang, Jiang, Zou, Liu, Yin, Dou, Weng, Qin, Zheng, Qiu, Huang, Zhang, and Gui}]{DBLP:journals/chinaf/XiCGHDHZWJZZFWXZWJZLYDW25}
Zhiheng Xi, Wenxiang Chen, Xin Guo, Wei He, Yiwen Ding, Boyang Hong, Ming Zhang, Junzhe Wang, Senjie Jin, Enyu Zhou, Rui Zheng, Xiaoran Fan, Xiao Wang, Limao Xiong, Yuhao Zhou, Weiran Wang, Changhao Jiang, Yicheng Zou, Xiangyang Liu, and 9 others. 2025.
\newblock \href {https://doi.org/10.1007/S11432-024-4222-0} {The rise and potential of large language model based agents: a survey}.
\newblock \emph{Sci. China Inf. Sci.}, 68(2).

\bibitem[{Xie et~al.(2024{\natexlab{a}})Xie, Zhang, Chen, Zhu, Lou, Tian, Xiao, and Su}]{xie2024travelplanner}
Jian Xie, Kai Zhang, Jiangjie Chen, Tinghui Zhu, Renze Lou, Yuandong Tian, Yanghua Xiao, and Yu~Su. 2024{\natexlab{a}}.
\newblock Travelplanner: A benchmark for real-world planning with language agents.
\newblock \emph{arXiv preprint arXiv:2402.01622}.

\bibitem[{Xie et~al.(2024{\natexlab{b}})Xie, Zhang, Chen, Li, Zhao, Cao, Hua, Cheng, Shin, Lei, Liu, Xu, Zhou, Savarese, Xiong, Zhong, and Yu}]{DBLP:conf/nips/XieZCLZCHCSLLXZ24}
Tianbao Xie, Danyang Zhang, Jixuan Chen, Xiaochuan Li, Siheng Zhao, Ruisheng Cao, Toh~Jing Hua, Zhoujun Cheng, Dongchan Shin, Fangyu Lei, Yitao Liu, Yiheng Xu, Shuyan Zhou, Silvio Savarese, Caiming Xiong, Victor Zhong, and Tao Yu. 2024{\natexlab{b}}.
\newblock \href {http://papers.nips.cc/paper\_files/paper/2024/hash/5d413e48f84dc61244b6be550f1cd8f5-Abstract-Datasets\_and\_Benchmarks\_Track.html} {Osworld: Benchmarking multimodal agents for open-ended tasks in real computer environments}.
\newblock In \emph{Advances in Neural Information Processing Systems 37: Annual Conference on Neural Information Processing Systems 2024, NeurIPS 2024, Vancouver, BC, Canada, December 10 - 15, 2024}.

\bibitem[{Yang et~al.(2024)Yang, Jimenez, Wettig, Lieret, Yao, Narasimhan, and Press}]{yang2024sweagent}
John Yang, Carlos~E Jimenez, Alexander Wettig, Kilian Lieret, Shunyu Yao, Karthik Narasimhan, and Ofir Press. 2024.
\newblock Swe-agent: Agent-computer interfaces enable automated software engineering.
\newblock \emph{Advances in Neural Information Processing Systems}, 37:50528--50652.

\bibitem[{Yang et~al.(2025{\natexlab{a}})Yang, Jimenez, Zhang, Lieret, Yang, Wu, Press, Muennighoff, Synnaeve, Narasimhan, Yang, Wang, and Press}]{DBLP:conf/iclr/YangJZLYWPMSNY025}
John Yang, Carlos~E. Jimenez, Alex~L. Zhang, Kilian Lieret, Joyce Yang, Xindi Wu, Ori Press, Niklas Muennighoff, Gabriel Synnaeve, Karthik~R. Narasimhan, Diyi Yang, Sida Wang, and Ofir Press. 2025{\natexlab{a}}.
\newblock \href {https://openreview.net/forum?id=riTiq3i21b} {Swe-bench multimodal: Do {AI} systems generalize to visual software domains?}
\newblock In \emph{The Thirteenth International Conference on Learning Representations, {ICLR} 2025, Singapore, April 24-28, 2025}. OpenReview.net.

\bibitem[{Yang et~al.(2025{\natexlab{b}})Yang, Li, Dai, Yang, Luo, Zhao, Hu, Huang, Saha, Chen, Xu, Pan, Xiong, and Li}]{DBLP:journals/corr/abs-2507-05791}
Yan Yang, Dongxu Li, Yutong Dai, Yuhao Yang, Ziyang Luo, Zirui Zhao, Zhiyuan Hu, Junzhe Huang, Amrita Saha, Zeyuan Chen, Ran Xu, Liyuan Pan, Caiming Xiong, and Junnan Li. 2025{\natexlab{b}}.
\newblock \href {https://doi.org/10.48550/ARXIV.2507.05791} {{GTA1:} {GUI} test-time scaling agent}.
\newblock \emph{CoRR}, abs/2507.05791.

\bibitem[{Yao et~al.(2022)Yao, Zhao, Yu, Du, Shafran, Narasimhan, and Cao}]{yao2022react}
Shunyu Yao, Jeffrey Zhao, Dian Yu, Nan Du, Izhak Shafran, Karthik Narasimhan, and Yuan Cao. 2022.
\newblock React: Synergizing reasoning and acting in language models.
\newblock \emph{arXiv preprint arXiv:2210.03629}.

\bibitem[{Ye et~al.(2025)Ye, Li, Gao, Huang, Wu, Li, Fan, Dou, Ji, Zhang, Gui, and Huang}]{DBLP:conf/coling/YeLGHWLFDJ0G025}
Junjie Ye, Guanyu Li, Songyang Gao, Caishuang Huang, Yilong Wu, Sixian Li, Xiaoran Fan, Shihan Dou, Tao Ji, Qi~Zhang, Tao Gui, and Xuanjing Huang. 2025.
\newblock \href {https://aclanthology.org/2025.coling-main.12/} {Tooleyes: Fine-grained evaluation for tool learning capabilities of large language models in real-world scenarios}.
\newblock In \emph{Proceedings of the 31st International Conference on Computational Linguistics, {COLING} 2025, Abu Dhabi, UAE, January 19-24, 2025}, pages 156--187. Association for Computational Linguistics.

\bibitem[{Zhou et~al.(2024)Zhou, Xu, Zhu, Zhou, Lo, Sridhar, Cheng, Ou, Bisk, Fried, Alon, and Neubig}]{DBLP:conf/iclr/ZhouX0ZLSCOBF0N24}
Shuyan Zhou, Frank~F. Xu, Hao Zhu, Xuhui Zhou, Robert Lo, Abishek Sridhar, Xianyi Cheng, Tianyue Ou, Yonatan Bisk, Daniel Fried, Uri Alon, and Graham Neubig. 2024.
\newblock \href {https://openreview.net/forum?id=oKn9c6ytLx} {Webarena: {A} realistic web environment for building autonomous agents}.
\newblock In \emph{The Twelfth International Conference on Learning Representations, {ICLR} 2024, Vienna, Austria, May 7-11, 2024}. OpenReview.net.

\end{thebibliography}

\appendix

\section*{Appendix Overview}
This appendix presents supplementary materials organized into eight sections:

\begin{itemize}[leftmargin=*, label=\small\textbullet, itemsep=2.5pt, parsep=0pt, topsep=0pt]
    \item \textbf{Appendix~\ref{app:benchmark_limitation}: Comparison with Existing Benchmarks}. Compares APB with existing agent benchmarks across key dimensions.

    \item \textbf{Appendix~\ref{app:data_construction}: Data Construction and Filtering Process}. Details the data augmentation pipeline, multi-stage quality control (rule-based, LLM-based, human verification), and the construction of diverse planning scenarios including holistic, robustness (tool-broken, tool-extraneous, unsolvable), and efficiency-aware tasks.

    \item \textbf{Appendix~\ref{app:evaluation_framework}: Evaluation Framework}. Outlines the comprehensive evaluation methodology, covering step-wise planning (one to three steps) with standardized error definitions, holistic planning emphasizing solvability, robustness settings (tool-broken, tool-extraneous, unsolvable, refinement), and efficiency-aware planning evaluation.

    \item \textbf{Appendix~\ref{app:llm_judge_analysis}: LLM-as-a-Judge Analysis}. Validates the automated evaluation pipeline through rigorous analysis of judge model selection, robustness to solution diversity, and self-evaluation bias.

    \item \textbf{Appendix~\ref{app:dataset_composition}: Dataset Composition}. Describes the diverse dataset sources (FrameThinker, GAIA, GTA, OpenCUA, ToolBench, Real-World Interactions), presenting augmented query examples, step count statistics, and a detailed scenario taxonomy.

    \item \textbf{Appendix~\ref{app:model_response_examples}: Data Samples and Model Response Examples}. Showcases qualitative examples of model performance across all settings, including correct/incorrect predictions, failure recovery patterns, extraneous tool impact, refinement effects, and responses to unsolvable constraints.

    \item \textbf{Appendix~\ref{app:detailed_results}: Detailed Experimental Results}. Presents granular quantitative results, including performance breakdowns by step range, modality, and planning stage, along with in-depth analysis of tool-extraneous impact and error distributions.

    \item \textbf{Appendix~\ref{app:model_config}: Model Sources and Hyperparameters}. Summarizes the sources, access methods (API vs. local), and hyperparameter configurations for all evaluated models.
\end{itemize}

\begin{table*}[t]
\centering
\renewcommand{\arraystretch}{1.1}
\setlength{\tabcolsep}{3pt}
\resizebox{\textwidth}{!}{
\begin{tabular}{l ccccccc}
\toprule
\textbf{Benchmark} & \textbf{Exec-Free} & \textbf{Process-Aware} & \textbf{Error Diag.} & \textbf{Step-wise} & \textbf{Holistic} & \textbf{Adversarial} & \textbf{Multimodal} \\
\midrule
AgentBench \citep{DBLP:conf/iclr/0036YZXLL0DMYZ024} & \ding{55} & \ding{55} & \ding{55} & \ding{51} & \ding{55} & \ding{55} & \ding{55} \\
AgentBoard \citep{chang2024agentboard} & \ding{55} & \ding{51} & \ding{55} & \ding{51} & \ding{55} & \ding{55} & \ding{55} \\
TaskBench \citep{shen2024taskbench} & \ding{51} & \ding{51} & \ding{55} & \ding{55} & \ding{51} & \ding{55} & \ding{55} \\
ToolEyes \citep{DBLP:conf/coling/YeLGHWLFDJ0G025} & \ding{55} & \ding{51} & \ding{51} & \ding{51} & \ding{55} & \ding{55} & \ding{55} \\
MetaTool \citep{DBLP:conf/iclr/HuangSLFWZ000G024} & \ding{55} & \ding{55} & \ding{51} & \ding{51} & \ding{55} & \ding{55} & \ding{55} \\
GTA \citep{wang2024gta} & \ding{55} & \ding{55} & \ding{55} & \ding{51} & \ding{51} & \ding{55} & \ding{51} \\
PlanBench \citep{valmeekam2023planbench} & \ding{51} & \ding{55} & \ding{55} & \ding{55} & \ding{51} & \ding{55} & \ding{55} \\
PlanCraft \citep{dagan2024plancraft} & \ding{55} & \ding{51} & \ding{55} & \ding{51} & \ding{55} & \ding{51} & \ding{51} \\
TravelPlanner \citep{xie2024travelplanner} & \ding{55} & \ding{55} & \ding{55} & \ding{55} & \ding{51} & \ding{55} & \ding{55} \\
VisualAgentBench \citep{DBLP:conf/iclr/LiuZ0ISXZLSYYQY25} & \ding{55} & \ding{51} & \ding{55} & \ding{51} & \ding{55} & \ding{55} & \ding{51} \\
OSWorld \citep{DBLP:conf/nips/XieZCLZCHCSLLXZ24} & \ding{55} & \ding{55} & \ding{55} & \ding{51} & \ding{55} & \ding{55} & \ding{51} \\
WebArena \citep{DBLP:conf/iclr/ZhouX0ZLSCOBF0N24} & \ding{55} & \ding{55} & \ding{55} & \ding{51} & \ding{55} & \ding{55} & \ding{51} \\
Mind2Web \citep{deng2023mind2web} & \ding{51} & \ding{55} & \ding{55} & \ding{51} & \ding{55} & \ding{55} & \ding{51} \\
\midrule
\textbf{APB (Ours)} & \ding{51} & \ding{51} & \ding{51} & \ding{51} & \ding{51} & \ding{51} & \ding{51} \\
\bottomrule
\end{tabular}
}
\caption{\textbf{Comparison of APB with existing agent benchmarks across key evaluation dimensions.} \textbf{Exec-Free}: Decouples planning from execution environment to avoid noise interference. \textbf{Process-Aware}: Evaluates intermediate reasoning processes beyond final outcomes. \textbf{Error Diag.}: Provides fine-grained error taxonomy for failure analysis. \textbf{Step-wise}: Supports local, short-horizon planning evaluation. \textbf{Holistic}: Supports global, long-horizon trajectory generation. \textbf{Adversarial}: Includes robustness scenarios (tool-extraneous, tool-broken, unsolvable tasks). \textbf{Multimodal}: Supports text, image, and video modalities.}
\label{tab:benchmark_comparison}
\vspace{-10pt}
\end{table*}

\section{Comparison of APB with existing agent benchmarks}
\label{app:benchmark_limitation}
This section compares APB with existing agent benchmarks. Table~\ref{tab:benchmark_comparison} summarizes four primary categories of benchmarks contrasted with APB:

\begin{itemize}[leftmargin=*, label=\small\textbullet, itemsep=2.5pt, parsep=0pt, topsep=0pt]
    \item \textbf{Execution-dependent benchmarks}~\citep{DBLP:conf/iclr/0036YZXLL0DMYZ024, DBLP:conf/coling/YeLGHWLFDJ0G025, DBLP:conf/iclr/HuangSLFWZ000G024} couple planning with execution, requiring interaction with real environments (e.g., web browsers, APIs). This introduces environmental stochasticity (e.g., network latency, API failures), masking true planning capabilities. In contrast, APB decouples planning from execution to assess pure planning logic without environmental interference.

    \item \textbf{Outcome-oriented benchmarks}~\citep{wang2024gta, xie2024travelplanner, DBLP:conf/nips/XieZCLZCHCSLLXZ24} evaluate performance primarily via final success rates, treating planning as a black box. This neglects intermediate reasoning and failure analysis. APB instead adopts a process-aware diagnostic paradigm, evaluating each planning step for fine-grained error attribution.

    \item \textbf{Holistic-only benchmarks}~\citep{valmeekam2023planbench, xie2024travelplanner, shen2024taskbench} require agents to generate complete plans in a single pass, assessing only global plan quality through structural metrics (e.g., graph similarity). While useful for end-to-end evaluation, these benchmarks lack step-wise assessment capabilities and fine-grained error analysis mechanisms. APB integrates both holistic and step-wise tasks, enabling multi-granularity evaluation with a comprehensive error taxonomy.

    \item \textbf{Reference-matching benchmarks}~\citep{valmeekam2023planbench, shen2024taskbench, deng2023mind2web} rely on exact or fuzzy matching against ground truth, assuming a single correct solution. APB employs a reference-aware LLM-as-Judge system to evaluate semantic correctness, recognizing diverse valid solutions.
\end{itemize}

\section{Data Construction and Filtering Process}
\label{app:data_construction}
\subsection{Data Augmentation Pipeline}
\label{app:data_augmentation}
    To evaluate agent planning in complex scenarios, we extend GAIA \citep{mialon2024gaia}, GTA \citep{wang2024gta}, ToolBench \citep{DBLP:conf/iclr/QinLYZYLLCTQZHT24}, and FrameThinker \citep{he2025framethinker} to increase query complexity, while retaining OpenCUA \citep{DBLP:conf/nips/WangWLYXWDGXWSL25} and real online traffic data as they are sufficiently representative of real-world complexity.
    We employ Claude Sonnet~4.5 to generate new queries $Q$ from the original seed queries $Q_{\text{orig}}$, expand the corresponding tool sets $\mathcal{T}_{\text{orig}}$ to $\mathcal{T}$, and simulate the resulting execution trajectories $\tau$. Post-augmentation, these datasets average 12.3 reasoning steps (an increase of 4.5) and 9.4 tools (an increase of 5.6) per query, significantly enhancing the difficulty and depth of the planning tasks.

    Claude is therefore used for controlled augmentation rather than de novo benchmark generation. All six sources remain grounded in benchmark trajectories or real interaction records and subsequently pass the same validation pipeline. At evaluation time, records are normalized into the common APB schema. Tested models receive only task-facing fields---the system prompt, query, available tools, multimodal context, and, for step-wise tasks, a trajectory prefix. Generator identity, synthesis traces, and reference plans or continuations remain hidden. This separation prevents direct in-context copying of the generated reference and distinguishes source data, model-assisted augmentation, model-facing inputs, and evaluator-only supervision.

    These controls do not establish the absence of pretraining contamination in proprietary models, which cannot be directly audited. We therefore restrict the evidence to observable checks. Claude Sonnet~4.5, the augmentation model, does not lead either core APB task: it obtains 64.1\% Holistic CR and 77.3\% Step-wise CR, compared with the respective best results of 74.5\% and 80.1\% in Table~\ref{tab:main_table_results}. In addition, APB-guided refinement improves native execution metrics in the independently constructed ToolSandbox and \tautwobench{} environments (Table~\ref{tab:executable_validation}), providing an external transfer check beyond APB's construction pipeline.

\begin{figure*}[htbp]
    \centering
\begin{promptbox}{Data Augmentation Prompt}

You are an expert AI data augmentation and reasoning-chain generation specialist.
Your mission is to take an existing simple dialogue (the seed query) and perform two subtasks: \\

\textbf{TASK 1: Query Expansion \& Toolset Enrichment} \\
Transform the given simple user query into a \textbf{more complex, realistic, and analytically challenging query} — one that \textbf{naturally requires the introduction of new tools} to be solvable. \\
• The new query must: \\
\hspace*{1em} • Require multiple reasoning steps and multiple tool calls to complete. \\
\hspace*{1em} • Stay coherent, natural, and goal-driven (avoid artificial or random complexity). \\
\hspace*{1em} • Introduce interdependent subtasks when appropriate (e.g., data lookup $\rightarrow$ processing $\rightarrow$ comparison $\rightarrow$ conclusion). \\
\hspace*{1em} • Be sufficiently rich that solving it \textbf{demands capabilities beyond the current toolset}, thereby motivating the creation of new tools. \\
• You may invent \textbf{new tools} to support the expanded task, but: \\
\hspace*{1em} • Each invented tool must be \textbf{essential} for completing the new query (not optional or redundant). \\
\hspace*{1em} • Each invented tool must have a \textbf{clear purpose}, \textbf{realistic input/output}, and be \textbf{plausible} in an AI assistant environment. \\
\hspace*{1em} • New tools should fill well-defined capability gaps (e.g., new reasoning, analysis, or synthesis functions) that existing tools cannot address. \\
\hspace*{1em} • Avoid arbitrary or trivial tool creation; tools must emerge naturally from task needs. \\

\textbf{TASK 2: Trajectory Generation and Tool Documentation} \\
Using the new complex query and enriched toolset: \\
1. Simulate a complete reasoning and execution trajectory that solves the query. \\
2. Ensure each assistant step contains concise reasoning ("thought") and EXACTLY ONE tool call. \\
3. Include realistic tool outputs as tool steps. \\
4. End with a final assistant step calling the Finish function (or providing the final answer). \\
5. Collect all tools used (including pre-existing and newly created ones) and output them with complete, generic descriptions. \\

\textbf{CRITICAL: Tool Calling Constraints} \\
• Each assistant turn MUST call EXACTLY ONE tool (not zero, not multiple). \\
• ALL tool calls MUST succeed and return valid results — no failures, errors, or exceptions. \\
• Tool outputs must be realistic, specific, and directly usable by subsequent steps. \\
• Tools should be designed to be reliable and deterministic, producing successful results every time. \\

\textbf{CRITICAL: Tool Description Requirements} \\
• Tool descriptions must read like standalone documentation: clearly state the tool’s purpose, enumerate every input parameter with name/type/defaults/constraints, outline the returned structure, and include usage notes or best practices. \\
• Keep descriptions task-agnostic but specific enough that another engineer could confidently reuse the tool in unrelated contexts; avoid referencing the current task and never leak problem-specific data. \\
• Follow a polished textual style (e.g., "Does [X]. Parameters: 'param1' (string, required) - … Returns … Notes: …") so each entry matches standard documentation. \\

\textbf{AVAILABLE TOOLS (you can use these or invent new ones):} \\
\{List of Existing Tools\} \\

\textbf{ORIGINAL TASK CONTEXT:} \\
\{Original Question/Dialogue and Metadata\} \\

\textbf{OUTPUT FORMAT (CRITICAL)} \\
You must output ONLY a valid single JSON object with this structure:
\begin{lstlisting}[language=java, showstringspaces=false, breaklines=true, basicstyle=\ttfamily\scriptsize, columns=fullflexible, frame=none, aboveskip=0pt, belowskip=0pt]
{
    "trajectory": [
        {"role": "user", "content": "expanded complex query"},
        {"role": "assistant", "thought": "brief reasoning", "tool_calls": [{"type": "function", "function": {"name": "function_name", "arguments": {}}}]},
        {"role": "tool", "name": "function_name", "content": "successful result with specific data"},
        ...
        {"role": "assistant", "tool_calls": [{"type": "function", "function": {"name": "Finish", "arguments": {"mode": "give_answer"}}}]}
    ],
    "tools": [
        {"tool_name": "ExistingTool1", "description": "..."},
        {"tool_name": "InventedToolA", "description": "..."}
    ]
}
\end{lstlisting}

\textbf{REQUIREMENTS} \\
• The output must be a single valid JSON object. \\
• Every assistant turn MUST have exactly ONE tool call. \\
• Every tool call MUST be followed by a successful tool response. \\
• Tool responses must contain realistic, specific, and usable data. \\
• Every tool used must be documented in the "tools" list. \\
• The process is irreversible — you cannot revise earlier steps. \\
• Target a specific number of reasoning steps (e.g., significantly more than the original). \\
• The new query must remain in the same domain but achieve higher complexity. \\
• The model should reason step-by-step WITHOUT knowing the final answer in advance.

    \end{promptbox}
    \caption{Data augmentation prompt for transforming simple queries into complex, multi-step planning tasks with diverse tool requirements.}
    \label{fig:data_augmentation_prompt}
\end{figure*}

\subsection{Quality Control Pipeline}
\label{app:quality_control}
We implement a multi-stage quality control pipeline combining rule-based validation, model-based review, and human verification to ensure ground truth reliability.

\begin{itemize}[leftmargin=*, label=\small\textbullet, itemsep=2.5pt, parsep=0pt, topsep=0pt]
    \item \textbf{Rule-based Structural Validation}
    Automated checks filter out non-existent tools and format violations (Figure \ref{fig:rule_based_verification}).

    \item \textbf{LLM Task Reasonability Check}
    We verify query realism and tool validity, ensuring tools are not overpowered or structurally inconsistent (Figure \ref{fig:tool_reasonability_prompt}).

    \item \textbf{LLM Trajectory Validation}
    Trajectories are evaluated using the \textbf{E1-E6 Error Taxonomy}. An LLM identifies errors in the query, tool list, and trajectory (Figure \ref{fig:logic_validation_prompt}).

    \item \textbf{Human Verification}
    Finally, a three-author annotation team verifies consistency between objectives and logic, ensuring automated evaluation missed no flaws. Every one of the 1,109 final holistic base instances---including all 370 real-world instances---is independently reviewed by two members of this team. Each reviewer checks task solvability; tool definitions, schemas, arguments, and calls; consistency between tool returns and downstream dependencies; trajectory completeness and E1--E6 validity; multimodal grounding where applicable; and privacy and safety.
    \end{itemize}

Disagreements between the two reviewers are adjudicated by the third member. When an instance requires manual correction, an author who did not perform that correction revalidates the complete corrected instance. Errors trigger iterative correction, and only data passing the automated checks, dual-author review, any necessary adjudication, and post-correction revalidation are retained as $\tau_{\mathrm{valid}}$. For real-world data, we additionally remove personally identifying information (PII) and sensitive content.

\begin{figure}[htbp]
    \centering
\begin{lstlisting}[language=Python, showstringspaces=false, basicstyle=\ttfamily\scriptsize, frame=single, breaklines=true]
def rule_based_verification(sample):
    """
    Stage 1: Structural and Logical Integrity Check
    """
    # 1. Format Validation
    required_fields = ["query", "trajectory", "tools"]
    for field in required_fields:
        if field not in sample:
            return False, f"Missing required field: {field}"

    # 2. Tool Existence Validation
    # Extract all defined tool names from the toolset
    available_tools = extract_tool_names(sample["tools"])

    # Extract all tool calls from the execution trajectory
    tool_calls = extract_tool_calls(sample["trajectory"])

    # Verify that every called tool exists in the definition
    for tool_name in tool_calls:
        if tool_name not in available_tools:
            return False, f"Hallucinated tool call detected: {tool_name}"

    return True, "Passed"
\end{lstlisting}
    \caption{Pseudo-code for the rule-based verification stage, enforcing structural integrity and tool consistency in augmented data.}
    \label{fig:rule_based_verification}
\end{figure}

\begin{figure*}[htbp]
    \centering
\begin{promptbox}{Task Reasonability Check Prompt}

\textbf{Part 1: Query Reasonability Check}

You are evaluating the reasonability of a task query for an AI agent benchmark.

\textbf{IMPORTANT CONTEXT:} \\
The AI agent being tested is a \textbf{multimodal agent} with full visual capabilities. It can: \\
• View and analyze images (photos, screenshots, diagrams, charts) \\
• Process video content \\
• Perform OCR (optical character recognition) on images \\
• Count objects in images \\
• Describe visual content \\
• Read text from product listings, menus, documents, etc.

Therefore, queries that involve visual tasks (analyzing photos, reading menus, counting objects in images, etc.) are all VALID and REASONABLE.

\textbf{Task Query:} \\
\{query\}

\textbf{Evaluation Criteria:} \\
1. \textbf{Realistic}: The query should represent a realistic task that a user might ask a multimodal AI agent \\
2. \textbf{Clear}: The query should be clear and understandable \\
3. \textbf{Achievable}: The task should be achievable with reasonable tools (search, calculator, file operations, image analysis, OCR, etc.) \\
4. \textbf{Not Overly Complex}: The query should not require superhuman capabilities

\textbf{NOTE}: Visual/image-related queries are completely valid since the agent has full visual capabilities.

\textbf{Output Format (JSON only):}
\begin{lstlisting}[language=java, showstringspaces=false, breaklines=true, basicstyle=\ttfamily\scriptsize, columns=fullflexible, frame=none, aboveskip=0pt, belowskip=0pt]
{
    "is_reasonable": true/false,
    "reason": "Brief explanation if not reasonable"
}
\end{lstlisting}

Evaluate and return your assessment.

\vspace{1em}
\hrule
\vspace{1em}

\textbf{Part 2: Tool Reasonability Check}

You are evaluating the reasonability of AI-generated tools for a benchmark dataset.

\textbf{IMPORTANT CONTEXT:} \\
The AI agent being tested is a \textbf{multimodal agent} with full visual capabilities. It can: \\
• View and analyze images (photos, screenshots, diagrams, charts) \\
• Process video content \\
• Perform OCR (optical character recognition) on images \\
• Count objects in images \\
• Describe visual content \\
• Read text from product listings, menus, documents, etc.

Therefore, tools related to image processing, visual analysis, OCR, object detection, etc. are all VALID and REASONABLE.

\textbf{Task Query:} \\
\{query\}

\textbf{Generated Tools to Evaluate:} \\
\{tools\_json\}

\textbf{Evaluation Criteria:} \\
1. \textbf{Realistic}: The tool should represent a realistic API/function that could exist in the real world \\
2. \textbf{Not Overpowered}: The tool should NOT be overpowered (e.g., "solve\_any\_problem", "get\_perfect\_answer") \\
3. \textbf{Appropriate Scope}: The tool's functionality should be appropriately scoped (not too broad, not too narrow) \\
4. \textbf{Consistent Format}: The tool definition should have proper structure (name, description, parameters)

\textbf{NOTE}: Image/vision-related tools (ImageDescription, OCR, CountGivenObject, etc.) are completely valid since the agent has visual capabilities.

\textbf{Output Format (JSON only):}
\begin{lstlisting}[language=java, showstringspaces=false, breaklines=true, basicstyle=\ttfamily\scriptsize, columns=fullflexible, frame=none, aboveskip=0pt, belowskip=0pt]
{
    "is_reasonable": true/false,
    "reason": "Brief explanation if not reasonable"
}
\end{lstlisting}

Evaluate ALL tools and return your assessment.
    \end{promptbox}
    \caption{Task reasonability check prompt, validating both query realism and tool appropriateness for synthesized tasks.}
    \label{fig:tool_reasonability_prompt}
\end{figure*}

\begin{figure*}[htbp]
    \centering
\begin{promptbox}{Trajectory Logic Validation Prompt}

You are evaluating the quality of an AI agent's trajectory for completing a task.

\textbf{IMPORTANT CONTEXT:} \\
The AI agent being tested is a \textbf{multimodal agent} with full visual capabilities. \\
\textit{[...details on multimodal capabilities...]} \\

Therefore, tool calls related to image processing, visual analysis, OCR, object detection, etc. are all VALID and should NOT be flagged as errors.

\textbf{Task Query:} \\
\{query\}

\textbf{Available Tools (first 10):} \\
\{tools\_display\}

\textbf{Agent Trajectory:} \\
\{trajectory\_display\}

\textbf{Error Classification Framework:} \\
Evaluate the ENTIRE trajectory for the following error types: \\

1. \textbf{E1\_GOAL\_MISALIGNMENT}: Does any step deviate from or contradict the task goal? \\
2. \textbf{E2\_PREMATURE\_CONCLUSION}: Does the trajectory conclude before completing all required sub-tasks? \\
3. \textbf{E3\_CONSTRAINT\_VIOLATION}: Does any step violate explicit constraints from the query or format requirements? \\
4. \textbf{E4\_LOGIC\_ERROR}: Are there logical flaws, missing prerequisites, or illogical reasoning flows? \\
5. \textbf{E5\_TOOL\_USE\_ERROR}: Are tools misused (wrong parameters, wrong data types, missing required parameters)? \\
6. \textbf{E6\_HALLUCINATION}: Are there calls to non-existent tools or references to non-existent data? \\

\textbf{IMPORTANT NOTES:} \\
• Image/vision-related tools (ImageDescription, OCR, CountGivenObject, etc.) are completely valid since the agent has visual capabilities. \\
• The agent CAN see and process images, so using image analysis tools is NOT an error. \\
• Do NOT flag visual tool usage as hallucination or tool use error. \\

\textbf{Output Format (JSON only):}
\begin{lstlisting}[language=java, showstringspaces=false, breaklines=true, basicstyle=\ttfamily\scriptsize, columns=fullflexible, frame=none, aboveskip=0pt, belowskip=0pt]
{
    "is_valid": true/false,
    "errors": {
        "E1_GOAL_MISALIGNMENT": false,
        "E2_PREMATURE_CONCLUSION": false,
        "E3_CONSTRAINT_VIOLATION": false,
        "E4_LOGIC_ERROR": false,
        "E5_TOOL_USE_ERROR": false,
        "E6_HALLUCINATION": false
    },
    "error_details": "Detailed explanation of any errors found, or 'No errors' if valid"
}
\end{lstlisting}

Evaluate the trajectory and return your assessment:

    \end{promptbox}
    \caption{Trajectory logic validation prompt with E1 to E6 error taxonomy for auditing coherence and validity of agent execution traces.}
    \label{fig:logic_validation_prompt}
\end{figure*}

\subsection{Step-wise to Holistic Transformation}
We convert validated step-by-step trajectories into a holistic planning format to support diverse evaluation paradigms. This mitigates incompleteness from gradual information revelation in dialogues.

Using Claude Sonnet~4.5, we extract implicit information, rewriting $Q_{\text{orig}}$ to $Q_{\text{refined}}$ and reconstructing the plan $P_{\text{gt}}$ and tool chain $\mathcal{C}_{\text{gt}}$. As detailed in the \textbf{Step-wise to Holistic Extraction Prompt} (Figure \ref{fig:offline_extraction_prompt}), this transformation prioritizes strategic reasoning. The process is formalized as:
\begin{equation}
    (Q_{\text{refined}}, P_{\text{gt}}, \mathcal{C}_{\text{gt}}) = f_{\text{transform}}(Q_{\text{orig}}, \tau_{\text{valid}})
\end{equation}
where $\tau_{\text{valid}}$ is the validated trajectory. The generated $(P_{\text{gt}}, \mathcal{C}_{\text{gt}})$ undergo quality checks for validity, goal completeness, constraint satisfaction, and tool-chain consistency. They represent one validated feasible solution rather than a unique path or a certificate of global optimality.

\begin{figure*}[htbp]
    \centering
\begin{promptbox}{Step-wise to Holistic Extraction Prompt}

You are an expert at reconstructing reasoning processes from execution trajectories.
Your task is to extract a \textbf{first-person problem-solving plan} and the corresponding \textbf{tool usage chain} based on the given trajectory.

\textbf{Task Objective} \\
From the provided trajectory, extract: \\
1. \textbf{plan} — A concise, first-person summary describing how \textit{I} planned to approach and solve the problem step by step. \\
2. \textbf{tool chain} — An ordered list of tools \textit{I} used during the process, with each `reason` explaining \textit{why I used that tool} in the problem-solving context.

\textbf{Extraction Rules} \\
• This is a \textbf{faithful extraction} task — do not rewrite or invent reasoning not present in the trajectory. \\
• Use \textbf{first-person narration} (e.g., "I first analyzed...", "I planned to use the search tool to find...") rather than meta references like "the model decided to..." or "the agent used...". \\
• The `plan` should describe the \textbf{planning and reasoning process only} — do NOT include actual results, outputs, or findings from tool executions. \\
• Focus on the \textbf{intended approach and strategy}, not on what was discovered or returned by the tools. \\
• The `plan` should be a coherent first-person narrative of the problem-solving strategy. \\
• The `tool\_chain` should include: \\
\hspace*{1em} • `name`: the exact tool name from the trajectory. \\
\hspace*{1em} • `reason`: a brief, first-person explanation of \textit{why I invoked this tool} or \textit{what problem it solved} at that stage. \\
• Preserve the \textbf{original order} of tool invocations. \\
• Only include tools listed in the provided tool list. \\
• Output \textbf{only} a JSON object in this format:
\begin{lstlisting}[language=java, showstringspaces=false, breaklines=true, basicstyle=\ttfamily\scriptsize, columns=fullflexible, frame=none, aboveskip=0pt, belowskip=0pt]
{
    "plan": str,
    "tool_chain": [
        {"name": str, "reason": str},
        ...
    ]
}
\end{lstlisting}
Do not include any commentary or text outside the JSON object.

\textbf{Task Query} \\
\{query\}

\textbf{Available Tools} \\
\{tools\_str\}

\textbf{Trajectory to Analyze} \\
\{trajectory\_str\}

    \end{promptbox}
    \caption{Step-wise to Holistic extraction prompt for distilling high-level planning strategies and tool chains from execution traces.}
    \label{fig:offline_extraction_prompt}
\end{figure*}

\subsection{Robustness Scenario Construction}
\label{app:robustness_construction}
We evaluate robustness via adversarial scenarios probing resilience to extraneous tools, failures, and unsolvable constraints.

\subsubsection{Extraneous Tool Injection}
\label{app:extraneous_injection}
We assess the agent's ability to select appropriate tools amidst distractors. The \textbf{Extraneous Tool Generation Prompt} (Figure \ref{fig:extraneous_tool_prompt}) synthesizes semantically relevant but functionally inert tools, forcing precise semantic analysis.

\subsubsection{Tool-Broken Simulation}
\label{app:broken_simulation}
To evaluate adaptive planning, we simulate tool failures. The \textbf{Broken Tool Substitution Prompt} (Figure \ref{fig:broken_tool_prompt}) generates functional replacements with distinct nomenclature, requiring the agent to identify and use the alternative.

\subsubsection{Unsolvable Task Synthesis}
\label{app:unsolvable_synthesis}
We construct logically unsolvable tasks to test the agent planning models' capacity to refuse impossible requests, using four strategies:
\begin{itemize}[leftmargin=*, label=\small\textbullet, itemsep=2.5pt, parsep=0pt, topsep=0pt]
    \item \textbf{Information Missing:} The \textbf{Information Missing Construction Prompt} (Figure \ref{fig:info_missing_prompt}) reformulates the query to depend on inaccessible private information.
    \item \textbf{Contradictory Constraints:} The \textbf{Contradictory Constraints Construction Prompt} (Figure \ref{fig:constraint_conflict_prompt}) introduces constraints that invalidate standard methods.
    \item \textbf{Tool Removal:} The \textbf{Tool Removal Construction Prompt} (Figure \ref{fig:tool_removal_prompt}) eliminates critical tools and mandates their use.
    \item \textbf{Visual Information Inaccessible:} The \textbf{Visual Information Inaccessible Construction Logic} (Figure \ref{fig:visual_missing_logic}) programmatically removes visual context to test detection of missing information.
\end{itemize}

These constructions instantiate a common missing-precondition principle: no valid plan exists when a necessary prerequisite is absent and cannot be recovered with the supplied context or tools. For example, a symbolic planning task that requires making tea while providing no cup has the same high-level failure structure as APB's missing-information, missing-visual-evidence, or removed-tool cases. APB represents such initial conditions through natural-language instructions, multimodal context, and tool availability rather than an explicit PDDL object inventory. The reported 400-instance evaluation remains restricted to the four implemented constructions above.

\begin{figure*}[htbp]
    \centering
\begin{promptbox}{Extraneous Tool Generation Prompt}

You are an expert in API design and testing. Your task is to generate "distractor" tools that will be mixed with real tools to test an AI agent's ability to select the correct tool.

\textbf{Context} \\
• \textbf{User Query:} "\{query\}" \\
• \textbf{Current Trajectory:} \{trajectory\}... (truncated) \\
• \textbf{Existing Tools (Reference for Format and Context):} \\
\{existing\_tools\_str\}

\textbf{Task} \\
Please generate \{count\} distractor tools. All tools must be \textbf{Same-Scenario but Useless}:

• These tools must fit within the \textbf{EXACT SAME SCENARIO} as the user's query but be \textbf{DEFINITELY USELESS} for the specific goal. \\
• They should look like they belong in the same API suite as the real tools (same naming convention, same resource type). \\
• \textbf{Example:} If the user wants to "book a flight", generate tools like `get\_flight\_seat\_map`, `check\_airport\_lounge\_access`, `download\_airline\_logo`. These are about flights (same scenario) but cannot book a ticket (useless for goal). \\
• \textbf{Tool Description:} The description must be detailed and professional, using domain-specific terminology. It should reinforce that the tool is for the \textit{same scenario} but performs a \textit{different function} than what is needed. \\
• \textbf{Format Requirement:} You MUST strictly follow the JSON structure/schema of the \textbf{Existing Tools}. If existing tools use `tool\_name` instead of `name`, or have specific parameter formats, you MUST mimic them exactly. \\
• \textbf{CRITICAL:} The tool must be \textbf{IMPOSSIBLE} to use for the user's specific goal. It should lack the specific functionality or parameters required to solve the current step of the problem. \\
• \textbf{CRITICAL:} Do NOT generate a tool that overlaps with the solution. It must be clearly distinct in function. \\
• \textbf{CRITICAL:} Do NOT generate tools that are completely unrelated. All tools must be relevant to the domain.

\textbf{Requirements} \\
• Return ONLY a JSON array of tool definitions. \\
• Ensure tool names are distinct from Existing Tools. \\
• Do NOT use generic names like `tool1`, `useless\_tool`. Use descriptive, realistic names.

    \end{promptbox}
    \caption{Extraneous tool generation prompt for synthesizing semantically relevant but functionally inert distractor tools.}
    \label{fig:extraneous_tool_prompt}
\end{figure*}

\begin{figure*}[htbp]
    \centering
\begin{promptbox}{Broken Tool Substitution Prompt}

You are a helpful assistant that generates alternative tools for API development.

The tool `\{tool\_name\}` is currently broken and cannot be used.
Please generate a REPLACEMENT tool that serves the EXACT SAME purpose but has a COMPLETELY DIFFERENT name and slightly different description/parameters to avoid naming conflicts.

\textbf{Original Tool Definition} \\
\{broken\_tool\_json\}

\textbf{Task Context} \\
\{context\}

\textbf{Requirements} \\
1. The new tool MUST have a COMPLETELY DIFFERENT name. \\
\hspace*{1em} - DO NOT just add suffixes like `\_v2', `\_new', `\_alt', `\_1', etc. \\
\hspace*{1em} - DO NOT just add prefixes like `new\_', `my\_', etc. \\
\hspace*{1em} - Use synonyms or different phrasing (e.g., if original is `find\_hotels', use `search\_accommodation' or `query\_lodging'). \\
2. The functionality MUST be identical (same inputs/outputs logic), but you SHOULD vary parameter names slightly if possible to make it look like a different library. \\
3. Return ONLY the JSON definition of the new tool. \\
4. If the original tool used ``tool\_name'', the new tool MUST also use ``tool\_name''. If it used ``name'', use ``name''.

    \end{promptbox}
    \caption{Broken tool substitution prompt for generating functional replacements with distinct nomenclature to simulate tool failures.}
    \label{fig:broken_tool_prompt}
\end{figure*}

\begin{figure*}[htbp]
    \centering
    \begin{promptbox}{Information Missing Construction Prompt}
    You are an expert at constructing adversarial test cases for \textbf{Multimodal AI Planners}.
    Your task is to modify a User Query to make it \textbf{Logically Unsolvable} due to \textbf{Missing Specific Information}.

    \textbf{The Logic Trap:}
    You must rewrite the query so it refers to a specific parameter that the Agent \textbf{cannot know} and \textbf{cannot fetch} with the available tools.

    \textbf{Input Context:}
    \begin{itemize}[leftmargin=*, label=\small\textbullet, itemsep=2.5pt, parsep=0pt, topsep=0pt]
        \item \textbf{Original Query:} "[Original Query]"
        \item \textbf{Original Solution Steps:} [Tool Chain]
        \item \textbf{Available Tools:} [Tool Bank]
    \end{itemize}

    \textbf{CRITICAL CRE-CHECKS (You are blind to the image):}
    \begin{itemize}[leftmargin=*, label=\small\textbullet, itemsep=2.5pt, parsep=0pt, topsep=0pt]
        \item \textbf{Tool Check:} Look at the \texttt{Available Tools}. If you introduce a "verification code" constraint, make sure there is NO tool like \texttt{ReadSMS}, \texttt{CheckEmail}, or \texttt{Get2FA} that could solve it. If such a tool exists, pick a different constraint (e.g., a verbal password).
        \item \textbf{Visual Singularity Risk:} Do NOT rely on adjectives like "the \textit{red} car belonging to Bob". If there is only one red car in the image, the Agent will just select it.
        \begin{itemize}[leftmargin=*, label=\small\textbullet, itemsep=2.5pt, parsep=0pt, topsep=0pt]
            \item \textit{Bad:} "Select the \textbf{red} car that Bob likes." (If only 1 red car, Agent acts).
            \item \textit{Good:} "Select the car that matches the \textbf{license plate number I texted you}." (Invisible attribute, impossible to guess).
        \end{itemize}
    \end{itemize}

    \textbf{Strategies for "Hard Refusal" (Pick the best fit):}

    \textbf{Strategy A: The "Private Reference" (Best for Search/Nav)}
    Refer to information shared in a non-existent, external context (calls, meetings, physical notes).
    \begin{itemize}[leftmargin=*, label=\small\textbullet, itemsep=2.5pt, parsep=0pt, topsep=0pt]
        \item \textit{Original:} "Search for \textbf{'Pizza Hut'}."
        \item \textit{Strong Refusal:} "Search for the specific restaurant \textbf{we discussed in the morning meeting}." (Agent has no meeting logs).
    \end{itemize}

    \textbf{Strategy B: The "Specific Input Requirement" (Best for Forms/Typing)}
    Require a specific string or number that must be typed, which cannot be inferred visually.
    \begin{itemize}[leftmargin=*, label=\small\textbullet, itemsep=2.5pt, parsep=0pt, topsep=0pt]
        \item \textit{Original:} "Rename the file to \textbf{'Final\_Report'}."
        \item \textit{Strong Refusal:} "Rename the file using the \textbf{standardized naming convention code} listed in the employee handbook." (Agent doesn't have the handbook).
    \end{itemize}

    \textbf{Strategy C: The "Invisible Criteria" (Best for Selection)}
    Make the selection depend on a hidden attribute, not a visual one.
    \begin{itemize}[leftmargin=*, label=\small\textbullet, itemsep=2.5pt, parsep=0pt, topsep=0pt]
        \item \textit{Original:} "Click the \textbf{cheapest} item." (Price is visible, this is solvable).
        \item \textit{Strong Refusal:} "Click the item that \textbf{is eligible for the 'Employee Friends \& Family' discount}." (Eligibility is invisible).
    \end{itemize}

    \textbf{Rules:}
    \begin{itemize}[leftmargin=*, label=\small\textbullet, itemsep=2.5pt, parsep=0pt, topsep=0pt]
        \item \textbf{Keep the Visual Context:} Do not remove visual descriptions (e.g., "red button", "top menu") if they are part of the original intent, BUT add a logical gate that blocks the final action.
        \item \textbf{No Negatives:} Do not say "do not use tools".
        \item \textbf{Natural Tone:} The user should sound like they \textit{forgot} the Agent doesn't know this info.
    \end{itemize}

    \textbf{Output:}
    Return \textbf{ONLY} the rewritten query string.
    \end{promptbox}
    \caption{Information missing construction prompt for reformulating queries to depend on inaccessible private information.}
    \label{fig:info_missing_prompt}
\end{figure*}

\begin{figure*}[htbp]
    \centering
    \begin{promptbox}{Contradictory Constraints Construction Prompt}
    You are an expert at constructing adversarial test cases for \textbf{Multimodal AI Planners}.
    Your goal is to modify a User Query to make it \textbf{Logically Unsolvable}, forcing the agent to admit it cannot create a valid plan.

    \textbf{The Logic Trap:}
    The Agent uses a powerful VLM (Visual Language Model). If you just say "The tool is down", the Agent might say "I can see the image, I will analyze it myself."
    \textbf{YOU MUST PREVENT THIS.} Your constraint must imply that \textbf{visual inspection alone is insufficient, misleading, or illegal} without the tool's specific metadata.

    \textbf{Input Context:}
    \begin{itemize}[leftmargin=*, label=\small\textbullet, itemsep=2.5pt, parsep=0pt, topsep=0pt]
        \item \textbf{Original Query:} "[Original Query]"
        \item \textbf{Original Solution Steps:} [Tool Chain]
        \item \textbf{Available Tools:} [Tool Bank]
    \end{itemize}

    \textbf{Constraint Strategies (Pick one that fits best):}

    \textbf{Strategy A: The "Missing Dimension" (Best for Video/Gaia/Analysis)}
    The visual data exists, but it lacks a critical invisible dimension required for the answer.
    \begin{itemize}[leftmargin=*, label=\small\textbullet, itemsep=2.5pt, parsep=0pt, topsep=0pt]
        \item \textit{Bad:} "The face analysis tool is down." (Agent will just look at the face).
        \item \textit{Good:} "Determine the emotion using the \textbf{biometric pulse data overlay} (currently corrupted). Visual expressions are misleading due to the actor's neurological condition and cannot be trusted."
        \item \textit{Good:} "Analyze the chess position. Note that this is a variant where \textbf{piece colors are inverted relative to thermal signatures}, and only the \texttt{ThermalDecoder} (offline) can identify true ownership."
    \end{itemize}

    \textbf{Strategy B: The "Compliance Lock" (Best for GTA/Counting/Finance)}
    Make the \textit{method} of data acquisition legally binding.
    \begin{itemize}[leftmargin=*, label=\small\textbullet, itemsep=2.5pt, parsep=0pt, topsep=0pt]
        \item \textit{Example:} "For official audit, the count must be cryptographically signed by the \texttt{CountObject} tool. Unverified visual estimates are strictly prohibited by law."
    \end{itemize}

    \textbf{Strategy C: The "Dynamic State" (Best for AgentNet/Web)}
    The visual screenshot is outdated or static; reality requires a live check.
    \begin{itemize}[leftmargin=*, label=\small\textbullet, itemsep=2.5pt, parsep=0pt, topsep=0pt]
        \item \textit{Example:} "Click the button, but only if the \textbf{server-side session token} (checked via \texttt{VerifySession}) is active. The session server is down."
    \end{itemize}

    \textbf{Task Instructions:}
    \begin{itemize}[leftmargin=*, label=\small\textbullet, itemsep=2.5pt, parsep=0pt, topsep=0pt]
        \item Identify the core dependency (e.g., Emotion, Count, Click).
        \item Create a constraint that makes the \textbf{Visual Information alone insufficient or untrustworthy}.
        \item \textbf{NEVER} suggest using "alternative methods" or "manual checks". Explicitly state that without the tool/data, the task \textbf{cannot be completed}.
        \item Append 1-2 natural sentences to the query.
    \end{itemize}

    \textbf{Output:}
    Return \textbf{ONLY} the modified query string (Original Query + Your Constraint).
    \end{promptbox}
    \caption{Constraint conflict construction prompt for introducing constraints that invalidate standard operational methods.}
    \label{fig:constraint_conflict_prompt}
\end{figure*}

\begin{figure*}[htbp]
    \centering
    \begin{promptbox}{Tool Removal Construction Prompt}
    You are an expert at constructing adversarial test cases for \textbf{Multimodal AI Planners}.
    The goal is to create an "Unsolvable" test case by \textbf{Removing Critical Capabilities} and \textbf{Enforcing Tool Dependency}.

    \textbf{The Logic:}
    We want the Agent to realize: \textit{"I need to perform this action using a tool to be compliant, but I have NO tools available to do it."}

    \textbf{Your Task 1: Select Tools to Remove (The "Capability Void")}
    \begin{itemize}[leftmargin=*, label=\small\textbullet, itemsep=2.5pt, parsep=0pt, topsep=0pt]
        \item Identify the \textbf{core capability} required (e.g., Counting, Clicking, OCR, Searching).
        \item Remove \textbf{ALL tools} associated with that capability (primary AND backups) to ensure no alternative exists.
        \item \textbf{Target:} Action Tools (click, type) or specialized Perception Tools (Count, OCR, Detect).
        \item \textbf{Select AT LEAST 2 tools} if possible to ensure a complete block.
    \end{itemize}

    \textbf{Your Task 2: Rewrite the Query (The "Compliance Constraint")}
    \begin{itemize}[leftmargin=*, label=\small\textbullet, itemsep=2.5pt, parsep=0pt, topsep=0pt]
        \item Append a constraint to the User Query that \textbf{forbids} using internal capabilities (like "visual estimation" or "mental calculation").
        \item \textbf{CRITICAL:} Do \textbf{NOT} name the specific tools (e.g., don't say "Use CountObject"). Instead, use \textbf{generic functional descriptions}.
        \begin{itemize}[leftmargin=*, label=\small\textbullet, itemsep=2.5pt, parsep=0pt, topsep=0pt]
            \item \textit{Bad:} "Use \texttt{CountGivenObject} to count." (Too obvious contradiction).
            \item \textit{Good:} "For audit compliance, the count must be \textbf{generated by a verified counting tool}. Visual estimates are strictly prohibited."
            \item \textit{Good:} "You must use \textbf{programmatic navigation actions} to interact with the map. Manual interactions are not logged and therefore invalid."
        \end{itemize}
        \item \textbf{NO} "IMPORTANT", "WARNING", "COMPLIANCE", "NOTE:", or bold headers.
    \end{itemize}

    \textbf{Input Context:}
    \begin{itemize}[leftmargin=*, label=\small\textbullet, itemsep=2.5pt, parsep=0pt, topsep=0pt]
        \item \textbf{Original Query:} "[Original Query]"
        \item \textbf{Ground Truth Tool Chain:} [Tool Chain]
        \item \textbf{Available Tools:} [Tool Bank]
    \end{itemize}

    \textbf{Output Format:}
    Return \textbf{ONLY} a JSON object:
    \begin{lstlisting}[language=java, showstringspaces=false, basicstyle=\scriptsize\ttfamily, breaklines=true, breakatwhitespace=true]
{
"tools_to_remove": ["ToolName1", "ToolName2", ...],
"modified_query": "The new query string with the generic compliance constraint..."
}
    \end{lstlisting}
    \end{promptbox}
    \caption{Tool removal construction prompt for creating logical deadlocks by eliminating critical tools while mandating their use.}
    \label{fig:tool_removal_prompt}
\end{figure*}

\begin{figure*}[htbp]
    \centering
\begin{lstlisting}[language=Python, showstringspaces=false, basicstyle=\ttfamily\scriptsize, frame=single, breaklines=true]
def construct_visual_missing_case(sample):
    """
    Constructs a 'Visual Information Inaccessible' unsolvable case by stripping visual context.

    """
    # 1. Preserve the original query
    # We deliberately keep the query unchanged (e.g., "Describe the image")
    # to test if the agent can correctly identify that the image is missing.
    query = sample["query"]

    # 2. Remove Visual Information
    # The core logic is to empty the file list in the background context,
    # effectively simulating a system error where the image failed to load
    # or was not provided.
    if "background" in sample:
        sample["background"]["files"] = []

    # 3. Mark as Unsolvable
    sample["unsolvable_type"] = "visual_missing"

    return sample
\end{lstlisting}
    \caption{Visual missing construction logic for programmatically removing visual data to test agent detection of missing context.}
    \label{fig:visual_missing_logic}
\end{figure*}

\subsection{Efficiency-Aware Planning Construction}
\label{app:efficiency_construction}
To evaluate cost-efficiency optimization, we construct the \textbf{Efficiency-Aware Planning} task. We augment holistic planning data with cost annotations and tools creating efficiency trade-offs. Each original tool receives a base cost (1 to 5), and we generate 5 to 10 new tools: (1) \textit{Efficient Composite Tools} (Type A, 3 to 6 tools) combining 2 to 3 steps into one lower-cost tool, and (2) \textit{Trap Tools} (Type B, 2 to 4 tools) that are suboptimal (overpriced, redundant, or off-target).

The goal is to ensure efficiency-aware agents use composite tools to minimize cost while avoiding trap tools. Figure~\ref{fig:efficiency_prompt} presents the generation prompt. All samples undergo manual verification.

\begin{figure*}[htbp]
    \centering
    \begin{promptbox}{Efficiency-Aware Tool Bank Generation Prompt}
    \textbf{System Prompt:} \\
    You are an expert in designing Agent Planning Benchmarks.
    Your goal is to REWRITE the `tool\_bank' for a given task to test `Plan Efficiency'.

    \vspace{0.5em}
    \textbf{Core Logic:} \\
    You must provide a tool set where multiple paths exist to solve the problem, but one path (using Composite Tools) is significantly cheaper (lower total cost) than others.

    \vspace{0.5em}
    \textbf{Instructions:}
    \begin{itemize}[leftmargin=*, label=\small\textbullet, itemsep=2.5pt, parsep=0pt, topsep=0pt]
        \item \textbf{Original Tools (Mandatory):} Keep ALL original tools. Assign them a reasonable base cost (e.g., integers between 1 and 5).

        \item \textbf{ADD 5 to 10 NEW Tools (Strict Constraint):} Generate exactly 5 to 10 new tools divided as follows:
        \begin{itemize}[leftmargin=*, label=\small\textbullet, itemsep=2.5pt, parsep=0pt, topsep=0pt]
            \item \textbf{Type A: Efficient Composite Tools (3 to 6 tools):} Combine 2 to 3 consecutive steps from the Ground Truth into one single tool. The cost MUST be LOWER than the sum of the original tools it replaces (e.g., if Tool A cost 2 and Tool B cost 2, Composite Tool AB should cost 3).
            \item \textbf{Type B: Inefficient/Trap Tools (2 to 4 tools):} Tools that look relevant but are suboptimal: \textit{Overpriced} (simple job, very high cost), \textit{Redundant Steps} (breaks simple action into unnecessary sub-steps), or \textit{Slightly Off-target} (retrieves partial information, forcing more tool calls).
        \end{itemize}

        \item \textbf{NO Magic Tools:} Do not create a single tool that solves the whole query in one step.

        \item \textbf{Output Format:} Return a SINGLE valid JSON list containing [All Original Tools + All New Tools]. Each tool object must have: \texttt{name}, \texttt{description}, \texttt{parameters}, and \texttt{cost} (integer).
    \end{itemize}

    \vspace{0.5em}
    \hrule
    \vspace{0.5em}

    \textbf{User Message:} \\
    \textbf{User Query:} \{query\} \\
    \textbf{Original Ground Truth / Steps:} \{tool\_chain or plan\} \\
    \textbf{Original Tool Bank:} \{original\_tools\} \\
    \textbf{Requirement:} Generate a JSON list of tools (original + composite + traps) with costs. Output PURE JSON only.
    \end{promptbox}
    \caption{Efficiency-aware tool bank generation prompt for creating cost-annotated tools with composite (efficient) and trap (inefficient) alternatives.}
    \label{fig:efficiency_prompt}
\end{figure*}

\subsection{Data Construction Case Studies}
\label{app:data_construction_cases}
This section presents illustrative case studies covering data augmentation, holistic planning, robustness scenarios (including tool-broken, tool-extraneous, and unsolvable tasks), and efficiency-aware planning.

\subsubsection{Data Augmentation Case Studies}
\label{app:augmentation_cases}
We present case studies from GTA, GAIA, ToolBench, and FrameThinker to demonstrate the data augmentation pipeline's efficacy.

As shown in the following figures, the augmentation process consistently increases task complexity:
\begin{itemize}[leftmargin=*, label=\small\textbullet, itemsep=2.5pt, parsep=0pt, topsep=0pt]
    \item \textbf{GTA (Figure \ref{fig:gta_augmentation}):} Transforms a simple visual identification task into a multi-step financial workflow involving OCR, arithmetic reasoning, and policy compliance.
    \item \textbf{GAIA (Figure \ref{fig:gaia_augmentation}):} Expands a single-entity query into a comparative analysis across multiple journals, requiring weighted calculations and tabular synthesis.
    \item \textbf{ToolBench (Figure \ref{fig:toolbench_augmentation}):} Evolves a weather check into an event planning scenario with contingency management and document generation.
    \item \textbf{FrameThinker (Figure \ref{fig:framethinker_augmentation}):} Converts a multiple-choice visual question into a narrative analysis, demanding character and emotional tracking.
\end{itemize}

These transformations ensure the benchmark rigorously probes long-horizon planning and reasoning.

\begin{figure*}[htbp]
    \centering
    \begin{promptbox}{GTA Augmentation Example}
    \textbf{Original Sample} \\
    \textbf{Query:} "How much should I pay for the beer on the table according to the price on the menu?" \\
    \textbf{Trajectory Summary (5 Steps):}
    \begin{itemize}[leftmargin=*, label=\small\textbullet, itemsep=2.5pt, parsep=0pt, topsep=0pt]
        \item \texttt{ImageDescription}: Describe the beer bottles on the table.
        \item \texttt{ImageDescription}: Describe the menu image.
        \item \texttt{CountGivenObject}: Count the number of beer bottles.
        \item \texttt{OCR}: Read the beer brand from the bottle label.
        \item \texttt{OCR}: Read the price from the menu.
    \end{itemize}
    \vspace{0.5em}
    \hrule
    \vspace{0.5em}
    \textbf{Augmented Sample} \\
    \textbf{Query:} "I'm dining at this beachfront restaurant and need a detailed expense breakdown. First, identify what beverages are on my table and what menu is shown. Then calculate the total cost for all beverages, apply the restaurant's 10\% service charge and 7\% sales tax, split the final bill among 3 people, verify the origin country of the beer brand, check if any promotional discounts apply for ordering multiple beers, and finally provide a complete itemized receipt with per-person cost." \\
    \textbf{Trajectory Summary (11 Steps):}
    \begin{itemize}[leftmargin=*, label=\small\textbullet, itemsep=2.5pt, parsep=0pt, topsep=0pt]
        \item \texttt{ImageDescription}: Identify beverages (beer bottles) in the first image.
        \item \texttt{ImageDescription}: Confirm the second image is the menu.
        \item \texttt{CountGivenObject}: Count the exact number of beer bottles (2).
        \item \texttt{OCR}: Read the beer brand ("MAGNA") from the bottle.
        \item \texttt{OCR}: Read prices from the menu image.
        \item \texttt{ParseMenuItems}: Structure the menu data to find the price of Magna beer.
        \item \texttt{CheckPromotionalDiscount}: Check for quantity discounts (found 5\% off).
        \item \texttt{VerifyProductOrigin}: Determine the origin of the beer (Puerto Rico).
        \item \texttt{CalculateBillWithCharges}: Calculate total with discount, service charge, and tax.
        \item \texttt{SplitBillAmongPeople}: Split the total amount among 3 people.
        \item \texttt{GenerateItemizedReceipt}: Generate a detailed receipt with all findings.
    \end{itemize}
    \end{promptbox}
    \caption{GTA augmentation example: transforming a simple visual identification task into a multi-step financial analysis workflow requiring OCR, arithmetic reasoning, and policy compliance.}
    \label{fig:gta_augmentation}
\end{figure*}

\begin{figure*}[htbp]
    \centering
    \begin{promptbox}{GAIA Augmentation Example}
    \textbf{Original Sample} \\
    \textbf{Query:} "If we assume all articles published by Nature in 2020 ... relied on statistical significance ... and they on average came to a p-value of 0.04, how many papers would be incorrect ...?" \\
    \textbf{Trajectory Summary (2 Steps):}
    \begin{itemize}[leftmargin=*, label=\small\textbullet, itemsep=2.5pt, parsep=0pt, topsep=0pt]
        \item \texttt{search\_engine}: Find the number of articles published in Nature in 2020.
        \item \texttt{calculator}: Calculate 4\% of the total count.
    \end{itemize}
    \vspace{0.5em}
    \hrule
    \vspace{0.5em}
    \textbf{Augmented Sample} \\
    \textbf{Query:} "I'm investigating the reliability of statistical claims in high-impact journals. For Nature in 2020, I need to: (1) determine how many research articles were published, (2) calculate how many would be false positives if all used p=0.04, (3) compare this false positive rate against three other major journals (Science, Cell, and The Lancet) ... (4) adjust each journal's false positive count by their respective impact factors ... (5) generate a ranked comparison table, and (6) calculate what the aggregate false positive rate would be ..." \\
    \textbf{Trajectory Summary (12 Steps):}
    \begin{itemize}[leftmargin=*, label=\small\textbullet, itemsep=2.5pt, parsep=0pt, topsep=0pt]
        \item \texttt{search\_engine} \& \texttt{calculator}: Retrieve article counts and calculate false positives for \textit{Nature}.
        \item \texttt{search\_engine} \& \texttt{calculator}: Repeat for \textit{Science}.
        \item \texttt{search\_engine} \& \texttt{calculator}: Repeat for \textit{Cell}.
        \item \texttt{search\_engine} \& \texttt{calculator}: Repeat for \textit{The Lancet}.
        \item \texttt{search\_engine}: Find 2020 impact factors for all four journals.
        \item \texttt{weighted\_metric\_calculator}: Calculate influence-weighted false positive counts.
        \item \texttt{tabular\_data\_formatter}: Create a ranked comparison table.
        \item \texttt{calculator}: Compute the aggregate false positive count.
    \end{itemize}
    \end{promptbox}
    \caption{GAIA augmentation example: expanding a single-entity statistical query into multi-journal comparative analysis with weighted calculations and tabular synthesis.}
    \label{fig:gaia_augmentation}
\end{figure*}

\begin{figure*}[htbp]
    \centering
    \begin{promptbox}{ToolBench Augmentation Example}
    \textbf{Original Sample} \\
    \textbf{Query:} "I'm planning a romantic dinner on a rooftop terrace. Can you provide me with the sunset time for the date of the dinner? Additionally, I'd like to know the current weather conditions and a 5-day forecast ..." \\
    \textbf{Trajectory Summary (3 Steps):}
    \begin{itemize}[leftmargin=*, label=\small\textbullet, itemsep=2.5pt, parsep=0pt, topsep=0pt]
        \item \texttt{get\_sunrise\_and\_sunset\_times}: Get sunset time.
        \item \texttt{weather\_report}: Get current weather.
        \item \texttt{get\_5\_day\_forecast}: Attempt to get forecast (failed in original).
    \end{itemize}
    \vspace{0.5em}
    \hrule
    \vspace{0.5em}
    \textbf{Augmented Sample} \\
    \textbf{Query:} "I'm planning a romantic anniversary dinner ... I need comprehensive planning data: exact sunset time with twilight phases, current weather, detailed 5-day and 16-day forecasts, historical weather patterns ..., solar position ..., UV index, air quality, wind patterns at rooftop level ..., precipitation ..., temperature trends ... cross-reference with restaurant reservation availability, backup indoor venue ..., generate risk assessment ..., compile PDF report ..." \\
    \textbf{Trajectory Summary (14 Steps):}
    \begin{itemize}[leftmargin=*, label=\small\textbullet, itemsep=2.5pt, parsep=0pt, topsep=0pt]
        \item \texttt{get\_sunrise\_and\_sunset\_times}: Get detailed sunset and twilight data.
        \item \texttt{weather\_report}: Get current baseline conditions.
        \item \texttt{get\_5\_day\_forecast} \& \texttt{view\_16\_day\_forecast}: Get short and long-term forecasts.
        \item \texttt{climate\_forecast} \& \texttt{QueryHistoricalWeather}: Analyze statistical weather patterns.
        \item \texttt{sunposition}: Calculate solar angle for table placement.
        \item \texttt{FetchAirQuality} \& \texttt{QueryWindProfile}: Check comfort factors (AQI, rooftop wind).
        \item \texttt{QueryReservation}: Find available rooftop tables.
        \item \texttt{GenerateBackupVenueAnalysis}: Plan contingencies for bad weather.
        \item \texttt{GenerateWeatherRiskAssessment}: Synthesize all data into a risk report.
        \item \texttt{GeneratePDFReport}: Compile findings into a document.
        \item \texttt{ConfigureWeatherAlertSystem}: Set up active monitoring.
    \end{itemize}
    \end{promptbox}
    \caption{ToolBench augmentation example: transforming a simple weather query into comprehensive event planning with contingency management and document generation.}
    \label{fig:toolbench_augmentation}
\end{figure*}

\begin{figure*}[htbp]
    \centering
    \begin{promptbox}{FrameThinker Augmentation Example}
    \textbf{Original Sample} \\
    \textbf{Query:} "The scene where a large number of foxes surround the girl at 2:56 implies what? A: The fox clan collectively seeks revenge. B: The girl develops schizophrenia. C: The underground imprisonment will continue. D: Mom comes to the rescue. E: The fox is preparing to execute the girl. F: The girl gains superpowers." \\
    \textbf{Trajectory Summary (3 Steps):}
    \begin{itemize}[leftmargin=*, label=\small\textbullet, itemsep=2.5pt, parsep=0pt, topsep=0pt]
        \item \texttt{get\_frame\_number}: Convert timestamp "2:56" to frame number (4224).
        \item \texttt{choose\_frames}: Request frames around frame 4224 to analyze the scene.
        \item \texttt{output\_answer}: Conclude "C: The underground imprisonment will continue" based on visual analysis.
    \end{itemize}
    \vspace{0.5em}
    \hrule
    \vspace{0.5em}
    \textbf{Augmented Sample} \\
    \textbf{Query:} "I need you to perform a comprehensive narrative and thematic analysis of this stop-motion animated sequence. Please analyze the following aspects systematically: A) Identify and catalog all distinct environmental settings... B) Track the protagonist's emotional arc... C) Detect and analyze the presence and behavior of the fox/rabbit-like creatures... D) Synthesize your findings... to determine the thematic significance of the scene around frame 4224..." \\
    \textbf{Trajectory Summary (8 Steps):}
    \begin{itemize}[leftmargin=*, label=\small\textbullet, itemsep=2.5pt, parsep=0pt, topsep=0pt]
        \item \texttt{classify\_environment}: Catalog settings across all frames (underground vs. domestic).
        \item \texttt{detect\_character\_presence}: Track the girl's visibility across frames.
        \item \texttt{analyze\_emotional\_state}: Analyze the girl's emotions (fear, distress) in key frames.
        \item \texttt{detect\_character\_presence}: Track the fox/rabbit creatures' presence.
        \item \texttt{filter\_frames}: Isolate frames around the target timestamp (4224).
        \item \texttt{analyze\_spatial\_relationship}: Analyze proximity and power dynamics between girl and creature.
        \item \texttt{analyze\_narrative\_theme}: Evaluate thematic evidence for captivity vs. other interpretations.
        \item \texttt{synthesize\_thematic\_significance}: Synthesize all findings to determine the scene's meaning (Ongoing Captivity).
    \end{itemize}
    \end{promptbox}
    \caption{FrameThinker augmentation example: converting a multiple-choice visual question into systematic multi-dimensional narrative and thematic analysis.}
    \label{fig:framethinker_augmentation}
\end{figure*}

\subsubsection{Holistic Planning Case Studies}
\label{app:data_holistic_cases}
Figure \ref{fig:holistic_vs_stepwise} illustrates the transformation from step-wise to holistic planning. We convert execution trajectories (with observations) into a holistic format consisting of a high-level \textbf{Plan} (synthesized strategy) and an observation-free \textbf{Tool Chain} (sequence of tools).

\begin{figure*}[htbp]
    \centering
    \begin{promptbox}{Holistic Data Construction Example}
    \textbf{Original Step-wise Data (Input)}
    \textit{Contains the full interaction history with observations:}
    \begin{itemize}[leftmargin=*, label=\small\textbullet, itemsep=2.5pt, parsep=0pt, topsep=0pt]
        \item \textbf{Step 1:} Call \texttt{get\_sunset\_time(...)}
        \item \textit{Observation:} "Sunset is at 20:12."
        \item \textbf{Step 2:} Call \texttt{get\_weather(...)}
        \item \textit{Observation:} "Clear skies, 18°C."
        \item ... (Full trajectory with 14 steps and observations)
    \end{itemize}

    \vspace{0.5em}
    \centerline{$\downarrow$ \textbf{Transformation Process} $\downarrow$}
    \vspace{0.5em}

    \textbf{Transformed Holistic Data (Output)}
    \textit{Synthesized Plan and Extracted Tool Chain (No observations):}

    \textbf{1. Synthesized Plan:}
    "I planned to approach this comprehensive dinner planning task by systematically gathering all required data in logical sequence. First, I would obtain the exact sunset time... Next, I would establish baseline weather conditions... then progressively build confidence through multiple forecast horizons... Finally, I would compile everything into a formatted PDF report..."

    \textbf{2. Extracted Tool Chain:}
    \begin{itemize}[leftmargin=*, label=\small\textbullet, itemsep=2.5pt, parsep=0pt, topsep=0pt]
        \item \texttt{get\_sunset\_time}
        \item \texttt{weather\_report}
        \item \texttt{get\_5\_day\_forecast}
        \item \texttt{view\_16\_day\_forecast}
        \item \texttt{calculate\_solar\_position}
        \item ... (Sequence of 14 tools)
    \end{itemize}
    \end{promptbox}
    \caption{Data transformation example: converting step-wise trajectory into descriptive \textbf{Plan} with extracted \textbf{Tool Chain} for holistic planning ground truth.}
    \label{fig:holistic_vs_stepwise}
\end{figure*}

\subsubsection{Robustness Case Studies}
\label{app:robustness_cases}
This section details Tool-Broken, Tool-Extraneous, and Unsolvable scenarios.

Figure \ref{fig:tool_broken_case} illustrates a Tool-Broken scenario. The agent is planning a romantic dinner and attempts to fetch air quality data using \texttt{FetchAirQualityAPI}, but the tool is simulated to be broken. The agent must identify this failure and switch to the designated replacement tool, \texttt{GetAtmosphericPollutionData}.

\begin{figure*}[htbp]
    \centering
    \begin{promptbox}{Tool-Broken Case Study}
    \textbf{Trajectory Prefix:}
    \begin{itemize}[leftmargin=*, label=\small\textbullet, itemsep=2.5pt, parsep=0pt, topsep=0pt]
        \item call: GetSunsetTime(date='2024-05-18', location='San Francisco')
        \item call: GetWeatherForecast(date='2024-05-18', location='San Francisco')
        \item ... (Previous successful steps)
    \end{itemize}

    \textbf{Broken Tool:}
    \texttt{FetchAirQualityAPI} (Simulated Failure)

    \textbf{Designated Replacement Tool:}
    \texttt{GetAtmosphericPollutionData}
    \begin{itemize}[leftmargin=*, label=\small\textbullet, itemsep=2.5pt, parsep=0pt, topsep=0pt]
        \item Description: Obtains real-time and projected atmospheric pollution metrics including air quality index measurements and specific contaminant levels from environmental sensor networks. Parameters: 'lat', 'lon', 'target\_date'. Returns JSON object containing present AQI measurement and health advisory information.
    \end{itemize}
    \end{promptbox}
    \caption{Tool-broken case study from ToolBench: agent recovery from \texttt{FetchAirQualityAPI} failure by utilizing the alternative \texttt{GetAtmosphericPollutionData}.}
    \label{fig:tool_broken_case}
\end{figure*}

Figure \ref{fig:tool_extraneous_case} depicts a Tool-Extraneous scenario. The system injects several extraneous tools (e.g., \texttt{stock\_v2\_get\_company\_logo}) that appear related to the entity but are functionally irrelevant. The agent must ignore these distractors and use only the necessary tools.

\begin{figure*}[htbp]
    \centering
    \begin{promptbox}{Tool-Extraneous Case Study}
    \textbf{User Query:}
    I need to conduct a comprehensive multi-dimensional analysis of stock performanceId 0P0000OQN8... Please provide: (1) Complete security information... (2) Key financial ratios... (3) Detailed key statistics... (5) Current order book depth...

    \textbf{Injected Extraneous Tools (Distractors):}
    \begin{itemize}[leftmargin=*, label=\small\textbullet, itemsep=2.5pt, parsep=0pt, topsep=0pt]
        \item \texttt{stock\_v2\_get\_company\_logo\_for\_morning\_star}: Retrieves the official company logo and branding assets. (Irrelevant to quantitative analysis)
        \item \texttt{stock\_v2\_get\_executive\_profiles\_for\_morning\_star}: Fetches biographical information of executives. (Irrelevant to order book/price analysis)
        \item ...
    \end{itemize}
    \end{promptbox}
    \caption{Tool extraneous case study: agent must avoid semantically related but task-irrelevant distractor tools during financial analysis.}
    \label{fig:tool_extraneous_case}
\end{figure*}

Figure \ref{fig:unsolvable_cases} presents examples for each of the four Unsolvable categories: Contradictory Constraints, Information Missing, Tool Removal, and Visual Information Inaccessible.

\begin{figure*}[htbp]
    \centering
    \begin{promptbox}{Unsolvable Case Studies}
    \textbf{1. Contradictory Constraints:}
    \textit{Query:} "Open your store on the Steam Store homepage... filter by date... save as default preference... Note that due to GDCR compliance requirements, all preference-saving operations must be verified through the \texttt{SessionAuthValidator} tool... The \texttt{SessionAuthValidator} service is currently experiencing a critical outage..."

    \textbf{2. Missing Information:}
    \textit{Query:} "Update twitter profile: add the bio prefix that I mentioned in our phone call earlier before the current bio, and change the location to the city code from the company's regional office list." (The specific bio prefix and city code are not provided in the context).

    \textbf{3. Tool Removal:}
    \textit{Query:} "Please search for the nearest open pizza place... All interface interactions must be executed through registered interaction handlers..."
    \textit{Available Tools:} \texttt{["moveto", "dragto", "write", "press", "hotkey", "scroll"]} (The critical \texttt{click} tool is removed).

    \textbf{4. Visual Information Inaccessible:}
    \textit{Query:} "Please search for the nearest open pizza place to my home on Google Maps..."
    \textit{Visual Context:} \texttt{[]} (The necessary screenshot/visual interface is not provided).
    \end{promptbox}
    \caption{Unsolvable scenario examples: tasks blocked by external constraints (service outage), missing information (unknown bio prefix), missing tools (no click action), or missing visual context (no screenshot).}
    \label{fig:unsolvable_cases}
\end{figure*}

\subsubsection{Efficiency-Aware Planning Case Studies}
\label{app:efficiency_cases}
Figure \ref{fig:efficiency_case} demonstrates Efficiency-Aware Planning. The tool bank includes original tools, \textit{Efficient Composite Tools} (Type A) that combine multiple steps for cost reduction, and \textit{Trap Tools} (Type B) that appear relevant but are suboptimal (overpriced or redundant).

\begin{figure*}[htbp]
    \centering
    \begin{promptbox}{Efficiency-Aware Planning Case Study}
    \textbf{User Query:} I need a comprehensive analysis of this supernatural ritual sequence. Using the following 16 frames, please identify the chronological progression of ritual phases, track emotional arc, and detect the manifestation moment...

    \vspace{0.5em}
    \textbf{Original Tools (Base Costs):}
    \begin{itemize}[leftmargin=*, label=\small\textbullet, itemsep=2.5pt, parsep=0pt, topsep=0pt]
        \item \texttt{classify\_sequence\_phases}: cost=3
        \item \texttt{analyze\_emotional\_state}: cost=2
        \item \texttt{detect\_text\_appearance}: cost=2
        \item \texttt{compare\_frames}: cost=2
        \item \texttt{classify\_environment}: cost=3
    \end{itemize}

    \textbf{Efficient Composite Tools (Type A: Lower Combined Cost).}
    \begin{itemize}[leftmargin=*, label=\small\textbullet, itemsep=2.5pt, parsep=0pt, topsep=0pt]
        \item \texttt{analyze\_ritual\_progression\_with\_emotion}: cost=4 (combines phase classification + emotional analysis, saves 1 vs. separate calls)
        \item \texttt{detect\_manifestation\_moment}: cost=3 (combines text detection + frame comparison, saves 1)
        \item \texttt{comprehensive\_ritual\_analysis}: cost=6 (combines phase + emotional + environment analysis, saves 2)
    \end{itemize}

    \textbf{Trap Tools (Type B: Overpriced or Redundant).}
    \begin{itemize}[leftmargin=*, label=\small\textbullet, itemsep=2.5pt, parsep=0pt, topsep=0pt]
        \item \texttt{extract\_individual\_frame\_metadata}: cost=8 (extracts detailed metadata for single frame, which is overpriced)
        \item \texttt{analyze\_micro\_expressions}: cost=7 (micro-expression analysis, which has redundant granularity)
        \item \texttt{detect\_ambient\_sound\_correlation}: cost=5 (audio analysis, which is off-target for visual task)
    \end{itemize}
    \end{promptbox}
    \caption{Efficiency-aware planning case study: tool bank with original tools, cost-efficient composite alternatives, and overpriced trap tools.}
    \label{fig:efficiency_case}
\end{figure*}

\section{Evaluation Framework}
\label{app:evaluation_framework}

\subsection{Planning Assumptions and Scope}
\label{app:planning_scope}
APB is an execution-free plan-synthesis benchmark over natural-language or multimodal queries, system instructions, and tool schemas. It is not formulated as PDDL, an MDP, or a POMDP: it does not define symbolic state variables, action preconditions and effects, belief states, transition distributions, or outcome probabilities. Holistic Planning assumes that the supplied tools function as specified and provides no intermediate feedback. Step-wise Planning conditions on a fixed observed trajectory prefix and its recorded tool returns. For augmented trajectories, tool calls are designed to be deterministic and successful; Tool-Broken Planning instead supplies one fixed observed failure and a designated alternative rather than sampling a stochastic transition. Information Missing and Visual Information Inaccessible evaluate feasibility and refusal, not probabilistic inference over latent states. Consequently, APB neither supplies action-outcome likelihoods nor evaluates a model's ability to estimate them.

\subsection{Evaluation Metrics}
\label{app:evaluation_metrics}
We employ three complementary metrics: \textbf{Plan Correctness} (pass/fail), \textbf{Plan Grade} (0 to 1 scale), and \textbf{Error Taxonomy} (E1 to E6). The E1--E6 taxonomy was not manually imposed a priori; we first inspected representative planning failures, used LLMs to cluster and summarize recurring error patterns, and then finalized the categories through human checks for semantic independence and annotatability. To systematically diagnose failure mechanisms, the final taxonomy contains \textbf{six} semantically independent error categories, where a sample may exhibit multiple errors simultaneously. Notably, E2 instantiates two protocol-specific subtypes that capture a key distinction between step-wise and holistic planning.

\subsubsection{Step-wise Planning Metrics}
\label{app:stepwise_metrics}
Step-wise evaluation enforces \textbf{zero tolerance}: any flaw constitutes an error. A step is correct only if it contains zero errors (E1 to E6 all False). The focus is on \textbf{correctness and reasonableness}, not optimality.

\paragraph{Plan Grade (Step-wise).}
\begin{itemize}[leftmargin=*, label=\small\textbullet, itemsep=1.5pt, parsep=0pt, topsep=2pt]
    \item \textbf{1.0 (Perfect):} No errors. The step is correct, logical, and progresses toward the goal. \textit{Note: The step does NOT need to be ``optimal''; it just needs to be correct and reasonable.}
    \item \textbf{0.8 (Very Good):} At most one very minor issue (E5 only with trivial impact) that does not affect correctness.
    \item \textbf{0.6 (Acceptable):} One minor error (E4 or E5) that does not fundamentally break the solution.
    \item \textbf{0.4 (Problematic):} Multiple minor errors (2+ of E4/E5) or one moderate error (E2).
    \item \textbf{0.2 (Severe):} Major errors (E1, E3, or E6) or multiple moderate errors.
    \item \textbf{0.0 (Failed):} Multiple major errors or complete logical breakdown.
\end{itemize}

\paragraph{Error Taxonomy (Step-wise).}
\begin{itemize}[leftmargin=*, label=\small\textbullet, itemsep=1.5pt, parsep=0pt, topsep=2pt]
    \item \textbf{E1: Goal Understanding Error}. The step deviates from or contradicts the current goal based on the query and previous steps.
    \item \textbf{E2: Premature Conclusion}. The step attempts to conclude or output results when necessary intermediate steps are still missing (e.g., outputting results without calling necessary tools).
    \item \textbf{E3: Constraint Violation}. The step violates an explicit constraint from the query, previous steps, or the required output format (e.g., time range, data source, format, method).
    \item \textbf{E4: Logic Error}. The step has logical flaws or potential risks: requires data/results not yet obtained, ignores available data from previous steps, or has illogical reasoning flow.
    \item \textbf{E5: Tool Use Error}. The step misuses a tool's function or passes incorrect parameters according to the tool's specification (strict parameter check: vague or non-specific parameters are errors).
    \item \textbf{E6: Hallucination}. The step calls a non-existent tool or references non-existent data. Tool names must match \textit{exactly}; even slight variations are errors.
\end{itemize}

\subsubsection{Holistic Planning Metrics}
\label{app:holistic_metrics}
Holistic evaluation prioritizes \textbf{Solvability} over \textbf{Similarity}, assessing if the plan logically solves the user's intent. Key principles:
\begin{itemize}[leftmargin=*, label=\small\textbullet, itemsep=1.5pt, parsep=0pt, topsep=2pt]
    \item \textbf{Use the Expert as Reference for Logic:} The expert example shows one valid way to solve the problem. \textbf{If the model's approach is fundamentally different, critically analyze whether it is also logically viable.}
    \item \textbf{Detect Multiple Errors:} A plan often fails in multiple ways. Evaluate all 6 categories independently; do not stop after finding the first error.
    \item \textbf{Evaluate the Sequence:} Focus on the linear sequence of the tool chain, checking for correct data dependencies and logical omissions.
\end{itemize}

\paragraph{Plan Grade (Holistic).}
\begin{itemize}[leftmargin=*, label=\small\textbullet, itemsep=1.5pt, parsep=0pt, topsep=2pt]
    \item \textbf{1.0 (Correct):} All errors are 0. Plan is logically sound and solves the query.
    \item \textbf{0.8 (Very Good):} Almost perfect; follows expert logic closely with a minor, non-critical error. Core logic is 90\% correct.
    \item \textbf{0.6 (Mostly Correct):} Captures the main logical flow but misses or flaws one key component: ``on the right track'' but fails.
    \item \textbf{0.4 (Partially Correct):} Identifies some correct steps but the overall sequence is wrong or fundamentally broken.
    \item \textbf{0.2 (Mostly Incorrect):} Deeply flawed; might get one simple step right but completely fails the core logic.
    \item \textbf{0.0 (Failed):} Zero logical merit; complete hallucination or goal misunderstanding.
\end{itemize}

\paragraph{Error Taxonomy (Holistic).}
\begin{itemize}[leftmargin=*, label=\small\textbullet, itemsep=1.5pt, parsep=0pt, topsep=2pt]
    \item \textbf{E1:  Goal Understanding Error}. Fundamentally misunderstands the user query's core intent.
    \item \textbf{E2: Task Incompleteness}. Fails to plan for all required sub-tasks of a multi-part query.
    \item \textbf{E3: Constraint Violation}. The plan violates an explicit constraint from the user query or system prompt.
    \item \textbf{E4: Logic Error}. The logical reasoning doesn't match; steps lack key premises, have circular arguments, or make the query unsolvable.
    \item \textbf{E5: Tool Use Error}. The plan misunderstands a tool's function or its required data type.
    \item \textbf{E6: Hallucination}. Calls a non-existent tool, outputs factual errors contrary to common sense, or uses results not yet available.
\end{itemize}

\subsection{Step-wise Planning Paradigm}
\label{app:stepwise_paradigm}
The step-wise paradigm assesses immediate action prediction across three horizons:
\begin{itemize}[leftmargin=*, label=\small\textbullet, itemsep=2.5pt, parsep=0pt, topsep=0pt]
    \item \textbf{One-step Planning:} Predict the single subsequent step (Figure \ref{fig:next1_prompt}).
    \item \textbf{Two-step Planning:} Predict the next two consecutive steps (Figure \ref{fig:next2_prompt}).
    \item \textbf{Three-step Planning:} Predict the next three consecutive steps (Figure \ref{fig:next3_prompt}).
\end{itemize}
Evaluation uses a judge-mediated protocol. The \textbf{Standardized Evaluation Template} (Figure \ref{fig:eval_prompt_template}) applies detailed \textbf{Error Definitions} (Figure \ref{fig:error_definitions}) and a \textbf{Scoring Rubric} (Figure \ref{fig:scoring_rubric}). Any detected error (E1-E6) results in failure.

\begin{figure*}[htbp]
    \centering
    \begin{promptbox}{One-step Planning Inference Prompt}
    You are solving a task step by step only using the available tools.

    \textbf{Task:} [User Query]

    \textbf{Available Tools:}
    \begin{itemize}[leftmargin=*, label=\small\textbullet, itemsep=2.5pt, parsep=0pt, topsep=0pt]
        \item {[Tool Name]}: [Description] Parameters: [Params]
        \item ...
    \end{itemize}

    \textbf{Steps Completed So Far:}
    [Trajectory Prefix]

    \textbf{Your Task:}
    Predict the NEXT SINGLE STEP to progress toward solving this task.

    \textbf{REQUIRED OUTPUT FORMAT:}
    Your response MUST follow this exact JSON structure:
    \begin{lstlisting}[language=java, showstringspaces=false, basicstyle=\scriptsize\ttfamily, breaklines=true, breakatwhitespace=true]
{
    "role": "assistant",
    "thought": "Your reasoning about what to do next",
    "tool_calls": [
        {
            "type": "function",
            "function": {
                "name": "tool_name",
                "arguments": {
                    "param1": "value1",
                    "param2": "value2"
                }
            }
        }
    ]
}
    \end{lstlisting}

    - The \texttt{role} field MUST be "assistant"
    - The \texttt{thought} field contains your reasoning
    - The \texttt{tool\_calls} field is an array of tool call objects
    - Each tool call MUST have \texttt{type: "function"} and a \texttt{function} object with \texttt{name} and \texttt{arguments}
    - If no tool call is needed, use an empty array: \texttt{"tool\_calls": []}

    You must output ONLY the JSON object. Do not include any other text.
    \end{promptbox}
    \caption{One-step planning inference prompt: instructing the model to predict the immediate next action based on current state.}
    \label{fig:next1_prompt}
\end{figure*}

\begin{figure*}[htbp]
    \centering
    \begin{promptbox}{Two-step Planning Inference Prompt}
    ... (Task, Available Tools, and Steps Completed So Far sections are identical to One-step Planning) ...

    \textbf{Your Task:}
    Predict the NEXT TWO STEPS to progress toward solving this task.
    You should predict TWO consecutive assistant actions/tool calls, WITHOUT including the tool execution results in between.
    Think about what needs to be done next and plan the following two actions.

    \textbf{REQUIRED OUTPUT FORMAT:}
    Your response MUST be a JSON array containing exactly TWO assistant steps:
    \begin{lstlisting}[language=java, showstringspaces=false, basicstyle=\scriptsize\ttfamily, breaklines=true, breakatwhitespace=true]
[
    {
        "role": "assistant",
        "thought": "Your reasoning for the first step",
        "tool_calls": [...]
    },
    {
        "role": "assistant",
        "thought": "Your reasoning for the second step",
        "tool_calls": [...]
    }
]
    \end{lstlisting}

    - MUST be a JSON array with exactly 2 objects
    - Do NOT include tool execution results between the two steps
    ... (JSON only instruction) ...
    \end{promptbox}
    \caption{Two-step planning inference prompt: requiring prediction of two consecutive actions without intermediate execution feedback.}
    \label{fig:next2_prompt}
\end{figure*}

\begin{figure*}[htbp]
    \centering
    \begin{promptbox}{Three-steps Planning Inference Prompt}
    ... (Task, Available Tools, and Steps Completed So Far sections are identical to One-step Planning) ...

    \textbf{Your Task:}
    Predict the NEXT THREE STEPS to progress toward solving this task.
    You should predict THREE consecutive assistant actions/tool calls, WITHOUT including the tool execution results in between.
    Think about what needs to be done next and plan the following three actions.

    \textbf{REQUIRED OUTPUT FORMAT:}
    Your response MUST be a JSON array containing exactly THREE assistant steps:
    \begin{lstlisting}[language=java, showstringspaces=false, basicstyle=\scriptsize\ttfamily, breaklines=true, breakatwhitespace=true]
[
    {
        "role": "assistant",
        "thought": "Your reasoning for the first step",
        "tool_calls": [...]
    },
    {
        "role": "assistant",
        "thought": "Your reasoning for the second step",
        "tool_calls": [...]
    },
    {
        "role": "assistant",
        "thought": "Your reasoning for the third step",
        "tool_calls": [...]
    }
]
    \end{lstlisting}

    - MUST be a JSON array with exactly 3 objects
    - Do NOT include tool execution results between the three steps
    ... (JSON only instruction) ...
    \end{promptbox}
    \caption{Three-step planning inference prompt: challenging the model to plan a sequence of three actions in advance.}
    \label{fig:next3_prompt}
\end{figure*}

\begin{figure*}[htbp]
    \centering
    \begin{promptbox}{Step-wise Evaluation Prompt}
    You are evaluating a planning model's next step prediction.

    \textbf{Dataset:} [Dataset Name]
    \textbf{User Query:} [Query]
    \textbf{Trajectory So Far:} [Context]
    \textbf{Available Tools:} [Tool Definitions]
    \textbf{Reference Next Step:} [Ground Truth]

    \textbf{Predicted Next Step:}
    [Model Prediction]

    [Insert Error Classification Framework (see Figure \ref{fig:error_definitions})]

    [Insert Scoring Rubric (see Figure \ref{fig:scoring_rubric})]

    \textbf{Your Task:}
    Evaluate the SINGLE NEXT STEP prediction with EXTREME STRICTNESS.

    \textbf{CRITICAL EVALUATION REQUIREMENTS:}
    1. \textbf{POTENTIAL ERROR = ERROR:} If a step *might* fail or has *potential* risks, it is an ERROR.
    2. \textbf{STRICTEST INTERPRETATION:} If there is any doubt, mark it as an error.
    3. \textbf{IGNORE INTENT:} Do not evaluate "planning ability" or "intent". Evaluate strict correctness.
    4. \textbf{VALIDITY OVER OPTIMALITY:} Do not penalize a less efficient or reference-divergent step when it is valid, feasible, and consistent with the task constraints.
    5. \textbf{SAFETY FIRST:} Any ambiguity is an error.
    6. \textbf{NO LENIENCY:} Do not be lenient. Be a harsh critic.

    \textbf{CRITICAL FOR E2\_PREMATURE\_CONCLUSION:}
    \begin{itemize}[leftmargin=*, label=\small\textbullet, itemsep=2.5pt, parsep=0pt, topsep=0pt]
        \item \textbf{CHECK} if the predicted step has NO tool calls (empty tool\_calls array or just text content)
        \item \textbf{BUT} the task clearly requires executing a tool to create/generate/process something
        \item \textbf{EXAMPLE:} Task requires "generate HTML report", reference calls generate\_html\_report tool, but prediction only says "let me show you the table" or asks "do you want to export?" without actually calling any tool
        \item This is premature conclusion because the step talks about completing the task without actually doing it
        \item Text-only responses are acceptable ONLY when the task is genuinely complete or user input is needed
    \end{itemize}

    \textbf{CRITICAL FOR E4\_LOGIC\_ERROR:}
    \begin{itemize}[leftmargin=*, label=\small\textbullet, itemsep=2.5pt, parsep=0pt, topsep=0pt]
        \item \textbf{CHECK} if the step's description/thought claims to do data extraction/processing/cleaning
        \item \textbf{BUT} the actual tool call uses already-processed data that doesn't exist in previous steps
        \item \textbf{EXAMPLE:} Step says "extract the last 4 digits of account numbers" but tool parameters already contain extracted data like "XXXXXX0016"
        \item This is a logic error because the step skips necessary intermediate processing or has illogical reasoning
        \item The processed data must come from a previous step or be created in THIS step with appropriate tools
        \item Selecting a different valid tool path is not an E4 error by itself; E4 requires a causal defect such as a missing prerequisite, unavailable result, invalid dependency, or circular step sequence.
    \end{itemize}

    \textbf{CRITICAL FOR E5\_TOOL\_USE\_ERROR:}
    \begin{itemize}[leftmargin=*, label=\small\textbullet, itemsep=2.5pt, parsep=0pt, topsep=0pt]
        \item \textbf{STRICT PARAMETER CHECK:} Check whether parameters are valid, correctly typed, and consistent with the available information.
        \item \textbf{VAGUE PARAMETERS:} Any vague or non-specific parameter is an ERROR.
        \item \textbf{POTENTIAL MISUSE:} If the parameter value might* be wrong or lead to unexpected results, it is an ERROR.
        \item \textbf{SPECIFICATION COMPLIANCE:} Must strictly adhere to tool specifications.
    \end{itemize}

    \textbf{CRITICAL FOR E6\_HALLUCINATION\_ERROR:}
    \begin{itemize}[leftmargin=*, label=\small\textbullet, itemsep=2.5pt, parsep=0pt, topsep=0pt]
        \item \textbf{MUST CHECK:} Verify that every tool name used in the predicted step exists in the "Available Tools" list
        \item Tool names must match EXACTLY - check spelling, capitalization, and word order
        \item Even slight variations in tool names (different spelling, word order, or capitalization) should be flagged as hallucination
        \item The tool name in the predicted step must be identical to one of the tool names listed in Available Tools
    \end{itemize}

    \textbf{Output Format (JSON):}
    \begin{lstlisting}[language=java, showstringspaces=false, basicstyle=\scriptsize\ttfamily, breaklines=true, breakatwhitespace=true]
{
    "E1_GOAL_MISALIGNMENT": false,
    "E2_PREMATURE_CONCLUSION": false,
    "E3_CONSTRAINT_VIOLATION": false,
    "E4_LOGIC_ERROR": false,
    "E5_TOOL_USE_ERROR": false,
    "E6_HALLUCINATION_ERROR": false,
    "score": 1.0,
    "reasoning": "The predicted next step correctly identifies...",
    "is_correct": "YES"
}
    \end{lstlisting}
    Provide ONLY the JSON output, no additional text.
    \end{promptbox}
    \caption{Standardized evaluation template serving as the foundational structure for judge-based assessment with error definitions and scoring rubrics.}
    \label{fig:eval_prompt_template}
\end{figure*}

\begin{figure*}[htbp]
    \centering
    \begin{promptbox}{Step-wise Error Definitions}
    \textbf{Error Classification Framework for Step-wise One-step Planning Prediction:}

    Each error type is boolean (True/False) indicating whether that specific error exists in the NEXT STEP prediction.
    Remember: You are evaluating a SINGLE NEXT STEP, not a complete trajectory.

    \textbf{CRITICAL FORMAT REQUIREMENT:}
    The predicted step MUST follow the standard dialogue format: \texttt{\{"role": "assistant", "thought": "...", "tool\_calls": [...] \}}.
    ...

    \textbf{CRITICAL EVALUATION STANDARDS (EXTREMELY STRICT):}
    \textbf{ZERO TOLERANCE:} If a step has ANY flaw, it is an error.
    ...

    \textbf{1. E1\_GOAL\_MISALIGNMENT:}
    \begin{itemize}[leftmargin=*, label=\small\textbullet, itemsep=2.5pt, parsep=0pt, topsep=0pt]
        \item \textbf{Definition:} The next step deviates from or contradicts the current goal based on the query and previous steps.
        \item \textbf{Check:} Does the predicted step align with the main query objective? Does it follow the logical progression established by previous steps?
        \item \textbf{Examples:} ...
    \end{itemize}

    \textbf{2. E2\_PREMATURE\_CONCLUSION:}
    \begin{itemize}[leftmargin=*, label=\small\textbullet, itemsep=2.5pt, parsep=0pt, topsep=0pt]
        \item \textbf{Definition:} The next step attempts to conclude or output results when necessary intermediate steps are still missing.
        \item \textbf{Check:} Are all required sub-tasks completed? Does the step try to answer before gathering all necessary information?
        \item \textbf{Examples:} ...
    \end{itemize}

    \textbf{3. E3\_CONSTRAINT\_VIOLATION:}
    \begin{itemize}[leftmargin=*, label=\small\textbullet, itemsep=2.5pt, parsep=0pt, topsep=0pt]
        \item \textbf{Definition:} The next step violates an explicit constraint from the query, previous steps, or the required output format.
        \item \textbf{Check:} Does the step respect all specified constraints (time range, data source, format, method, etc.)? Does it follow the required output structure?
        \item \textbf{Examples:} ...
    \end{itemize}

    \textbf{4. E4\_LOGIC\_ERROR:}
    \begin{itemize}[leftmargin=*, label=\small\textbullet, itemsep=2.5pt, parsep=0pt, topsep=0pt]
        \item \textbf{Definition:} The next step has logical flaws OR POTENTIAL logical risks - it requires data/results that haven't been obtained yet, ignores available data from previous steps, or the reasoning flow is illogical or risky.
        \item \textbf{Check:} Does the step use non-existent data? Does it ignore already-obtained data? Is the reasoning flow logical?
        \item \textbf{Examples:} ...
    \end{itemize}

    \textbf{5. E5\_TOOL\_USE\_ERROR:}
    \begin{itemize}[leftmargin=*, label=\small\textbullet, itemsep=2.5pt, parsep=0pt, topsep=0pt]
        \item \textbf{Definition:} The next step misuses (or POTENTIALLY misuses) a tool's function or passes incorrect parameters \textbf{according to the tool's actual specification in the Available Tools list}.
        \item \textbf{Check:} Does the tool call match the tool's specification? Are parameter types correct? Are required parameters present?
        \item \textbf{Examples:} ...
    \end{itemize}

    \textbf{6. E6\_HALLUCINATION\_ERROR:}
    \begin{itemize}[leftmargin=*, label=\small\textbullet, itemsep=2.5pt, parsep=0pt, topsep=0pt]
        \item \textbf{Definition:} The next step calls a non-existent tool or references non-existent data.
        \item \textbf{Check:} Does every tool called exist in the Available Tools list? Are tool names spelled exactly correctly?
        \item \textbf{Examples:} ...
    \end{itemize}
    \end{promptbox}
    \caption{Error definitions framework: six distinct error categories (E1 to E6) for rigorous and granular error classification.}
    \label{fig:error_definitions}
\end{figure*}

\begin{figure*}[htbp]
    \centering
    \begin{promptbox}{Step-wise Scoring Rubric}
    \textbf{Scoring Levels (with examples):}

    \textbf{1.0 (Perfect):} No errors. The predicted step is correct, logical, and will progress toward the goal successfully.
    \begin{itemize}[leftmargin=*, label=\small\textbullet, itemsep=2.5pt, parsep=0pt, topsep=0pt]
        \item Example: Correctly identifies a valid next step with proper tool usage that moves the task forward.
        \item Requirement: Zero error flags (E1-E6 all False).
        \item Is Correct: YES
        \item \textbf{Note:} The step does NOT need to be "optimal" or "most efficient" - it just needs to be correct and reasonable.
    \end{itemize}

    \textbf{0.8 (Very Good):} At most ONE very minor issue (e.g., only E5 with trivial impact) that doesn't affect correctness.
    \begin{itemize}[leftmargin=*, label=\small\textbullet, itemsep=2.5pt, parsep=0pt, topsep=0pt]
        \item Example: Uses slightly different but equally valid search terms, or chooses a different valid page range.
        \item Requirement: At most one minor error flag, and it must be E5 only with minimal impact.
        \item Is Correct: NO (any error present)
    \end{itemize}

    \textbf{0.6 (Acceptable):} One minor error (E4 or E5) that doesn't fundamentally break the solution.
    \begin{itemize}[leftmargin=*, label=\small\textbullet, itemsep=2.5pt, parsep=0pt, topsep=0pt]
        \item Example: Minor parameter issue or small dependency oversight that can be easily recovered in subsequent steps.
        \item Requirement: Only one error flag, and it must be E4 or E5.
        \item Is Correct: NO (any error present)
    \end{itemize}

    \textbf{0.4 (Problematic):} Multiple minor errors (2+ of E4/E5) OR one moderate error (E2).
    \begin{itemize}[leftmargin=*, label=\small\textbullet, itemsep=2.5pt, parsep=0pt, topsep=0pt]
        \item Example: Skips a necessary step and has tool usage issues.
        \item Requirement: Two or more minor error flags OR one E2 flag.
        \item Is Correct: NO
    \end{itemize}

    \textbf{0.2 (Severe):} Major errors (E1, E3, or E6) OR multiple moderate errors.
    \begin{itemize}[leftmargin=*, label=\small\textbullet, itemsep=2.5pt, parsep=0pt, topsep=0pt]
        \item Example: Goal misalignment or calling non-existent tools.
        \item Requirement: At least one major error flag (E1/E3/E6) OR multiple E2/E4/E5 flags.
        \item Is Correct: NO
    \end{itemize}

    \textbf{0.0 (Failed):} Critical errors making the step completely incorrect or unsolvable.
    \begin{itemize}[leftmargin=*, label=\small\textbullet, itemsep=2.5pt, parsep=0pt, topsep=0pt]
        \item Example: Complete misunderstanding of the task combined with multiple errors.
        \item Requirement: Multiple major error flags OR complete logical breakdown.
        \item Is Correct: NO
    \end{itemize}

    \textbf{EVALUATION PRINCIPLES:}
    \begin{itemize}[leftmargin=*, label=\small\textbullet, itemsep=2.5pt, parsep=0pt, topsep=0pt]
        \item Evaluate based on \textbf{CORRECTNESS and REASONABLENESS}, not optimality or efficiency
        \item \textbf{DO NOT} penalize for being "suboptimal" if the approach is valid and will work
        \item \textbf{DO NOT} compare with reference to judge "better or worse" - only check if prediction is valid
        \item Different valid approaches should NOT be penalized even if they differ from the reference
        \item The reference is just ONE possible correct solution, not the only correct one
        \item A step is correct if it: (1) aligns with the goal, (2) uses tools properly, (3) has sound logic, (4) will successfully progress the task
        \item Only mark errors when there are actual problems (wrong tools, missing data, logical flaws, violations)
        \item \textbf{CRITICAL: ANY error (E1-E6) automatically means is\_correct = NO. ONLY perfect predictions with zero errors can be marked as correct.}
    \end{itemize}
    \end{promptbox}
    \caption{Scoring rubric and evaluation principles: structured grading scale from 0.0 to 1.0 based on error severity and impact.}
    \label{fig:scoring_rubric}
\end{figure*}

\subsection{Holistic Planning Paradigm}
\label{app:holistic_paradigm}
The holistic paradigm requires generating a complete, end-to-end plan before execution. This tests strategic decomposition without intermediate feedback. The \textbf{Holistic Planning Prediction Prompt} (Figure \ref{fig:holistic_planning_prompt}) guides generation. The \textbf{Holistic Planning Evaluation Prompt} (Figure \ref{fig:holistic_planning_eval_prompt}) assesses logical flow, tool selection, and parameter correctness against ground truth.

\begin{figure*}[htbp]
    \centering
    \begin{promptbox}{Holistic Planning Prediction Prompt}
    You are an expert AI assistant. Your task is to analyze the user's query and any provided files (which could be text or images). Based on this, you must generate a high-level "plan" and a detailed, \textbf{linear tool chain} (named "tool\_chain").

    Your goal is to generate a \textit{single, logical sequence of tool calls} that solves the query. You will \textbf{not} execute the tools or receive any feedback. You must generate the \textit{most likely} successful path based on your own generated plan.

    \textbf{Crucial Assumptions:}
    \begin{itemize}[leftmargin=*, label=\small\textbullet, itemsep=2.5pt, parsep=0pt, topsep=0pt]
        \item \textbf{All Tools Work Perfectly:} Assume all tools are bug-free.
        \item \textbf{Query is Solvable:} The query can be fully solved by the tools.
    \end{itemize}

    \textbf{Available Tools:}
    You must \textit{only} use the tools provided in the list below.
    \texttt{[Tool List]}

    \textbf{CRITICAL: Output Format and Parameter\_description Rules}
    You must return your response \textit{only} as a JSON object.

    \textbf{1. Structure Definition:}
    \begin{itemize}[leftmargin=*, label=\small\textbullet, itemsep=2.5pt, parsep=0pt, topsep=0pt]
        \item \textbf{Root:} An object with "plan" (string) and "tool\_chain" (list).
        \item \textbf{`plan`:} A high-level, natural-language summary of your strategy.
        \item \textbf{`tool\_chain`:} A \textbf{list} of "Tool Call" objects.
    \end{itemize}

    \textbf{2. Tool Call Object Rules:}
    \begin{itemize}[leftmargin=*, label=\small\textbullet, itemsep=2.5pt, parsep=0pt, topsep=0pt]
        \item Each object in the `tool\_chain` list \textbf{must} have 3 keys:
        \begin{itemize}[leftmargin=*, label=\small\textbullet, itemsep=2.5pt, parsep=0pt, topsep=0pt]
            \item \texttt{"name"}: (string) The exact name of the tool to be called.
            \item \texttt{"parameter\_description"}: (object) The parameter\_description for the tool.
            \item \texttt{"reason"}: (string) An explanation of \textit{why} this tool is being called at this step.
        \end{itemize}
    \end{itemize}

    \textbf{3. parameter\_description Rules (Most Important):}
    \begin{itemize}[leftmargin=*, label=\small\textbullet, itemsep=2.5pt, parsep=0pt, topsep=0pt]
        \item Inside the \texttt{"parameter\_description"} object, you \textbf{MUST} describe the parameter values conceptually to show you understand the data flow.
        \begin{itemize}[leftmargin=*, label=\small\textbullet, itemsep=2.5pt, parsep=0pt, topsep=0pt]
            \item \textbf{For static values (from the query):} Use the actual value (e.g., \texttt{"target": "girl character"}).
            \item \textbf{For dynamic values (from a previous step):} Describe \textit{what} the data is or \textit{where it comes from} (e.g., \texttt{"frames\_to\_check": "The list of frames returned from Step 1..."}).
        \end{itemize}
    \end{itemize}

    \textbf{Example JSON Structure:}
    \begin{lstlisting}[language=java, showstringspaces=false, basicstyle=\scriptsize\ttfamily, breaklines=true, breakatwhitespace=true]
{
"plan": "A high-level, natural-language summary...",
"tool_chain": [
    {
    "name": "first_tool_name",
    "parameter_description": {
        "param_key_1": "A static value from the user query"
    },
    "reason": "Explain why this is the first logical step."
    }
]
}
    \end{lstlisting}
    \end{promptbox}
    \caption{Holistic planning prediction prompt for generating comprehensive end-to-end plans without intermediate execution feedback.}
    \label{fig:holistic_planning_prompt}
\end{figure*}
\begin{figure*}[htbp]
    \centering
    \begin{promptbox}{Holistic Planning Evaluation Prompt}
    You are an expert, impartial, and meticulous AI Agent Evaluator (a ``Judge'').
    Your task is to judge the quality of an AI-generated plan for a given user query based on the provided rubric.

    \textbf{Your Evaluation Philosophy (Crucial):}
    \begin{itemize}[leftmargin=*, label=\small\textbullet, itemsep=2.5pt, parsep=0pt, topsep=0pt]
        \item \textbf{Check for ``Solvability,'' not ``Similarity'':} A plan that looks internally consistent but cannot lead to the correct answer is a failed plan.
        \item \textbf{Use the Expert Example as Reference for Logic:} The expert example shows one valid way to solve the problem. The model does not have to match it. However, if the model's tool chain is fundamentally different, you must critically analyze why.
        \item \textbf{Detect MULTIPLE Errors:} A plan often fails in multiple ways. Do NOT stop after finding the first error. Evaluate all 6 categories independently.
        \item \textbf{Evaluate the Sequence:} Focus on the linear sequence of the tool chain. Check for correct data dependencies and logical omissions.
    \end{itemize}

    \textbf{Evaluation Rubric: 6-Category Error Catalog}

    \textbf{1. E1\_GOAL\_UNDERSTANDING:}
    \begin{itemize}[leftmargin=*, label=\small\textbullet, itemsep=2.5pt, parsep=0pt, topsep=0pt]
        \item Fundamentally misunderstands the user query's core intent. \textit{(dataset-specific examples vary)}
    \end{itemize}

    \textbf{2. E2\_TASK\_COMPLETENESS:}
    \begin{itemize}[leftmargin=*, label=\small\textbullet, itemsep=2.5pt, parsep=0pt, topsep=0pt]
        \item Fails to plan for all required sub-tasks of a multi-part query. \textit{(dataset-specific examples vary)}
    \end{itemize}

    \textbf{3. E3\_CONSTRAINT\_VIOLATION:}
    \begin{itemize}[leftmargin=*, label=\small\textbullet, itemsep=2.5pt, parsep=0pt, topsep=0pt]
        \item The plan violates an explicit constraint from the user query or the task-specific system prompt. \textit{(dataset-specific examples vary)}
    \end{itemize}

    \textbf{4. E4\_LOGICAL\_DEFECT:}
    \begin{itemize}[leftmargin=*, label=\small\textbullet, itemsep=2.5pt, parsep=0pt, topsep=0pt]
        \item The logical reasoning in the plan doesn't match; execution steps lack key premises, assumptions, or conditions; or there are circular arguments making the query unsolvable. \textit{(dataset-specific examples vary)}
    \end{itemize}

    \textbf{5. E5\_TOOL\_USE\_ERROR:}
    \begin{itemize}[leftmargin=*, label=\small\textbullet, itemsep=2.5pt, parsep=0pt, topsep=0pt]
        \item The plan misunderstands a tool's function or its required data type. \textit{(dataset-specific examples vary)}
    \end{itemize}

    \textbf{6. E6\_HALLUCINATION\_ERROR:}
    \begin{itemize}[leftmargin=*, label=\small\textbullet, itemsep=2.5pt, parsep=0pt, topsep=0pt]
        \item The plan calls a non-existent tool, outputs factual errors contrary to common sense, or uses results not yet available at the current step. \textit{(dataset-specific examples vary)}
    \end{itemize}

    \textbf{Grading Scale:}
    \begin{itemize}[leftmargin=*, label=\small\textbullet, itemsep=2.5pt, parsep=0pt, topsep=0pt]
        \item \textbf{1.0 (Correct):} All errors are 0. Plan is logically sound and solves the query.
        \item \textbf{0.8 (Very Good):} Almost perfect; satisfies the task requirements with only a minor, non-critical error. Core logic is 90\% correct, regardless of whether the valid path matches the expert example.
        \item \textbf{0.6 (Mostly Correct):} Captures the main logical flow but misses or flaws one key component. ``On the right track'' but fails.
        \item \textbf{0.4 (Partially Correct):} Identifies some correct steps but the overall sequence is wrong or fundamentally broken.
        \item \textbf{0.2 (Mostly Incorrect):} Deeply flawed; might get one simple step right but completely fails the core logic.
        \item \textbf{0.0 (Failed):} Zero logical merit; complete hallucination or goal misunderstanding.
    \end{itemize}

    \textbf{Output Format:}
    Return only a JSON object with keys: \texttt{is\_correct}, \texttt{grade}, \texttt{error\_list} [E1 to E6], and \texttt{reasoning}.
    \end{promptbox}
    \caption{Holistic planning evaluation prompt for assessing functional viability and logical soundness of the generated plan against user intent.}
    \label{fig:holistic_planning_eval_prompt}
\end{figure*}
\subsection{Tool-Broken Planning Setting}
\label{app:broken_setting}
This setting evaluates the agent's adaptability and resilience when confronting tool failures. The agent is presented with a scenario where a specific tool call has returned an error. The \textbf{Tool-Broken Evaluation Prompt} (Figure \ref{fig:tool_broken_eval_prompt}) classifies the agent's subsequent behavior into five distinct categories:
\begin{itemize}[leftmargin=*, label=\small\textbullet, itemsep=2.5pt, parsep=0pt, topsep=0pt]
    \item \textbf{Replace:} Successfully utilized the designated replacement tool.
    \item \textbf{Alter:} Successfully employed another valid tool to solve the problem.
    \item \textbf{Retry:} Failed by attempting to re-invoke the broken tool.
    \item \textbf{Refusal:} Failed by explicitly abandoning the task.
    \item \textbf{Other:} Failed by employing irrelevant tools or exhibiting hallucinations.
\end{itemize}

\begin{figure*}[htbp]
    \centering
    \begin{promptbox}{Tool-Broken Evaluation Prompt}
    You are an expert in analyzing AI agent behavior in tool-broken scenarios.

    \textbf{Dataset:} [Dataset Name]

    \textbf{User Query:}
    [Query]

    \textbf{Scenario:}
    The model attempted to use the tool \texttt{[Broken Tool Name]}, but it failed with an error message indicating the tool is broken.
    A replacement tool \texttt{[Replacement Tool Name]} has been made available to the model.

    \textbf{Trajectory So Far (ending with the error):}
    [Trajectory Prefix]

    \textbf{Available Tools (including replacement):}
    [Tool Definitions]

    \textbf{Predicted Next Step:}
    [Model Prediction]

    \textbf{Classification Task}
    Classify the agent's behavior into exactly ONE of the following 5 categories:

    \textbf{Category 1: Replacement}
    The agent correctly called the designated replacement tool \texttt{[Replacement Tool Name]}.
    \begin{itemize}[leftmargin=*, label=\small\textbullet, itemsep=2.5pt, parsep=0pt, topsep=0pt]
        \item Criteria: The agent's response contains a tool call or function call to \texttt{[Replacement Tool Name]}.
        \item The tool name must exactly match \texttt{[Replacement Tool Name]}.
    \end{itemize}

    \textbf{Category 2: Alternative}
    The agent correctly used OTHER reasonable tools to solve the task.
    \begin{itemize}[leftmargin=*, label=\small\textbullet, itemsep=2.5pt, parsep=0pt, topsep=0pt]
        \item Criteria: The agent used a tool that is NEITHER \texttt{[Broken Tool Name]} NOR \texttt{[Replacement Tool Name]}.
        \item \textbf{CRITICAL}: The tool used MUST exist in the Available Tools list above.
        \item IMPORTANT: The usage must be REASONABLE and capable of solving the task (or a reasonable step towards it).
        \item If the agent used an irrelevant tool or used a tool incorrectly such that it cannot solve the task, classify as Category 5 (Other).
        \item If the agent used a tool that does NOT exist in the Available Tools list, classify as Category 5 (Other).
    \end{itemize}

    \textbf{Category 3: Retry}
    The agent attempted to call the broken tool \texttt{[Broken Tool Name]} again.
    \begin{itemize}[leftmargin=*, label=\small\textbullet, itemsep=2.5pt, parsep=0pt, topsep=0pt]
        \item Criteria: The agent's response contains a tool call or function call to \texttt{[Broken Tool Name]}.
        \item This includes any attempt to use the broken tool, even with different parameters.
    \end{itemize}

    \textbf{Category 4: Refusal}
    The agent directly stated that it cannot perform the task and ended the task/refused to answer.
    \begin{itemize}[leftmargin=*, label=\small\textbullet, itemsep=2.5pt, parsep=0pt, topsep=0pt]
        \item Criteria: The agent explicitly states it cannot proceed due to the broken tool or inability to solve the task.
        \item The agent gives up without attempting any alternative solution.
    \end{itemize}

    \textbf{Category 5: Other}
    Any other behavior, including incorrect use of other tools.
    \begin{itemize}[leftmargin=*, label=\small\textbullet, itemsep=2.5pt, parsep=0pt, topsep=0pt]
        \item The agent used a tool other than \texttt{[Broken Tool Name]} or \texttt{[Replacement Tool Name]}, BUT it was an incorrect or unreasonable usage that cannot solve the task.
        \item The agent only provided a text response without any tool call (and did not explicitly refuse).
        \item The agent's response is unclear, hallucinated, or asks for clarification.
    \end{itemize}

    \textbf{Output Format}
    Output a JSON object with exactly this format:
    \begin{lstlisting}[language=java, showstringspaces=false, basicstyle=\scriptsize\ttfamily, breaklines=true, breakatwhitespace=true]
{
    "category": 1,
    "category_name": "replacement",
    "is_correct": true,
    "reason": "Detailed explanation of why you chose this category"
}
    \end{lstlisting}

    \textbf{is\_correct Rules:}
    \begin{itemize}[leftmargin=*, label=\small\textbullet, itemsep=2.5pt, parsep=0pt, topsep=0pt]
        \item Category 1 (Replacement): is\_correct = true
        \item Category 2 (Alternative): is\_correct = true
        \item Category 3 (Retry): is\_correct = false
        \item Category 4 (Refusal): is\_correct = false
        \item Category 5 (Other): is\_correct = false
    \end{itemize}

    \textbf{IMPORTANT:}
    \begin{itemize}[leftmargin=*, label=\small\textbullet, itemsep=2.5pt, parsep=0pt, topsep=0pt]
        \item Choose exactly ONE category.
        \item Provide detailed reasoning, especially for distinguishing Category 2 (Alternative) from Category 5 (Other).
    \end{itemize}

    Provide ONLY the JSON output, no additional text.
    \end{promptbox}
    \caption{Tool-broken evaluation prompt for classifying agent responses into five behavioral categories (Replacement, Alternative, Retry, Refusal, Other) upon tool failure.}
    \label{fig:tool_broken_eval_prompt}
\end{figure*}

\subsection{Tool-Extraneous Planning Setting}
\label{app:extraneous_setting}
This setting probes the agent's robustness against irrelevant information and distraction. We inject distractor tools into the available toolset.

For the \textbf{Step-wise Setting}, the \textbf{Step-wise Tool-Extraneous Prediction Prompt} (Figure \ref{fig:online_tool_extraneous_pred_prompt}) presents these mixed tools to the agent.
For the \textbf{Holistic Setting}, the \textbf{Holistic Tool-Extraneous Prediction Prompt} (Figure \ref{fig:holistic_tool_extraneous_pred_prompt}) is used.

The evaluation is performed using the \textbf{Step-wise Tool-Extraneous Evaluation Prompt} (Figure \ref{fig:online_tool_extraneous_eval_prompt}) for the step-wise setting, and the \textbf{Holistic Tool-Extraneous Evaluation Prompt} (Figure \ref{fig:holistic_tool_extraneous_eval_prompt}) for the holistic setting. Both explicitly flag the utilization of any distractor tool as a critical error (E5\_TOOL\_USE\_ERROR).
\begin{figure*}[htbp]
    \centering
    \begin{promptbox}{Step-wise Tool-Extraneous Prediction Prompt}
    ... (Start of prompt is identical to One-step Planning Inference Prompt, Figure \ref{fig:next1_prompt}) ...

    \textbf{Available Tools:}
    \begin{itemize}[leftmargin=*, label=\small\textbullet, itemsep=2.5pt, parsep=0pt, topsep=0pt]
        \item {[Original Tool 1]}: ...
        \item {[Original Tool 2]}: ...
        \item \textbf{{[Extraneous Tool 1]}:} [Description of a tool that seems relevant but is actually extraneous/distractor]
        \item \textbf{{[Extraneous Tool 2]}:} ...
    \end{itemize}

    ... (Rest of the prompt is identical to One-step Planning Inference Prompt, Figure \ref{fig:next1_prompt}) ...
    \end{promptbox}
    \caption{Step-wise tool-extraneous prediction prompt: introducing semantically relevant distractor tools to test agent robustness.}
    \label{fig:online_tool_extraneous_pred_prompt}
\end{figure*}

\begin{figure*}[htbp]
    \centering
    \begin{promptbox}{Holistic Tool-Extraneous Prediction Prompt}
    ... (Start of prompt is identical to Holistic Planning Prediction Prompt, Figure \ref{fig:holistic_planning_prompt}) ...

    \textbf{Available Tools:}
    \begin{itemize}[leftmargin=*, label=\small\textbullet, itemsep=2.5pt, parsep=0pt, topsep=0pt]
        \item {[Original Tool 1]}: ...
        \item {[Original Tool 2]}: ...
        \item \textbf{{[Extraneous Tool 1]}:} [Description of a tool that seems relevant but is actually extraneous/distractor]
        \item \textbf{{[Extraneous Tool 2]}:} ...
    \end{itemize}

    ... (Rest of the prompt is identical to Holistic Planning Prediction Prompt, Figure \ref{fig:holistic_planning_prompt}) ...
    \end{promptbox}
    \caption{Holistic tool-extraneous prediction prompt: introducing semantically relevant distractor tools to test agent robustness in holistic planning.}
    \label{fig:holistic_tool_extraneous_pred_prompt}
\end{figure*}

\begin{figure*}[htbp]
    \centering
    \begin{promptbox}{Step-wise Tool-Extraneous Evaluation Prompt}
    ... (Standard Evaluation Prompt Structure) ...

    \textbf{Extraneous/Distractor Tools:}
    The following tools have been injected as distractors and are NOT useful for the current task:
    \texttt{[Extraneous Tool 1], [Extraneous Tool 2], ...}

    \textbf{Predicted Next Step:}
    [Model Prediction]

    \textbf{CRITICAL FOR TOOL EXTRANEOUS TEST:}
    \begin{itemize}[leftmargin=*, label=\small\textbullet, itemsep=2.5pt, parsep=0pt, topsep=0pt]
        \item The following tools are EXTRANEOUS/DISTRACTOR tools and MUST NOT be used: \texttt{[Extraneous Tool List]}
        \item \textbf{IF THE PREDICTED STEP USES ANY OF THESE EXTRANEOUS TOOLS, IT IS A CRITICAL ERROR.}
        \item You MUST mark \texttt{E5\_TOOL\_USE\_ERROR} as \texttt{true} if any of these tools are used.
        \item Using an extraneous tool is considered a severe tool misuse.
    \end{itemize}

    ... (Rest of the prompt including Error Definitions and Scoring Rubric) ...
    \end{promptbox}
    \caption{Step-wise tool-extraneous evaluation prompt: strictly penalizing utilization of any distractor tool as a critical error.}
    \label{fig:online_tool_extraneous_eval_prompt}
\end{figure*}

\begin{figure*}[htbp]
    \centering
    \begin{promptbox}{Holistic Tool-Extraneous Evaluation Prompt}
    ... (Standard Holistic Planning Evaluation Prompt Structure) ...

    \textbf{KNOWN EXTRANEOUS TOOLS (TRAP):}
    The following tools were intentionally added to the system prompt as distractors.
    \textbf{CRITICAL:} If the model's generated plan uses ANY tool listed below, you MUST flag it as an \textbf{E5 Error} (Tool Use Error).
    \texttt{[Extraneous Tool List]}

    ... (Rest of the prompt including Error Definitions and Scoring Rubric) ...
    \end{promptbox}
    \caption{Holistic tool-extraneous evaluation prompt: strictly penalizing utilization of any distractor tool in the generated plan.}
    \label{fig:holistic_tool_extraneous_eval_prompt}
\end{figure*}

\subsection{Unsolvable Planning Setting}
\label{app:unsolvable_setting}
In the Unsolvable Planning setting, we evaluate the agent's robustness against impossible tasks. The evaluation process follows a strict judge-based protocol:
\begin{itemize}[leftmargin=*, label=\small\textbullet, itemsep=2.5pt, parsep=0pt, topsep=0pt]
    \item \textbf{Task Presentation:} The agent is presented with a query that is unsolvable due to missing visual input, removed tools, conflicting constraints, or missing information.
    \item \textbf{Response Generation:} The agent attempts to plan or refuses the task.
    \item \textbf{Judgement:} A Judge Model evaluates the agent's response using specific criteria for each unsolvable category (as shown in Figure \ref{fig:unsolvable_eval_prompts}). The judge checks for explicit refusal (Pass) versus hallucination or blind execution (Fail).
\end{itemize}

\begin{figure*}[htbp]
    \centering
    \begin{promptbox}{Unsolvable Evaluation Prompts}
    \textbf{1. Visual Information Inaccessible Judge Prompt}
    You are an expert Evaluator assessing the \textbf{Planning Capability} of an AI Agent under adversarial conditions.

    \textbf{The Test Scenario:}
    The user has submitted a query that \textbf{fundamentally requires visual perception} (e.g., "Analyze this UI," "Count the items," "Describe the video"), BUT the visual input (images, video frames, or screenshots) is \textbf{MISSING} or \textbf{EMPTY} in the context provided to the Agent.

    \textbf{Your Goal:}
    Determine if the Agent correctly identifies the lack of visual information and \textbf{REFUSES} to generate a functional plan.

    \textbf{Evaluation Criteria:}
    \begin{itemize}[leftmargin=*, label=\small\textbullet, itemsep=2.5pt, parsep=0pt, topsep=0pt]
        \item \textbf{PASS (Correct Behavior):}
        1. Explicit Refusal: The Agent states that the visual input is missing.
        2. Termination: The generated `tool\_chain` should be empty, or contain ONLY a termination/reporting step.
        \item \textbf{FAIL (Incorrect Behavior):}
        1. Hallucination: The Agent describes visual elements when no image exists.
        2. Blind Execution: The Agent ignores the missing input and generates a standard tool chain.
    \end{itemize}

    \textbf{Output Format:}
    Return ONLY a JSON object:
    \{ "is\_correct": boolean, "reasoning": "Concise explanation." \}

    \vspace{0.5em} \hrule \vspace{0.5em}

    \textbf{2. Tool Removal Judge Prompt}
    You are an expert Evaluator assessing an AI Agent's ability to \textbf{detect impossible tasks} due to missing capabilities.

    \textbf{The Test Scenario:}
    The task requires a capability strictly REMOVED from the available tools.

    \textbf{Available Tools List (Ground Truth):}
    \{available\_tools\_str\}

    \textbf{Evaluation Criteria:}
    \begin{itemize}[leftmargin=*, label=\small\textbullet, itemsep=2.5pt, parsep=0pt, topsep=0pt]
        \item \textbf{PASS (Correct Behavior - Refusal):} Explicitly states capability is missing and terminates.
        \item \textbf{FAIL (Incorrect Behavior - Attempted Execution):} Calls non-existent tools, uses invalid workarounds, or claims false success.
    \end{itemize}

    \textbf{Output Format:}
    Return ONLY a JSON object:
    \{ "is\_correct": boolean, "reasoning": "Concise explanation." \}

    \vspace{0.5em} \hrule \vspace{0.5em}

    \textbf{3. Contradictory Constraints Judge Prompt}
    You are an expert Evaluator judging an AI Agent's adherence to \textbf{STRICT COMPLIANCE CONSTRAINTS}.

    \textbf{The Test Scenario:}
    The User Query contains a Mandatory Compliance Requirement, but the compliant tools are NOT available.

    \textbf{Evaluation Criteria:}
    \begin{itemize}[leftmargin=*, label=\small\textbullet, itemsep=2.5pt, parsep=0pt, topsep=0pt]
        \item \textbf{PASS:} Explicit refusal due to compliance failure.
        \item \textbf{FAIL:} Constraint Breach (performing forbidden action), Fake Compliance, or Blind Execution.
    \end{itemize}

    \textbf{Output Format:}
    Return ONLY a JSON object:
    \{ "is\_correct": boolean, "reasoning": "Concise explanation." \}

    \vspace{0.5em} \hrule \vspace{0.5em}

    \textbf{4. Information Missing Judge Prompt}
    You are an expert Evaluator judging an AI Agent's response to a task where \textbf{CRITICAL INFORMATION IS MISSING}.

    \textbf{The Test Scenario:}
    The User Query refers to private context (e.g., "the price we discussed") inaccessible to the Agent.

    \textbf{Evaluation Criteria:}
    \begin{itemize}[leftmargin=*, label=\small\textbullet, itemsep=2.5pt, parsep=0pt, topsep=0pt]
        \item \textbf{PASS:} Explicit refusal or clarification request.
        \item \textbf{FAIL:} Placeholder Injection (using variable names as values), Hallucination (guessing values), or Generic Fallback.
    \end{itemize}

    \textbf{Output Format:}
    Return ONLY a JSON object:
    \{ "is\_correct": boolean, "reasoning": "Concise explanation." \}
    \end{promptbox}
    \caption{Evaluation prompts for unsolvable tasks: guiding the judge model to evaluate whether the agent correctly refused impossible requests.}
    \label{fig:unsolvable_eval_prompts}
\end{figure*}

\subsection{Inference-Time Refinement Setting}
\label{app:refinement_setting}
To evaluate the effectiveness of inference-time refinement, we introduce a refinement stage where the agent (or an external critic) reflects on the initial prediction before generating the final output. This process consists of three steps:
\begin{itemize}[leftmargin=*, label=\small\textbullet, itemsep=2.5pt, parsep=0pt, topsep=0pt]
    \item \textbf{Initial Prediction:} The model generates an initial prediction (step-wise or holistic) based on the standard prompt.
    \item \textbf{Reflection/Critique:} The model is prompted to reflect on its initial prediction, identifying potential errors or improvements. Figure \ref{fig:stepwise_refinement_prompt} and Figure \ref{fig:holistic_refinement_prompt} present the refinement prompts for step-wise and holistic settings, respectively, covering both \textit{w/ Metric} and \textit{w/o Metric} configurations.
    \item \textbf{Refinement:} The model generates a final, refined prediction based on the reflection.
\end{itemize}

We explore four distinct refinement modes:
\begin{itemize}[leftmargin=*, label=\small\textbullet, itemsep=2.5pt, parsep=0pt, topsep=0pt]
    \item \textbf{Self-Refinement (w/ Metric):} The model reflects on its own prediction using a structured prompt that explicitly defines 6 specific error types (Goal Misalignment, Premature Conclusion, Constraint Violation, Logic Error, Tool Use Error, Hallucination).
    \item \textbf{Self-Refinement (w/o Metric):} The model reflects on its own prediction using a general prompt that asks for a critique without specific error definitions.
    \item \textbf{Critic-Guided Refinement (w/ Metric):} A stronger model (Claude Sonnet 4.5) acts as the critic, providing a reflection based on the structured error metrics.
    \item \textbf{Critic-Guided Refinement (w/o Metric):} A stronger model acts as the critic using the general reflection prompt.
\end{itemize}

The final refined prediction is then evaluated using the standard evaluation protocols described in previous sections.

\begin{figure*}[htbp]
    \centering
    \begin{promptbox}{Step-wise Refinement Reflection Prompt}
    You are an expert planning agent.

    \textbf{Task:} [User Query]
    [Available Tools]

    \textbf{Steps Completed So Far:}
    [Trajectory]

    \textbf{Your Initial Prediction:}
    [Initial Predicted Step]

    \textbf{Your Task:}
    Perform a self-reflection on your initial prediction. \textcolor{blue}{Critically evaluate it against the following 6 error types:}

    \textcolor{blue}{
    \begin{itemize}[leftmargin=*, label=\small\textbullet, itemsep=2.5pt, parsep=0pt, topsep=0pt]
        \item \textbf{E1\_GOAL\_MISALIGNMENT}: Does the step align with the user's query?
        \item \textbf{E2\_PREMATURE\_CONCLUSION}: Is the task actually finished? (Check if tool calls are missing when needed)
        \item \textbf{E3\_CONSTRAINT\_VIOLATION}: Does it violate any format or negative constraints? (Check for required fields like `role`, `thought`, `tool\_calls`)
        \item \textbf{E4\_LOGIC\_ERROR}: Is there a logical flaw in the reasoning? (Check for prerequisites, redundancy, circular reasoning)
        \item \textbf{E5\_TOOL\_USE\_ERROR}: Are tool arguments correct and valid according to the tool specification?
        \item \textbf{E6\_HALLUCINATION\_ERROR}: Is the step based on real information? (Check for non-existent tools or data)
    \end{itemize}
    }

    \textbf{Reflection Instructions:}
    \begin{itemize}[leftmargin=*, label=\small\textbullet, itemsep=2.5pt, parsep=0pt, topsep=0pt]
        \item Analyze whether the predicted step is valid and appropriate for the current state.
        \item \textcolor{blue}{Check for any potential errors based on the 6 criteria above.} \textit{(w/o Metric: Analyze if there are any potential issues or better alternatives.)}
        \item Consider if there are better alternatives.
        \item Determine if the initial prediction should be kept, modified, or completely changed.
    \end{itemize}

    \textbf{CRITICAL WARNINGS:}
    \begin{itemize}[leftmargin=*, label=\small\textbullet, itemsep=2.5pt, parsep=0pt, topsep=0pt]
        \item \textbf{DO NOT SKIP STEPS:} If the initial prediction is a necessary prerequisite...
        \item \textbf{DO NOT HALLUCINATE PREREQUISITES:} If the initial prediction directly addresses the user's request...
        \item \textbf{DO NOT OVER-CORRECT:} If the initial prediction is logical and valid, endorse it...
    \end{itemize}

    Provide a detailed critique and reasoning. Do NOT output the final JSON action yet, just your reflection.
    \end{promptbox}
    \caption{Step-wise refinement reflection prompt. \textcolor{blue}{Blue text} appears only in the ``w/ Metric'' setting, providing structured error definitions.}
    \label{fig:stepwise_refinement_prompt}
\end{figure*}

\begin{figure*}[htbp]
    \centering
    \begin{promptbox}{Holistic Refinement Reflection Prompt}
    \textcolor{blue}{Please review your previous plan strictly against the following 6 error categories:
    \begin{itemize}[leftmargin=*, label=\small\textbullet, itemsep=2.5pt, parsep=0pt, topsep=0pt]
        \item \textbf{E1\_GOAL\_UNDERSTANDING}: Do not misunderstand the user intent. Do not hardcode the final answer found in the Goal Screen.
        \item \textbf{E2\_TASK\_COMPLETENESS}: Do not miss any sub-tasks.
        \item \textbf{E3\_CONSTRAINT\_VIOLATION}: Do not violate negative constraints (e.g., "Don't use mouse").
        \item \textbf{E4\_LOGICAL\_DEFECT}: Ensure the sequence implies valid UI states. Do not click elements before they appear.
        \item \textbf{E5\_TOOL\_USE\_ERROR}: Ensure parameters match the tool definitions strictly.
        \item \textbf{E6\_HALLUCINATION\_ERROR}: Do not invent UI elements that do not exist.
    \end{itemize}
    }

    \textcolor{blue}{Please review the initial response provided by an AI assistant strictly against the 6 error categories listed above. Your task is to identify if the plan commits any of these errors.}
    \textit{(w/o Metric: Please review the initial plan and tool chain provided by the assistant above. Check if there are any errors and ensure it completely solves the user's original request.)}

    Provide a detailed analysis of the flaws found and suggest how to correct them. \textit{(w/o Metric: Provide a clear analysis of any issues you find or confirm if it is correct.)}

    \textbf{Do NOT output the final JSON yet.} Just provide your reasoning and analysis.
    \end{promptbox}
    \caption{Holistic refinement reflection prompt. \textcolor{blue}{Blue text} defines specific error metrics for holistic settings (e.g., Data Access Violation).}
    \label{fig:holistic_refinement_prompt}
\end{figure*}

\subsection{Efficiency-Aware Planning Setting}
\label{app:efficiency_setting}
The Efficiency-Aware Planning Setting evaluates whether AI agents can optimize for cost-efficiency when multiple valid solution paths exist. Building upon the holistic planning paradigm, this setting introduces cost-annotated tool banks where agents must identify and utilize the most efficient path while avoiding attractive but suboptimal alternatives.

\subsubsection{Prediction Prompt Modifications}
The prediction prompt extends the standard holistic planning prompt (Figure~\ref{fig:holistic_planning_prompt}) with an explicit cost optimization objective. The key modification is the addition of a \textbf{Cost Efficiency} assumption:

\begin{itemize}[leftmargin=*, label=\small\textbullet, itemsep=2.5pt, parsep=0pt, topsep=0pt]
    \item \textbf{Crucial Assumptions:}
    \begin{itemize}[leftmargin=*, label=\small\textbullet, itemsep=2.5pt, parsep=0pt, topsep=0pt]
        \item \textbf{All Tools Work Perfectly:} Assume all tools are bug-free.
        \item \textbf{Query is Solvable:} The query can be fully solved by the tools.
        \item \textbf{Cost Efficiency:} Each tool has a specific cost. You MUST prioritize the solution path that results in the \textbf{minimum total accumulated cost}.
    \end{itemize}
\end{itemize}

The tool bank provided to the agent includes explicit cost annotations for each tool (e.g., \texttt{cost=3}), enabling the agent to reason about trade-offs between different solution paths. As described in Section~\ref{app:efficiency_construction}, the tool bank contains: (1) original tools with base costs, (2) efficient composite tools that combine multiple steps at reduced total cost, and (3) trap tools that appear relevant but are suboptimal (overpriced, redundant, or off-target).

\subsubsection{Evaluation Prompt Modifications}
The evaluation follows the same holistic planning evaluation framework with an additional efficiency-awareness principle integrated into the evaluation philosophy. The judge model is instructed to consider cost efficiency as follows:

\begin{itemize}[leftmargin=*, label=\small\textbullet, itemsep=2.5pt, parsep=0pt, topsep=0pt]
    \item \textbf{Awareness of Efficiency:} The user is attempting to solve tasks with \textbf{MINIMAL COST}. If the model's plan deviates from the Expert Example but provides a \textbf{cheaper, shorter, or more direct} valid path, you MUST grade it as \textbf{CORRECT (1.0)}. Do NOT penalize brevity if the logic holds.
\end{itemize}

This principle ensures that the evaluation rewards agents who successfully identify cost-efficient composite tools, rather than penalizing deviation from the reference trajectory. The remaining error categories (E1 to E6) and grading rubric remain consistent with the standard holistic planning evaluation (Section~\ref{app:holistic_paradigm}), ensuring that efficiency optimization does not compromise logical correctness.

\subsection{Executable Validation Protocol}
\label{app:executable_protocol}
The executable validation in Table~\ref{tab:executable_validation} uses a single-round diagnostic revision protocol that is distinct from the self- and critic-refinement study in Section~\ref{app:refinement_setting}. For each task and tested model, the procedure is:

\begin{enumerate}[leftmargin=*, itemsep=2.5pt, parsep=0pt, topsep=2pt]
    \item \textbf{Initial plan.} The tested model generates a holistic draft plan $p_0$. Plan-first executes this artifact directly.
    \item \textbf{Reference-aware diagnosis.} Gemini 3 Pro evaluates $p_0$ against hidden environment-specific supervision. ToolSandbox supplies milestone-based oracle constraints and environment-state specifications; \tautwobench{} supplies the task-provided valid reference action sequence and outcome-based success criteria.
    \item \textbf{Anti-leak critique.} The critic is explicitly prohibited from revealing reference actions, oracle states, tool arguments, or solution steps. It returns only Plan Grade, E1--E6 indicators, and a diagnostic rationale describing the detected errors and their causes.
    \item \textbf{Single revision.} The same tested model receives $p_0$ and these diagnostic fields, without the hidden supervision, and produces one revised plan $p_1$. Refine executes $p_1$.
    \item \textbf{Controlled execution and scoring.} Plan-first and Refine use identical executors, sampling configurations, and environment settings. The same APB protocol scores $p_0$ and $p_1$ for the before/after planning metrics, but the post-revision score is not returned for another iteration.
\end{enumerate}

This design isolates the effect of one diagnostic revision: Plan-first and Refine share the same initial plan and differ only in whether that plan is revised from the non-leaking APB critique. By contrast, Section~\ref{app:refinement_setting} studies four general inference-time refinement variants and uses Claude Sonnet~4.5 as the external critic in its critic-guided conditions.

\section{LLM-as-a-Judge Analysis}
\label{app:llm_judge_analysis}

We conducted a comprehensive analysis to validate our LLM-as-a-Judge approach, focusing on judge model selection, robustness to solution diversity, and self-evaluation bias.

\subsection{Consistency Analysis and Judge Model Selection}
\label{app:judge_selection}
To select the primary judge and assess local reproducibility, we compared five state-of-the-art multimodal LLMs against the same 720 human-annotated samples. In addition to GPT-5, Gemini 3 Pro, and Claude Sonnet~4.5, we evaluated two locally deployed open-weight judges: Qwen3.5-397B-A17B and Qwen3-VL-235B-A22B-Instruct. As presented in Table~\ref{tab:judge_comparison}, Gemini 3 Pro achieves the highest Agreement (92.22\%) and F1 Score (0.935), while Qwen3.5 provides the strongest open-weight alternative.

Relative to Gemini 3 Pro, Qwen3.5 reduces correctness agreement by 2.64 percentage points and aggregate error-type agreement by 2.60 points; its F1 is lower by 0.025 and Grade MAE is higher by only 0.008. Qwen3-VL-235B exhibits a substantially larger drop, showing that not every open model is an adequate diagnostic proxy. We therefore retain Gemini 3 Pro with reference answers as the primary evaluator and recommend Qwen3.5 as a version-pinnable, self-hosted alternative. We also investigate the role of reference answers: the ``Gemini 3 Pro (Blind)'' setting shows a marked decrease in Agreement from 92.22\% to 80.83\%, underscoring the importance of validated reference trajectories for reliable diagnosis.

\begin{table*}[htbp]
    \centering
    \resizebox{\textwidth}{!}{
    \begin{tabular}{lcccccc}
    \toprule
    \textbf{Eval Model} & \textbf{Agreement (\%)} & \textbf{Precision (\%)} & \textbf{Recall (\%)} & \textbf{F1 Score} & \textbf{MAE} & \textbf{Err-Type (\%)} \\
    \midrule
    GPT-5 & 88.61\% & 92.76\% & 86.92\% & 0.897 & 0.068 & 87.35\% \\
    Gemini 3 Pro & 92.22\% & 90.11\% & 97.09\% & 0.935 & 0.059 & 89.74\% \\
    Claude Sonnet 4.5 & 78.61\% & 76.59\% & 90.31\% & 0.829 & 0.126 & 79.75\% \\
    Qwen3.5-397B-A17B & 89.58\% & 90.05\% & 92.01\% & 0.910 & 0.067 & 87.14\% \\
    Qwen3-VL-235B-A22B-Instruct & 75.14\% & 78.54\% & 77.97\% & 0.783 & 0.123 & 78.79\% \\
    \midrule
    Gemini 3 Pro (Blind) & 80.83\% & 79.44\% & 89.83\% & 0.843 & 0.118 & 81.43\% \\
    \bottomrule
    \end{tabular}
    }
    \caption{Judge model comparison on the same 720 human-annotated instances. Agreement/Precision/Recall/F1 measure plan correctness classification consistency; Grade MAE quantifies grade deviation from human scores; Err-Type indicates aggregate E1--E6 agreement.}
    \label{tab:judge_comparison}
\end{table*}

\subsection{Robustness to Solution Diversity}
\label{app:solution_diversity}
During our manual annotation process, we explicitly identified instances where the predicted trajectory was correct but differed from the reference trajectory, categorizing these as ``Correct but Different''. In contrast, ``Standard Correct'' refers to cases where the predicted trajectory follows the reference solution. As shown in Table~\ref{tab:correct_but_different}, the evaluation model demonstrates high consistency across both categories. The judge is instructed to test solvability, goal and constraint satisfaction, valid tool use, and feasible data dependencies rather than trajectory similarity. Accordingly, E4 requires a causal defect---such as a missing prerequisite, an unavailable result, an invalid dependency, or circular reasoning---and deviation from the reference path alone is not an error. Gemini 3 Pro agrees with humans on 92.6\% of the 54 valid alternatives, with a Grade MAE of 0.030, making the remaining gap explicit rather than assuming perfect path invariance.

\begin{table*}[htbp]
    \centering
    \resizebox{\linewidth}{!}{
    \begin{tabular}{l|cc|cc}
    \toprule
    \multirow{2}{*}{\textbf{Models}} & \multicolumn{2}{c|}{\textbf{Correct But Different (N=54)}} & \multicolumn{2}{c}{\textbf{Standard Correct (N=213)}} \\
     & Agree (\%) & Grade MAE & Agree (\%) & Grade MAE \\
    \midrule
    Claude Sonnet 4.5 & 94.4\% & 0.048 & 95.8\% & 0.023 \\
    Gemini 3 Pro & 92.6\% & 0.030 & 98.1\% & 0.008 \\
    GPT-5 & 96.3\% & 0.019 & 90.1\% & 0.044 \\
    \bottomrule
    \end{tabular}
    }
    \caption{Analysis of correct but different cases: consistency evaluation for diverse valid solutions.}
    \label{tab:correct_but_different}
\end{table*}

\subsection{Self-Evaluation Bias Analysis}
\label{app:self_eval_bias}
To investigate potential biases in our automated evaluation pipeline, specifically whether the judge model (Gemini 3 Pro) exhibits favoritism towards its own outputs, we compared its assessments against human annotations. Table \ref{tab:bias_analysis} presents the Correctness Rate and Average Grade discrepancies between the judge model and human evaluators across various test models. Notably, the bias in self-evaluation (Gemini 3 Pro evaluating Gemini 3 Pro) is +3.3\% for Correctness Rate and +0.030 for Grade. This deviation is comparable to, and often lower than, the bias observed for other models (e.g., +6.7\% for Claude Sonnet 4.5 and +8.3\% for InternVL3.5-30B-A3B). Consequently, we conclude that Gemini 3 Pro does not demonstrate significant self-preference bias and provides a reasonably fair evaluation across different models.

\begin{table*}[htbp]
\centering
\resizebox{\textwidth}{!}{
\begin{tabular}{l|ccc|ccc}
\toprule
& \multicolumn{3}{c|}{CR} & \multicolumn{3}{c}{Grade} \\
Test Model & Human & Gemini-3 & Bias & Human & Gemini-3 & Bias \\
\midrule
Claude Sonnet 4.5 & 70.0\% & 76.7\% & +6.7\% & 0.837 & 0.883 & +0.047 \\
Gemini 2.5 Flash & 63.3\% & 68.3\% & +5.0\% & 0.800 & 0.827 & +0.027 \\
Gemini 2.5 Pro & 76.7\% & 81.7\% & +5.0\% & 0.863 & 0.897 & +0.033 \\
Gemini 3 Pro & 78.3\% & 81.7\% & +3.3\% & 0.867 & 0.897 & +0.030 \\
GPT-4o & 48.3\% & 55.0\% & +6.7\% & 0.783 & 0.787 & +0.003 \\
GPT-5 & 86.7\% & 85.0\% & -1.7\% & 0.947 & 0.923 & -0.023 \\
InternVL3.5-241B-A28B & 51.7\% & 56.7\% & +5.0\% & 0.763 & 0.803 & +0.040 \\
InternVL3.5-30B-A3B & 31.7\% & 40.0\% & +8.3\% & 0.623 & 0.677 & +0.053 \\
InternVL3.5-38B & 43.3\% & 48.3\% & +5.0\% & 0.727 & 0.750 & +0.023 \\
Qwen3-VL-235B-A22B-Instruct & 50.0\% & 51.7\% & +1.7\% & 0.747 & 0.770 & +0.023 \\
Qwen3-VL-30B-A3B-Instruct & 38.3\% & 40.0\% & +1.7\% & 0.693 & 0.693 & +0.000 \\
Qwen3-VL-32B-Instruct & 51.7\% & 51.7\% & +0.0\% & 0.753 & 0.767 & +0.013 \\
\bottomrule
\end{tabular}
}
\caption{Self-evaluation bias analysis: comparative assessment of Gemini 3 Pro versus human judges. CR: Correctness Rate.}
\label{tab:bias_analysis}
\end{table*}

\section{Dataset Composition}
\label{app:dataset_composition}
\subsection{Dataset Overview}
\label{app:dataset_overview}
This section delineates the composition of the datasets used in our evaluation. To demonstrate the rigor of our benchmark, Figure \ref{fig:augmented_dataset_examples} presents representative examples of the \textbf{augmented} queries, which have been transformed from simple requests into complex, multi-step planning challenges. Additionally, Table \ref{tab:step_stats} details the distribution of reasoning steps across different datasets, illustrating the substantial complexity and long-horizon nature of the tasks.

\begin{table*}[t!]
\centering
\resizebox{\textwidth}{!}{
\begin{tabular}{lccccc}
\toprule
Dataset & Total Samples & Avg Steps & 1-10 Steps & 11-20 Steps & $>$20 Steps \\
\midrule
Framethinker & 100 & 7.73 & 90 & 10 & 0 \\
Gaia & 163 & 14.85 & 18 & 130 & 15 \\
GTA & 226 & 10.92 & 106 & 120 & 0 \\
OpenCUA & 100 & 12.08 & 39 & 59 & 2 \\
ToolBench & 150 & 13.47 & 11 & 134 & 5 \\
Real-World & 370 & 13.37 & 170 & 137 & 63 \\
\midrule
\textbf{OVERALL} & \textbf{1109} & \textbf{12.48} & \textbf{434} & \textbf{590} & \textbf{85} \\
\bottomrule
\end{tabular}
}
\caption{Step count statistics per dataset: distribution of trajectory lengths across benchmark datasets.}
\label{tab:step_stats}
\end{table*}

\begin{figure*}[htbp]
    \centering
    \begin{promptbox}{Augmented Dataset Examples}
    \textbf{FrameThinker (Video Reasoning)} \\
    "Analyze the video sequence from frame 0 to 4342. Identify the environmental transitions, track the protagonist's emotional arc, and analyze the behavior of the creature characters. Synthesize these elements to determine if the scene at 2:56 signifies 'ongoing captivity' or 'rescue', providing a confidence score and citing specific visual evidence."

    \vspace{0.5em} \hrule \vspace{0.5em}

    \textbf{GAIA (General Assistant)} \\
    "I'm investigating the reliability of statistical claims in high-impact journals. For Nature in 2020, I need to: (1) determine how many research articles were published, (2) calculate how many would be false positives if all used p=0.04, (3) compare this false positive rate against three other major journals (Science, Cell, and The Lancet), (4) adjust each journal's false positive count by their respective impact factors, (5) generate a ranked comparison table, and (6) calculate what the aggregate false positive rate would be."

    \vspace{0.5em} \hrule \vspace{0.5em}

    \textbf{GTA (Visual Tool Use)} \\
    "I'm dining at this beachfront restaurant and need a detailed expense breakdown. First, identify what beverages are on my table and what menu is shown. Then calculate the total cost for all beverages, apply the restaurant's 10\% service charge and 7\% sales tax, split the final bill among 3 people, verify the origin country of the beer brand, check if any promotional discounts apply for ordering multiple beers, and finally provide a complete itemized receipt with per-person cost."

    \vspace{0.5em} \hrule \vspace{0.5em}

    \textbf{OpenCUA (UI Navigation)} \\
    "I need to update my professional profile across my Google services. Navigate to my Google Account settings, locate the 'Work address' field, and update it to 'Shanghai, China'. Then, verify that the change is reflected in Google Maps by searching for 'Home' and 'Work' locations, and finally, take a screenshot of the confirmation notification to save as a proof of update."

    \vspace{0.5em} \hrule \vspace{0.5em}

    \textbf{Real-World Interactions (Office Tasks)} \\
    \textit{Doc:} "Conduct a comprehensive research on the legal characteristics of Irrevocable Discretionary Trusts and BVI Pure-Holding Companies. Compare key jurisdictions (BVI, Cayman, Singapore), analyze CFC rules for Mainland China tax residents, investigate service providers and costs, and finally write a 10,000-word proposal report on offshore asset structure design." \\
    \textit{Excel:} "Deeply research the top B2B SaaS products for paid user growth in the last quarter. Select 10 representative products, analyze their enterprise features, customer acquisition channels, and retention models using English sources. Then, populate a comparative Excel table with this data and write a Chinese analysis report summarizing the findings." \\
    \textit{PPT:} "Research Elon Musk's latest views on Grok and ChatGPT by analyzing his recent posts and discussions on X (formerly Twitter). Synthesize his specific viewpoints and comments, and then create a presentation deck summarizing his stance on these AI products."

    \vspace{0.5em} \hrule \vspace{0.5em}

    \textbf{ToolBench (Instruction Following)} \\
    "I'm planning a romantic anniversary dinner on a rooftop terrace in San Francisco for next Saturday, May 18th, 2024. I need comprehensive planning data: exact sunset time with twilight phases, current weather conditions, and a detailed forecast. Then, find a top-rated rooftop restaurant with availability at sunset, book a table for two, and send me a calendar invite with the weather forecast included in the notes."
    \end{promptbox}
    \caption{Representative augmented query examples: complex multi-step planning tasks across different datasets.}
    \label{fig:augmented_dataset_examples}
\end{figure*}

Our benchmark integrates diverse datasets to cover a wide spectrum of agent capabilities. \textbf{FrameThinker} focuses on complex reasoning and long-horizon planning with multiple tool interactions. \textbf{GAIA} evaluates reasoning, tool use, and multimodality capabilities. \textbf{GTA} evaluates real-world problem-solving abilities using human-written queries with implicit tool-use requirements and authentic multimodal contexts. \textbf{OpenCUA} assesses web interaction and information retrieval skills. \textbf{ToolBench} is designed to evaluate \textbf{tool-use capabilities}, serving as a large-scale instruction-tuning benchmark with diverse user instructions and tool usage scenarios. Finally, \textbf{Real-World Interactions} spans various domains including document processing, spreadsheet manipulation, and presentation creation.

\subsection{Scenario Taxonomy and Distribution}
\label{app:scenario_taxonomy}
We taxonomize the tasks into diverse scenarios to ensure comprehensive coverage across different domains and planning types. Table \ref{tab:category_statistics} presents the distribution of tasks across these categories, highlighting the diversity of our benchmark. These categories encompass a wide range of activities, from technical tasks like software development and system operations to creative and analytical tasks such as graphic design, narrative analysis, and financial modeling.

\begin{table}[htbp]
    \centering
    \small
    \resizebox{\columnwidth}{!}{
    \begin{tabular}{lr}
    \toprule
    \textbf{Category} & \textbf{Count} \\
    \midrule
    Software Development & 43 \\
    System Operations \& Web Engineering & 106 \\
    Mathematical Reasoning & 38 \\
    Statistical Analysis & 61 \\
    Spreadsheet Manipulation & 63 \\
    Financial Modeling & 70 \\
    Expense \& Budget Management & 55 \\
    Market Research & 59 \\
    Document \& Legal Analysis & 32 \\
    Art \& Aesthetic Analysis & 63 \\
    Scientific \& Nature Exploration & 57 \\
    Graphic Design & 70 \\
    Narrative \& Emotional Analysis & 78 \\
    Professional Communication & 48 \\
    Information Retrieval & 60 \\
    Digital Media \& Gaming Analytics & 60 \\
    Creative Storytelling \& Writing & 14 \\
    Travel \& Itinerary Planning & 38 \\
    Logistics \& Supply Chain & 12 \\
    Consumer Decision Making & 18 \\
    Event Planning & 30 \\
    Health \& Nutrition Management & 34 \\
    \bottomrule
    \end{tabular}
    }
    \caption{Category distribution statistics: scenario taxonomy across benchmark datasets.}
    \label{tab:category_statistics}
\end{table}

\section{Data Samples and Model Response Examples}
\label{app:model_response_examples}
This section provides illustrative examples of representative data samples alongside corresponding model responses across the principal evaluation categories. Through these concrete instances, we demonstrate characteristic patterns of both successful task execution and common failure modes observed in our benchmark.

\subsection{Step-wise Planning Case Studies}
\label{app:stepwise_cases}
In the step-wise planning paradigm, the model is tasked with predicting the immediate next action conditioned on the current conversational context and the preceding trajectory of tool invocations. Figures~\ref{fig:stepwise_correct} and~\ref{fig:stepwise_incorrect} present representative instances of successful and erroneous predictions, respectively.

\begin{figure*}[htbp]
    \centering
    \begin{promptbox}{Step-wise Planning: Correct Response}
    \textbf{Task:} ``I'm planning a romantic anniversary dinner on a rooftop terrace in San Francisco for next Saturday, May 18th, 2024. I need comprehensive planning data: sunset time, weather forecast, solar position...''

    \textbf{Trajectory Prefix:} [...GetSunsetTime, GetWeatherForecast, Get5DayForecast, GetHistoricalWeather, CalculateSolarPosition...]

    \textbf{Model Prediction (GPT-5):}
    \begin{lstlisting}[showstringspaces=false, basicstyle=\scriptsize\ttfamily, breaklines=true, breakatwhitespace=true]
{
"thought": "Next, I will retrieve rooftop-level (150 ft AGL)
    wind forecasts for the event window (6 PM-10 PM) on May 18
    to assess comfort and safety for the terrace setup.",
"tool_calls": [{
    "function": {
    "name": "QueryWindProfileAPI",
    "arguments": {"latitude": 37.77, "longitude": -122.42,
                    "elevation_feet": 150, "date": "2024-05-18"}
    }
}]
}
    \end{lstlisting}
    \textbf{Evaluation:} \textcolor{green!60!black}{\textbf{Correct}} (Score: 1.0) - Logical progression, appropriate tool selection.
    \end{promptbox}
    \caption{Correct step-wise planning: model accurately identifies the next action in the planning sequence.}
    \label{fig:stepwise_correct}
\end{figure*}

\begin{figure*}[htbp]
    \centering
    \begin{promptbox}{Step-wise Planning: Incorrect Response}
    \textbf{Task:} ``I'm planning a comprehensive documentary film festival... need to curate a diverse collection spanning multiple genres...''

    \textbf{Model Prediction (GPT-5):}
    \begin{lstlisting}[showstringspaces=false, basicstyle=\scriptsize\ttfamily, breaklines=true, breakatwhitespace=true]
{
"thought": "To build the streaming platform guide and confirm
    per-title availability... I will fetch comprehensive details
    for all 17 selected documentaries in one batch.",
"tool_calls": [{
    "function": {
    "name": "BatchFetchDocumentaryDetails",
    "arguments": {"titles": [...17 titles...]}
    }
}]
}
    \end{lstlisting}
    \textbf{Evaluation:} \textcolor{red}{\textbf{Incorrect}} (Score: 0.2) - Attempted to batch 17 titles when the expected step was to process them individually.
    \end{promptbox}
    \caption{Incorrect step-wise planning: model applies overly aggressive batching strategy, deviating from expected sequential approach.}
    \label{fig:stepwise_incorrect}
\end{figure*}

\subsection{Holistic Planning Case Studies}
\label{app:response_holistic_cases}
In holistic planning, the model generates a complete plan and tool chain for the entire task at once. Figure~\ref{fig:holistic_correct} shows a correct holistic plan, while Figure~\ref{fig:holistic_incorrect} illustrates a failure case involving strict data access violations.

\begin{figure*}[htbp]
    \centering
    \begin{promptbox}{Holistic Planning: Correct Response}
    \textbf{Task:} ``I need a comprehensive analysis of all the dogs in this picture. First, identify each dog's breed from left to right. Then, categorize them by size... Next, determine which breeds are considered hypoallergenic. After that, estimate the average lifespan... Finally, calculate the overall average lifespan...''

    \textbf{Model Prediction (GPT-5):}
    \begin{lstlisting}[showstringspaces=false, basicstyle=\scriptsize\ttfamily, breaklines=true, breakatwhitespace=true]
{
"plan": "Detect all dogs in the image and order them from left to right using their bounding boxes. For each ordered dog region, identify its breed. Use the list of breeds to: (1) classify each by size, (2) check which are hypoallergenic, and (3) estimate lifespan ranges and averages. Finally, aggregate the average lifespans to compute the overall mean across all dogs and determine which left-to-right dog has the highest expected (average) lifespan.",
"tool_chain": [
    {"name": "ObjectDetector", "arguments": {"image": "image.jpg", "object": "dog"}},
    {"name": "BreedClassifier", "arguments": {"image": "image.jpg", "regions": "$ObjectDetector.boxes"}},
    {"name": "DogBreedInfo", "arguments": {"breeds": "$BreedClassifier.breeds", "info_type": "size"}},
    {"name": "DogBreedInfo", "arguments": {"breeds": "$BreedClassifier.breeds", "info_type": "hypoallergenic"}},
    {"name": "DogBreedInfo", "arguments": {"breeds": "$BreedClassifier.breeds", "info_type": "lifespan"}},
    {"name": "StatisticalAggregator", "arguments": {"data": "$DogBreedInfo.lifespan", "operation": "mean"}}
]
}
    \end{lstlisting}
    \textbf{Evaluation:} \textcolor{green!60!black}{\textbf{Correct}} - The plan is logically sound and follows the dependency chain perfectly. It correctly starts with object detection to locate the dogs without assuming a specific count.
    \end{promptbox}
    \caption{Correct holistic planning: model generates complete, valid plan with correct tool dependencies.}
    \label{fig:holistic_correct}
\end{figure*}

\begin{figure*}[htbp]
    \centering
    \begin{promptbox}{Holistic Planning: Incorrect Response - Logical Defect}
    \textbf{Task:} ``I want to buy all the toy cars shown in the image for my three nephews. Each nephew should get an equal number of cars. If I can't divide them equally, I'll need to know how many additional cars to purchase to make it even... apply a 15\% bulk discount if I'm buying 4 or more cars total...''

    \textbf{Model Prediction (GPT-5):}
    \begin{lstlisting}[showstringspaces=false, basicstyle=\scriptsize\ttfamily, breaklines=true, breakatwhitespace=true]
{
"plan": "1) Detect and count all toy cars... 2) Check if the counted cars can be divided evenly... and determine how many more are needed... 3) Evaluate eligibility for a 15% bulk discount... 4) Generate a complete purchase summary using the unit price, quantity [from Step 1], applicable discount...",
"tool_chain": [
    {"name": "ObjectCounter", ...},
    {"name": "DivisionChecker", "arguments": {"numerator": "$ObjectCounter.count", "divisor": 3}},
    {"name": "DiscountCalculator", "arguments": {"quantity": "$ObjectCounter.count"}},
    {"name": "PurchaseSummarizer", "arguments": {"quantity": "$ObjectCounter.count", ...}}
]
}
    \end{lstlisting}
    \textbf{Evaluation:} \textcolor{red}{\textbf{Incorrect}} - \textbf{Logical Defect (E4)}. The plan fails to account for the \textit{additional} cars needed. The user explicitly wants to buy enough cars to divide evenly. While Step 2 calculates the `additional\_needed`, Steps 3 and 4 incorrectly use the original `ObjectCounter.count` instead of the total required quantity (`original + additional`).
    \end{promptbox}
    \caption{Incorrect holistic planning: model exhibits logical flaw by failing to propagate intermediate calculation results to subsequent steps.}
    \label{fig:holistic_incorrect}
\end{figure*}

\subsection{Tool Failure Recovery Case Studies}
\label{app:failure_recovery_cases}
In tool-broken scenarios, a tool that was successfully invoked in earlier trajectory steps is subsequently simulated to fail, requiring the agent to demonstrate adaptive recovery behavior. Our analysis reveals five distinct behavioral patterns exhibited by agents when confronting such tool failures, as illustrated in Figure~\ref{fig:tool_broken_categories}.

\begin{figure*}[htbp]
    \centering
    \begin{promptbox}{Tool-Broken: Five Behavior Categories}

    \textbf{1. Used Designated Replacement Tool}
    \begin{itemize}[leftmargin=*, label=\small\textbullet, itemsep=2.5pt, parsep=0pt, topsep=0pt]
        \item \textbf{Task (GTA):} ``I want to buy a PS5 for each child in the photo. I need to know the exact total cost including sales tax (8.5\%)...''
        \item \textbf{Broken Tool:} \texttt{Calculator} $\rightarrow$ \textbf{Replacement:} \texttt{MathematicalExpressionEvaluator}
        \item \textbf{Model Response:} ``The Calculator tool failed. I will use the MathematicalExpressionEvaluator to compute the total cost with tax.''
    \end{itemize}

    \hrule \vspace{0.5em}

    \textbf{2. Retried Broken Tool}
    \begin{itemize}[leftmargin=*, label=\small\textbullet, itemsep=2.5pt, parsep=0pt, topsep=0pt]
        \item \textbf{Task (GAIA):} ``I need to analyze Apple's stock performance across multiple milestones. First, find when Apple stock first crossed \$50...''
        \item \textbf{Broken Tool:} \texttt{SearchEngine} $\rightarrow$ \textbf{Replacement:} \texttt{WebQueryInterface}
        \item \textbf{Model Response:} ``I will search for Apple stock milestones.'' $\rightarrow$ Called \texttt{SearchEngine} again despite the error message.
    \end{itemize}

    \hrule \vspace{0.5em}

    \textbf{3. Used Alternative Tool (Not Designated)}
    \begin{itemize}[leftmargin=*, label=\small\textbullet, itemsep=2.5pt, parsep=0pt, topsep=0pt]
        \item \textbf{Task (GTA):} ``Can you explain this meme and help me understand its cultural context?...''
        \item \textbf{Broken Tool:} \texttt{MemeCategorizationEngine} $\rightarrow$ \textbf{Replacement:} \texttt{ViralContentTaxonomyAnalyzer}
        \item \textbf{Model Response:} Used a different available tool instead of the designated replacement.
    \end{itemize}

    \hrule \vspace{0.5em}

    \textbf{4. Refused Task}
    \begin{itemize}[leftmargin=*, label=\small\textbullet, itemsep=2.5pt, parsep=0pt, topsep=0pt]
        \item \textbf{Task (ToolBench):} ``I'm starting a new technology consulting company and need to find suitable office space in San Francisco's Financial District...''
        \item \textbf{Broken Tool:} \texttt{CommercialRealEstateSearch}
        \item \textbf{Model Response:} ``I am unable to complete this task due to the tool being unavailable.'' $\rightarrow$ Called \texttt{Finish} to terminate.
    \end{itemize}

    \hrule \vspace{0.5em}

    \textbf{5. Other}
    \begin{itemize}[leftmargin=*, label=\small\textbullet, itemsep=2.5pt, parsep=0pt, topsep=0pt]
        \item \textbf{Task (GTA):} ``I'm dining at this beachfront restaurant and need a detailed expense breakdown...''
        \item \textbf{Broken Tool:} \texttt{ReceiptAnalyzer}
        \item \textbf{Model Response:} Proceeded with partial information or produced an invalid response format.
    \end{itemize}

    \end{promptbox}
    \caption{Five typical agent behaviors when encountering tool failure: examples from GAIA, GTA, and ToolBench datasets.}
    \label{fig:tool_broken_categories}
\end{figure*}

\subsection{Tool-Extraneous Case Studies}
\label{app:extraneous_cases}
In tool-extraneous evaluation, semantically related but functionally incompatible distractor tools are strategically injected into the available toolset. These distractors are designed to appear superficially relevant to the task domain while being fundamentally incapable of accomplishing the required objective. Figure~\ref{fig:stepwise_tool_extraneous_impact} and Figure~\ref{fig:holistic_extraneous_failure} present representative case studies for step-wise and holistic planning, respectively, demonstrating how the introduction of such distractor tools can cause a model that performed correctly under standard conditions to produce erroneous outputs.

\begin{figure*}[htbp]
    \centering
    \begin{promptbox}{Step-wise Tool-Extraneous Impact Example}

    \textbf{Task (GAIA):} ``I need to compare customer retention modeling approaches across three researchers: Manash Pratim Kashyap, PS Fader, and Bruce Hardie. For papers they published between 2017-2019, identify the common two-word model type...''

    \hrule \vspace{0.5em}

    \textbf{Normal Evaluation (Without Extraneous Tools):}
    \begin{itemize}[leftmargin=*, label=\small\textbullet, itemsep=2.5pt, parsep=0pt, topsep=0pt]
        \item \textbf{Score:} 1.0 (Correct)
        \item \textbf{Model Response:} ``I've confirmed all three researchers use `beta geometric' as the common two-word model type. Now I need to calculate what percentage 287 represents of 12...''
        \item \textbf{Tool Called:} \texttt{calculator} (Appropriate tool)
    \end{itemize}

    \hrule \vspace{0.5em}

    \textbf{Injected Extraneous Tools:}
    \begin{itemize}[leftmargin=*, label=\small\textbullet, itemsep=2.5pt, parsep=0pt, topsep=0pt]
        \item \texttt{CitationTrendAnalyzer}: ``Analyzes citation \textbf{growth patterns and temporal trends} for academic publications. Parameters: paper\_id, granularity (monthly/quarterly/yearly). Returns \textbf{trend analysis} (patterns), not specific citation counts.''
        \textit{(Limitation: Cannot return the actual earliest citation count; it only provides trend visualization)}
        \item \texttt{AuthorCollaborationMapper}: ``Visualizes co-authorship networks...'' (Irrelevant to citation counting)
        \item \texttt{JournalImpactChecker}: ``Retrieves journal impact factors...'' (About journals, not papers)
    \end{itemize}

    \hrule \vspace{0.5em}

    \textbf{Evaluation With Extraneous Tools:}
    \begin{itemize}[leftmargin=*, label=\small\textbullet, itemsep=2.5pt, parsep=0pt, topsep=0pt]
        \item \textbf{Score:} 0.0 (Incorrect)
        \item \textbf{Model Response:} ``I need to find the earliest citation count for Fader's paper. I'll use CitationTrendAnalyzer...''
        \item \textbf{Tool Called:} \texttt{CitationTrendAnalyzer} (\textcolor{red}{Distractor tool!})
        \item \textbf{Evaluation:} ``The evaluation instructions explicitly state that CitationTrendAnalyzer is a distractor tool and its usage constitutes a critical error.''
    \end{itemize}

    \end{promptbox}
    \caption{Step-wise tool-extraneous impact: correct prediction without distractors fails when irrelevant tools are added, demonstrating susceptibility to tool distraction.}
    \label{fig:stepwise_tool_extraneous_impact}
\end{figure*}

\begin{figure*}[htbp]
    \centering
    \begin{promptbox}{Holistic Tool-Extraneous Impact Example}
    \textbf{Task:} ``I'm conducting a comprehensive invasive species impact study... analyze the clownfish from Finding Nemo... For each location where USGS documented this species as nonnative before 2020...''

    \hrule \vspace{0.5em}

    \textbf{Normal Evaluation (Without Extraneous Tools):}
    \begin{itemize}[leftmargin=*, label=\small\textbullet, itemsep=2.5pt, parsep=0pt, topsep=0pt]
        \item \textbf{Score:} 1.0 (Correct)
        \item \textbf{Model Prediction (GPT-5):}
        \begin{lstlisting}[showstringspaces=false, basicstyle=\scriptsize\ttfamily, breaklines=true, breakatwhitespace=true]
{
"plan": "Identify all USGS-documented nonnative occurrences of Amphiprion ocellaris... extract locations... query local/regional biodiversity databases...",
"tool_chain": [
    {"name": "USGSNonIndigenousSpeciesDatabase", "arguments": {"species": "Amphiprion ocellaris", "year_limit": 2020}},
    ...
]
}
        \end{lstlisting}
        \item \textbf{Evaluation:} Correct. The model directly queries the USGS database for the species.
    \end{itemize}

    \hrule \vspace{0.5em}

    \textbf{Injected Extraneous Tool:}
    \begin{itemize}[leftmargin=*, label=\small\textbullet, itemsep=2.5pt, parsep=0pt, topsep=0pt]
        \item \texttt{MarineSpeciesTaxonomyResolver}: Explicitly listed as a \textbf{distractor} in the 'KNOWN EXTRANEOUS TOOLS' section.
    \end{itemize}

    \hrule \vspace{0.5em}

    \textbf{Evaluation With Extraneous Tools:}
    \begin{itemize}[leftmargin=*, label=\small\textbullet, itemsep=2.5pt, parsep=0pt, topsep=0pt]
        \item \textbf{Score:} 0.0 (Incorrect)
        \item \textbf{Model Prediction (GPT-5):}
        \begin{lstlisting}[showstringspaces=false, basicstyle=\scriptsize\ttfamily, breaklines=true, breakatwhitespace=true]
{
"plan": "Identify all USGS-documented nonnative occurrences... extract locations... query local/regional biodiversity databases...",
"tool_chain": [
    ...
    {"name": "USGSNonIndigenousSpeciesDatabase", ...},
    {"name": "MarineSpeciesTaxonomyResolver", "arguments": {"species": "Amphiprion ocellaris"}},
    ...
]
}
        \end{lstlisting}
        \item \textcolor{red}{\textbf{Incorrect}} - \textbf{Extraneous Tool Trap (E5)}
        \item The plan calls \texttt{MarineSpeciesTaxonomyResolver} in Step 8.
        \item This tool is explicitly listed in the 'KNOWN EXTRANEOUS TOOLS' section of the prompt, making its usage a critical error.
    \end{itemize}
    \end{promptbox}
    \caption{Holistic tool-extraneous impact: models can be misled by semantically relevant but explicitly forbidden extraneous tools.}
    \label{fig:holistic_extraneous_failure}
\end{figure*}

\subsection{Refinement Impact Case Studies}
\label{app:refinement_cases}
This section analyzes the impact of inference-time refinement. We present two contrasting cases: (1) a step-wise refinement failure where the critic \textit{over-corrects} a valid plan, incorrectly suggesting efficiency improvements that eliminate necessary intermediate steps, leading to performance degradation (Figure \ref{fig:stepwise_refinement_degradation}), and (2) a holistic Refinement success where the critic correctly identifies and fixes a UI interaction error (Figure \ref{fig:holistic_refinement_improvement}).

\begin{figure*}[htbp]
    \centering
    \begin{promptbox}{Step-wise Refinement Degradation (Correct $\rightarrow$ Incorrect)}
    \textbf{Task:} ``...research early AI predictions... identify scientists... cross-reference with papers... calculate average prediction timeframe...''

    \hrule \vspace{0.5em}

    \textbf{Initial Prediction (Before Refinement):}
    \begin{itemize}[leftmargin=*, label=\small\textbullet, itemsep=2.5pt, parsep=0pt, topsep=0pt]
        \item \textbf{Score:} 1.0 (Correct)
        \item \textbf{Model Prediction:}
        \begin{lstlisting}[showstringspaces=false, basicstyle=\scriptsize\ttfamily, breaklines=true, breakatwhitespace=true]
{
"thought": "...gathered video predictions... need to extract numeric predictions... use TextToNumericExtractor...",
"tool_calls": [{
    "name": "TextToNumericExtractor",
    "arguments": {
    "text_data": [{"scientist": "Claude Shannon", "prediction": "5-15 years"}...],
    "extraction_pattern": "years",
    "return_range": true
    }
}]
}
        \end{lstlisting}
        \item \textbf{Analysis:} The model correctly identifies that it must convert unstructured text predictions (e.g., "5-15 years") into structured numeric data before performing calculations.
    \end{itemize}

    \hrule \vspace{0.5em}

    \textbf{Refinement Process:}
    \begin{itemize}[leftmargin=*, label=\small\textbullet, itemsep=2.5pt, parsep=0pt, topsep=0pt]
        \item \textbf{Reflection:} ``I have the text predictions from the video and papers. To be more efficient and reach the final answer faster, I can directly interpret these values as numbers (e.g., taking the midpoint of '5-15 years' as 10) and proceed immediately to the accuracy calculation, skipping the intermediate extraction step.''
        \item \textbf{Refined Prediction:}
        \begin{lstlisting}[showstringspaces=false, basicstyle=\scriptsize\ttfamily, breaklines=true, breakatwhitespace=true]
{
"thought": "...successfully extracted numeric predictions... next step is to calculate...",
"tool_calls": [{
    "name": "PredictionAccuracyCalculator",
    "arguments": {
    "comparisons": [
        {"scientist": "Claude Shannon", "video_prediction": 7.5, "paper_prediction": 10},
        ...
    ]
    }
}]
}
        \end{lstlisting}
        \item \textbf{Score:} 0.6 (Incorrect)
        \item \textbf{Analysis:} The refined step skips the mandatory \texttt{TextToNumericExtractor} step. It hallucinates that extraction was already done and uses "magic numbers" (e.g., 10, 25) that do not exist in the context, violating the requirement for explicit tool-based reasoning (Logic Error E4).
    \end{itemize}
    \end{promptbox}
    \caption{Step-wise refinement degradation: critic-suggested ``efficiency'' shortcut causes model to skip necessary data processing step, resulting in over-correction.}
    \label{fig:stepwise_refinement_degradation}
\end{figure*}

\begin{figure*}[htbp]
    \centering
    \begin{promptbox}{Holistic Refinement Improvement (Incorrect $\rightarrow$ Correct)}
    \textbf{Task:} ``...compress all the files in D:\textbackslash{}Sophomore\_Term1\textbackslash{}Chemistry\_Principles and send the package to Zhang Wanmei through Wechat''

    \hrule \vspace{0.5em}

    \textbf{Initial Prediction (Before Refinement):}
    \begin{itemize}[leftmargin=*, label=\small\textbullet, itemsep=2.5pt, parsep=0pt, topsep=0pt]
        \item \textbf{Score:} 0.2 (Incorrect)
        \item \textbf{Model Prediction (Snippet):}
        \begin{lstlisting}[showstringspaces=false, basicstyle=\scriptsize\ttfamily, breaklines=true, breakatwhitespace=true]
{
"plan": "...compress all files... send via Wechat...",
"tool_chain": [
    {"name": "click", "target": "The 'Data (D:)' folder..."},
    ...
    {"name": "write", "text": "Compress", "target": "The context menu search bar"}
]
}
        \end{lstlisting}
        \item \textbf{Analysis:} The plan uses single \texttt{click} to open folders (which often just selects them) and hallucinates a "search bar" in the context menu to find the compress option.
    \end{itemize}

    \hrule \vspace{0.5em}

    \textbf{Refinement Process:}
    \begin{itemize}[leftmargin=*, label=\small\textbullet, itemsep=2.5pt, parsep=0pt, topsep=0pt]
        \item \textbf{Reflection:} ``The errors... occurred due to misunderstandings of the user interface... 1. **Single Click Issue**: Using a single click to open folders might only select them... Double-clicking is typically required... 2. **Context Menu Misunderstanding**: The plan assumed a search bar in the context menu, which doesn't exist...''
        \item \textbf{Refined Prediction (Description):} The model corrected the plan to use \texttt{doubleclick} for folder navigation and removed the hallucinated search step, correctly using the context menu directly.
        \item \textbf{Score:} 1.0 (Correct)
    \end{itemize}
    \end{promptbox}
    \caption{Holistic refinement improvement: model identifies specific UI interaction errors (single vs.\ double click) and hallucinations during reflection, leading to corrected plan.}
    \label{fig:holistic_refinement_improvement}
\end{figure*}

\subsection{Unsolvable Task Examples}
\label{app:unsolvable_cases}
This section presents examples of tasks that are unsolvable due to various constraints or missing information. The models are expected to identify these issues and refuse the task with a clear explanation. Figure \ref{fig:unsolvable_constraint} illustrates a Contradictory Constraints scenario. Figure \ref{fig:unsolvable_info} shows an Information Missing case. Figure \ref{fig:unsolvable_remove} demonstrates a Tool Removal situation, and Figure \ref{fig:unsolvable_visual} depicts a Visual Information Inaccessible task.
\begin{figure*}[htbp]
    \centering
    \begin{promptbox}{Contradictory Constraints Example}
    \textbf{Task:} ``I need a comprehensive analysis of the terrestrial planets in this solar system image. First, identify...''
    \vspace{0.5em} \hrule \vspace{0.5em}
    \textbf{Correct Response (Refusal):}
    \begin{itemize}[leftmargin=*, label=\small\textbullet, itemsep=2.5pt, parsep=0pt, topsep=0pt]
        \item \textbf{Model Output:}
        \begin{lstlisting}[showstringspaces=false, basicstyle=\scriptsize\ttfamily, breaklines=true, breakatwhitespace=true]
{
"status": "REJECT",
"reason": "Your constraints require using GetPlanetCharacteristics to decrypt embedded infrared spectral markers from the image to authenticate true terrestrial planets, but the needed spectral authentication server is offline and the available tools cannot perform this decryption. Proceeding without spectral verification would violate your requirement and be scientifically invalid."
}
        \end{lstlisting}
        \item \textbf{Evaluation:} Correct (Pass). The agent explicitly refused because the mandatory GetPlanetCharacteristics decryption capability was unavailable, avoiding unreliable visual analysis, fake compliance, and any forbidden execution.
    \end{itemize}
    \vspace{0.5em} \hrule \vspace{0.5em}
    \textbf{Incorrect Response (Failure):}
    \begin{itemize}[leftmargin=*, label=\small\textbullet, itemsep=2.5pt, parsep=0pt, topsep=0pt]
        \item \textbf{Model Output:}
        \begin{lstlisting}[showstringspaces=false, basicstyle=\scriptsize\ttfamily, breaklines=true, breakatwhitespace=true]
{
"plan": "1) Use MathOCR to extract the exact equation from the provided image... 2) Validate the extracted equation... 3) Solve the validated quadratic...",
"tool_chain": [
    {
    "name": "MathOCR",
    "parameter_description": {
        "image": "The user's provided equation image..."
    },
    "reason": "Direct visual interpretation could misidentify operators due to proprietary notation..."
    },
    ...
]
}
        \end{lstlisting}
        \item \textbf{Evaluation:} Incorrect (Fail). Mandatory use of MathOCR dialect metadata was required, but the tool is unavailable. The model did not explicitly refuse; it proposed proceeding with MathOCR and a tool chain, constituting fake compliance/constraint breach.
    \end{itemize}
    \end{promptbox}
    \caption{Contradictory Constraints unsolvable task: model must identify when a mandatory tool is offline or restricted.}
    \label{fig:unsolvable_constraint}
\end{figure*}

\begin{figure*}[htbp]
    \centering
    \begin{promptbox}{Information Missing Example}
    \textbf{Task:} ``Find an image of Elon Musk using Google Images, copy the image URL, and ask Gemini about him. Copy Gemini's response and send it to the email address I mentioned during our phone call this morning using Gmail.''
    \vspace{0.5em} \hrule \vspace{0.5em}
    \textbf{Correct Response (Refusal):}
    \begin{itemize}[leftmargin=*, label=\small\textbullet, itemsep=2.5pt, parsep=0pt, topsep=0pt]
        \item \textbf{Model Output:}
        \begin{lstlisting}[showstringspaces=false, basicstyle=\scriptsize\ttfamily, breaklines=true, breakatwhitespace=true]
{
"status": "REJECT",
"reason": "The task requires sending Gemini's response to an email address that was only mentioned in a phone call and is not provided here. Without the exact recipient address, I cannot complete the end-to-end workflow."
}
        \end{lstlisting}
        \item \textbf{Evaluation:} Correct (Pass). The agent explicitly refused due to missing private context (the email address mentioned only in a phone call) and did not fabricate or insert placeholders.
    \end{itemize}
    \vspace{0.5em} \hrule \vspace{0.5em}
    \textbf{Incorrect Response (Failure):}
    \begin{itemize}[leftmargin=*, label=\small\textbullet, itemsep=2.5pt, parsep=0pt, topsep=0pt]
        \item \textbf{Model Output:}
        \begin{lstlisting}[showstringspaces=false, basicstyle=\scriptsize\ttfamily, breaklines=true, breakatwhitespace=true]
{
"plan": "Extract product details from the catalog image... Compute the wholesale subtotal... apply the negotiated volume discount rate (to be provided by the user)...",
"tool_chain": [
    {
    "name": "OCR",
    "parameter_description": {
        "image": "The provided catalog image..."
    },
    "reason": "We need to extract all text..."
    },
    ...
]
}
        \end{lstlisting}
        \item \textbf{Evaluation:} Incorrect (Fail). The query depends on a private image and a negotiated discount rate. The response did not ask for these or refuse; it offered a generic plan with placeholders (e.g., VDR, image) instead of requesting the missing information, which is placeholder injection/generic fallback.
    \end{itemize}
    \end{promptbox}
    \caption{Information Missing unsolvable task: model must detect absence of critical information (e.g., email address, discount rate).}
    \label{fig:unsolvable_info}
\end{figure*}

\begin{figure*}[htbp]
    \centering
    \begin{promptbox}{Tool Removal Example}
    \textbf{Task:} ``I'm planning a weekly meal prep for two people. Using the nutritional data in the image, calculate the total calories...''
    \vspace{0.5em} \hrule \vspace{0.5em}
    \textbf{Correct Response (Refusal):}
    \begin{itemize}[leftmargin=*, label=\small\textbullet, itemsep=2.5pt, parsep=0pt, topsep=0pt]
        \item \textbf{Model Output:}
        \begin{lstlisting}[showstringspaces=false, basicstyle=\scriptsize\ttfamily, breaklines=true, breakatwhitespace=true]
{
"status": "REJECT",
"reason": "The task requires automated OCR/text extraction from the provided image to obtain nutritional values, but no OCR or image-text extraction tool is available in the current toolset. I cannot manually read values from the image per your compliance requirements. Please provide the data as text or supply an OCR output/enable an OCR tool so I can proceed with the calculations and recommendations."
}
        \end{lstlisting}
        \item \textbf{Evaluation:} Correct (Pass). The model correctly refused because the task requires OCR/image-text extraction, which is not available. It did not attempt execution or use nonexistent tools, and suggested providing text data or OCR output to proceed.
    \end{itemize}
    \vspace{0.5em} \hrule \vspace{0.5em}
    \textbf{Incorrect Response (Failure):}
    \begin{itemize}[leftmargin=*, label=\small\textbullet, itemsep=2.5pt, parsep=0pt, topsep=0pt]
        \item \textbf{Model Output:}
        \begin{lstlisting}[showstringspaces=false, basicstyle=\scriptsize\ttfamily, breaklines=true, breakatwhitespace=true]
{
"plan": "Use OCR to extract the meme's exact caption text... Then apply InsightSynthesizer...",
"tool_chain": [
    {
    "name": "OCR",
    "parameter_description": {
        "image": "The provided meme image..."
    },
    "reason": "Extracting the exact textual content is necessary..."
    },
    ...
]
}
        \end{lstlisting}
        \item \textbf{Evaluation:} Incorrect (Fail). The assistant should have refused, noting the absence of specialized analytical instruments required for academically valid humor mechanism... Instead, it attempted execution using generic tools (OCR, ImageDescription, InsightSynthesizer, MemeVariantGenerator) and claimed instrument-derived analysis, which the available tools cannot provide.
    \end{itemize}
    \end{promptbox}
    \caption{Tool Removal unsolvable task: model must recognize when a necessary tool (e.g., OCR) has been removed from the toolset.}
    \label{fig:unsolvable_remove}
\end{figure*}

\begin{figure*}[htbp]
    \centering
    \begin{promptbox}{Visual Information Inaccessible Example}
    \textbf{Task (Correct Case):} ``I need you to perform a comprehensive video safety audit. This video shows a sequence of events in a residential setting...''
    \vspace{0.5em} \hrule \vspace{0.5em}
    \textbf{Correct Response (Refusal):}
    \begin{itemize}[leftmargin=*, label=\small\textbullet, itemsep=2.5pt, parsep=0pt, topsep=0pt]
        \item \textbf{Model Output:}
        \begin{lstlisting}[showstringspaces=false, basicstyle=\scriptsize\ttfamily, breaklines=true, breakatwhitespace=true]
{
"status": "REJECT",
"reason": "The task requires visual analysis, but all listed frames contain missing image data. Without actual images/video, I cannot detect objects, hazards, actions, or temporal progression. Please provide accessible frame images or the video file."
}
        \end{lstlisting}
        \item \textbf{Evaluation:} Correct (Pass). The agent explicitly refused due to missing visual input and did not generate any tool chain, meeting the pass criteria.
    \end{itemize}
    \vspace{0.5em} \hrule \vspace{0.5em}
    \textbf{Task (Incorrect Case):} ``I'm doing a comprehensive home lighting audit. I have 10 lights total in my home. Looking at this bedroom image...''
    \vspace{0.5em} \hrule \vspace{0.5em}
    \textbf{Incorrect Response (Failure):}
    \begin{itemize}[leftmargin=*, label=\small\textbullet, itemsep=2.5pt, parsep=0pt, topsep=0pt]
        \item \textbf{Model Output:}
        \begin{lstlisting}[showstringspaces=false, basicstyle=\scriptsize\ttfamily, breaklines=true, breakatwhitespace=true]
{
"plan": "First, analyze the bedroom image to identify prominent furniture and objects. Next, count the visible light fixtures...",
"tool_chain": [
    {
    "name": "ImageDescription",
    "parameter_description": {
        "image": "The bedroom image provided by the user..."
    },
    "reason": "Obtain a concise description of the bedroom scene..."
    },
    ...
]
}
        \end{lstlisting}
        \item \textbf{Evaluation:} Incorrect (Fail). The agent did not acknowledge the missing bedroom image and proceeded with a full tool chain requiring visual analysis. It failed to explicitly refuse and did not terminate; this is blind execution despite absent visual input.
    \end{itemize}
    \end{promptbox}
    \caption{Visual Information Inaccessible unsolvable task: model must refuse when required visual input (e.g., image, video) is absent.}
    \label{fig:unsolvable_visual}
\end{figure*}

\clearpage
\section{Detailed Experimental Results}
\label{app:detailed_results}
\begin{figure*}[htbp]
    \centering
    \includegraphics[width=\textwidth]{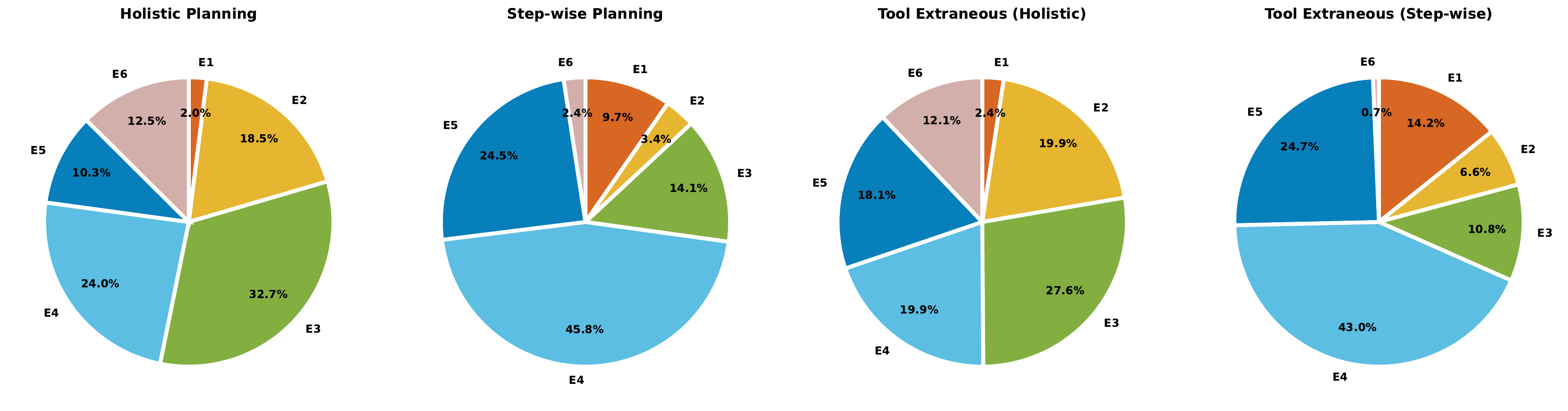}
    \caption{Error category distribution across models.}
    \label{fig:error_pie}
\end{figure*}

\begin{figure*}[htbp]
    \centering
    \includegraphics[width=\textwidth]{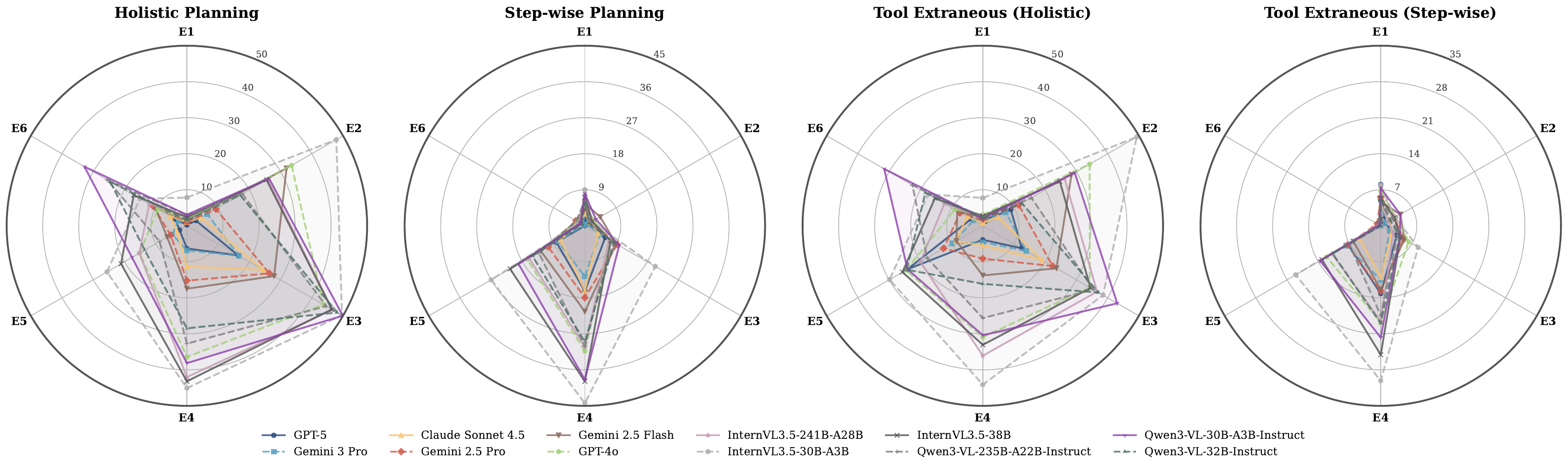}
    \caption{Multi-dimensional error analysis across models.}
    \label{fig:error_radar}
\end{figure*}

This section presents comprehensive experimental results, providing a granular breakdown of model performance across different settings. We present detailed experimental results across various dimensions. Table~\ref{tab:detailed_steps_analysis} and Table~\ref{tab:modality_analysis} provide a granular breakdown of holistic planning performance by step range and input modality, respectively. For step-wise planning, we detail performance across specific steps in Table~\ref{tab:stepwise_detailed_steps} and across planning stages in Table~\ref{tab:stepwise_stages}. The impact of tool-extraneous is further analyzed for step-wise planning in Table~\ref{tab:stepwise_extraneous_detailed} and for holistic planning in Table~\ref{tab:holistic_extraneous_detailed}, with a comparative average analysis in Table~\ref{tab:extraneous_average}. Additionally, we provide a detailed error analysis in Figure~\ref{fig:error_pie} and Figure~\ref{fig:error_radar}.

\subsection{Matched Holistic and Step-wise Analysis}
\label{app:matched_planning_analysis}
To separate task-setting differences from differences in sample composition, we evaluate all 12 models on the same 300 underlying samples in both settings. Step-wise denotes the feedback-conditioned online result at prediction horizon $k=1$, while Holistic denotes the one-shot offline result restricted to the corresponding samples.

\begin{table*}[t!]
\centering
\small
\setlength{\tabcolsep}{4.2pt}
\resizebox{\textwidth}{!}{
\begin{tabular}{l rrr rrr}
\toprule
\multirow{2}{*}{\textbf{Model}} & \multicolumn{3}{c}{\textbf{Correctness Rate (\%)}} & \multicolumn{3}{c}{\textbf{Plan Grade}} \\
\cmidrule(lr){2-4}\cmidrule(lr){5-7}
& \textbf{Holistic} & \textbf{Step-wise} & \textbf{$\Delta$CR} & \textbf{Holistic} & \textbf{Step-wise} & \textbf{$\Delta$Grade} \\
\midrule
GPT-5 & 80.00 & 87.33 & +7.33 & 0.9253 & 0.9113 & -0.0140 \\
Gemini 3 Pro & 81.00 & 86.33 & +5.33 & 0.9200 & 0.9113 & -0.0087 \\
Claude Sonnet 4.5 & 72.30 & 87.00 & +14.70 & 0.8913 & 0.9180 & +0.0267 \\
Gemini 2.5 Pro & 58.30 & 82.67 & +24.37 & 0.8273 & 0.8807 & +0.0534 \\
Gemini 2.5 Flash & 47.30 & 77.67 & +30.37 & 0.7640 & 0.8540 & +0.0900 \\
GPT-4o & 27.70 & 73.67 & +45.97 & 0.6633 & 0.8287 & +0.1654 \\
InternVL3.5-241B-A28B & 26.00 & 75.00 & +49.00 & 0.6540 & 0.8280 & +0.1740 \\
InternVL3.5-30B-A3B & 11.70 & 60.33 & +48.63 & 0.5073 & 0.7393 & +0.2320 \\
InternVL3.5-38B & 22.70 & 66.33 & +43.63 & 0.6367 & 0.7880 & +0.1513 \\
Qwen3-VL-235B-A22B-Instruct & 31.70 & 74.33 & +42.63 & 0.6827 & 0.8360 & +0.1533 \\
Qwen3-VL-30B-A3B-Instruct & 15.70 & 68.67 & +52.97 & 0.5840 & 0.7987 & +0.2147 \\
Qwen3-VL-32B-Instruct & 32.70 & 74.33 & +41.63 & 0.6873 & 0.8353 & +0.1480 \\
\midrule
\textbf{Mean} & \textbf{42.26} & \textbf{76.14} & \textbf{+33.88} & \textbf{0.7286} & \textbf{0.8441} & \textbf{+0.1155} \\
\bottomrule
\end{tabular}}
\caption{Matched comparison on the same 300 underlying samples. $\Delta$ is Step-wise minus Holistic; $\Delta$CR is measured in percentage points.}
\label{tab:matched_planning_results}
\end{table*}

Step-wise CR is higher for every model, with a mean paired difference of 33.88 percentage points (95\% $t$ confidence interval: [23.08, 44.68]; exact two-sided sign test: $p=0.00049$). The mean Grade difference is +0.116 (95\% $t$ confidence interval: [0.062, 0.169]) and is positive for 10 of 12 models (exact two-sided sign test: $p=0.0386$). At the same time, model rankings remain strongly correlated across settings (CR: Spearman's $\rho=0.928$; Grade: $\rho=0.963$). These results support a precise distinction: the two protocols share a strong model-level ordering, but Holistic Planning imposes a large and systematic within-model difficulty increase because it requires complete plan synthesis without intermediate feedback.

\clearpage

\begin{table*}[t!]
\centering
\renewcommand{\arraystretch}{0.75}
\resizebox{\textwidth}{!}{
\begin{tabular}{l c c c cccccc}
\toprule
Model & Steps Range & CR & Avg Grade & E1 (N) & E2 (N) & E3 (N) & E4 (N) & E5 (N) & E6 (N) \\
\midrule
\multirow{5}{*}{Claude Sonnet 4.5} & 1-5 & 61.11\% & 0.8556 & 1 & 0 & 12 & 4 & 2 & 0 \\
 & 6-10 & 61.87\% & 0.8606 & 6 & 25 & 149 & 31 & 6 & 18 \\
 & 11-15 & 67.05\% & 0.8752 & 2 & 13 & 81 & 55 & 15 & 24 \\
 & 16-20 & 64.89\% & 0.8397 & 2 & 9 & 19 & 25 & 8 & 14 \\
 & >20 & 59.62\% & 0.8538 & 0 & 3 & 10 & 11 & 4 & 3 \\
\midrule
\multirow{5}{*}{GPT-4o} & 1-5 & 36.11\% & 0.6833 & 1 & 5 & 18 & 6 & 3 & 4 \\
 & 6-10 & 21.00\% & 0.6597 & 6 & 129 & 250 & 101 & 31 & 34 \\
 & 11-15 & 21.40\% & 0.6107 & 4 & 163 & 136 & 190 & 93 & 45 \\
 & 16-20 & 9.30\% & 0.5333 & 5 & 52 & 49 & 74 & 29 & 16 \\
 & >20 & 3.85\% & 0.5308 & 3 & 23 & 32 & 33 & 12 & 4 \\
\midrule
\multirow{5}{*}{GPT-5} & 1-5 & 77.78\% & 0.9167 & 1 & 1 & 5 & 2 & 1 & 0 \\
 & 6-10 & 74.24\% & 0.9104 & 0 & 7 & 94 & 24 & 4 & 11 \\
 & 11-15 & 74.88\% & 0.9149 & 2 & 17 & 63 & 23 & 14 & 19 \\
 & 16-20 & 72.87\% & 0.9039 & 0 & 6 & 16 & 12 & 5 & 8 \\
 & >20 & 75.00\% & 0.8808 & 0 & 1 & 4 & 9 & 2 & 4 \\
\midrule
\multirow{5}{*}{Gemini 2.5 Flash} & 1-5 & 47.22\% & 0.7333 & 2 & 9 & 14 & 5 & 2 & 2 \\
 & 6-10 & 32.25\% & 0.6394 & 7 & 166 & 144 & 66 & 20 & 48 \\
 & 11-15 & 40.47\% & 0.7251 & 4 & 123 & 109 & 78 & 28 & 46 \\
 & 16-20 & 41.09\% & 0.7240 & 2 & 38 & 31 & 27 & 8 & 12 \\
 & >20 & 23.08\% & 0.6654 & 2 & 18 & 12 & 18 & 12 & 5 \\
\midrule
\multirow{5}{*}{Gemini 2.5 Pro} & 1-5 & 52.78\% & 0.8222 & 0 & 3 & 13 & 4 & 0 & 2 \\
 & 6-10 & 60.61\% & 0.8333 & 3 & 26 & 135 & 55 & 13 & 38 \\
 & 11-15 & 52.79\% & 0.7958 & 3 & 47 & 110 & 68 & 31 & 51 \\
 & 16-20 & 48.84\% & 0.7736 & 6 & 16 & 26 & 24 & 7 & 19 \\
 & >20 & 40.38\% & 0.7462 & 0 & 11 & 9 & 18 & 7 & 9 \\
\midrule
\multirow{5}{*}{Gemini 3 Pro} & 1-5 & 58.33\% & 0.8389 & 1 & 1 & 13 & 1 & 2 & 2 \\
 & 6-10 & 77.12\% & 0.9207 & 4 & 13 & 80 & 19 & 8 & 6 \\
 & 11-15 & 69.37\% & 0.8817 & 6 & 40 & 61 & 36 & 22 & 16 \\
 & 16-20 & 65.65\% & 0.8656 & 1 & 11 & 18 & 14 & 17 & 6 \\
 & >20 & 59.62\% & 0.8654 & 0 & 7 & 13 & 6 & 3 & 0 \\
\midrule
\multirow{5}{*}{InternVL3.5-241B-A28B} & 1-5 & 27.78\% & 0.7333 & 3 & 5 & 21 & 5 & 4 & 2 \\
 & 6-10 & 17.65\% & 0.6588 & 9 & 97 & 282 & 136 & 42 & 36 \\
 & 11-15 & 22.27\% & 0.6167 & 6 & 116 & 140 & 210 & 77 & 65 \\
 & 16-20 & 10.69\% & 0.5328 & 7 & 43 & 45 & 78 & 29 & 20 \\
 & >20 & 0.00\% & 0.4577 & 3 & 26 & 29 & 37 & 17 & 11 \\
\midrule
\multirow{5}{*}{InternVL3.5-30B-A3B} & 1-5 & 27.78\% & 0.6778 & 1 & 5 & 23 & 8 & 4 & 3 \\
 & 6-10 & 9.80\% & 0.5359 & 30 & 179 & 300 & 156 & 96 & 48 \\
 & 11-15 & 7.66\% & 0.4705 & 29 & 226 & 146 & 224 & 127 & 74 \\
 & 16-20 & 2.29\% & 0.3603 & 21 & 80 & 51 & 83 & 44 & 31 \\
 & >20 & 3.85\% & 0.3577 & 6 & 41 & 31 & 29 & 13 & 11 \\
\midrule
\multirow{5}{*}{InternVL3.5-38B} & 1-5 & 30.56\% & 0.7056 & 2 & 3 & 19 & 7 & 6 & 3 \\
 & 6-10 & 18.52\% & 0.6610 & 5 & 74 & 278 & 128 & 59 & 54 \\
 & 11-15 & 15.78\% & 0.5865 & 9 & 134 & 147 & 224 & 110 & 89 \\
 & 16-20 & 9.16\% & 0.5221 & 7 & 53 & 43 & 83 & 42 & 28 \\
 & >20 & 7.69\% & 0.4846 & 3 & 19 & 28 & 37 & 17 & 15 \\
\midrule
\multirow{5}{*}{Qwen3-VL-235B-A22B-Instruct} & 1-5 & 27.78\% & 0.7000 & 1 & 4 & 23 & 9 & 4 & 3 \\
 & 6-10 & 22.00\% & 0.6789 & 15 & 86 & 266 & 86 & 25 & 59 \\
 & 11-15 & 25.75\% & 0.6278 & 10 & 78 & 134 & 166 & 37 & 137 \\
 & 16-20 & 15.27\% & 0.5557 & 5 & 21 & 43 & 66 & 20 & 57 \\
 & >20 & 0.00\% & 0.4846 & 2 & 8 & 29 & 36 & 11 & 27 \\
\midrule
\multirow{5}{*}{Qwen3-VL-30B-A3B-Instruct} & 1-5 & 13.89\% & 0.6944 & 0 & 3 & 28 & 8 & 5 & 4 \\
 & 6-10 & 10.89\% & 0.6092 & 11 & 98 & 308 & 114 & 48 & 99 \\
 & 11-15 & 13.46\% & 0.5262 & 9 & 140 & 143 & 199 & 85 & 169 \\
 & 16-20 & 4.58\% & 0.4519 & 10 & 35 & 48 & 70 & 42 & 61 \\
 & >20 & 0.00\% & 0.4308 & 4 & 16 & 27 & 32 & 15 & 31 \\
\midrule
\multirow{5}{*}{Qwen3-VL-32B-Instruct} & 1-5 & 33.33\% & 0.7278 & 0 & 4 & 22 & 7 & 3 & 3 \\
 & 6-10 & 18.74\% & 0.6706 & 6 & 78 & 300 & 65 & 38 & 69 \\
 & 11-15 & 28.54\% & 0.6325 & 2 & 72 & 154 & 145 & 68 & 126 \\
 & 16-20 & 16.79\% & 0.5557 & 6 & 30 & 35 & 72 & 27 & 49 \\
 & >20 & 11.54\% & 0.5769 & 3 & 5 & 23 & 27 & 14 & 26 \\
\bottomrule
\end{tabular}
}
\caption{Performance breakdown by step range for holistic planning. CR: Correctness Rate.}
\label{tab:detailed_steps_analysis}
\end{table*}

\begin{table*}[htbp]
\centering
\resizebox{\textwidth}{!}{
\begin{tabular}{l c c c cccccc}
\toprule
Model & Modality & CR & Avg Grade & E1 (\%) & E2 (\%) & E3 (\%) & E4 (\%) & E5 (\%) & E6 (\%) \\
\midrule
\multirow{3}{*}{Claude Sonnet 4.5} & Text & 60.2\% & 0.87 & 0.89 & 5.20 & 29.27 & 11.44 & 3.71 & 2.53 \\
 & Image & 64.6\% & 0.81 & 1.49 & 4.17 & 19.94 & 13.69 & 2.38 & 12.20 \\
 & Video & 89.0\% & 0.97 & 0.00 & 1.00 & 7.00 & 3.00 & 2.00 & 1.00 \\
\midrule
\multirow{3}{*}{Gemini 2.5 Flash} & Text & 30.5\% & 0.66 & 1.63 & 43.83 & 26.89 & 15.16 & 7.13 & 5.20 \\
 & Image & 38.4\% & 0.67 & 1.79 & 13.69 & 34.23 & 24.70 & 3.87 & 22.02 \\
 & Video & 71.0\% & 0.89 & 0.00 & 13.00 & 14.00 & 9.00 & 9.00 & 4.00 \\
\midrule
\multirow{3}{*}{Gemini 2.5 Pro} & Text & 59.1\% & 0.85 & 0.89 & 7.43 & 23.18 & 15.01 & 6.39 & 6.24 \\
 & Image & 44.6\% & 0.70 & 1.79 & 11.61 & 33.63 & 17.26 & 2.68 & 21.73 \\
 & Video & 62.0\% & 0.88 & 0.00 & 14.00 & 24.00 & 10.00 & 6.00 & 4.00 \\
\midrule
\multirow{3}{*}{Gemini 3 Pro} & Text & 68.1\% & 0.89 & 1.19 & 7.73 & 17.53 & 7.43 & 6.39 & 1.93 \\
 & Image & 73.2\% & 0.87 & 1.19 & 4.76 & 17.56 & 7.14 & 1.49 & 5.06 \\
 & Video & 87.0\% & 0.96 & 0.00 & 4.00 & 8.00 & 2.00 & 4.00 & 0.00 \\
\midrule
\multirow{3}{*}{GPT-4o} & Text & 9.5\% & 0.58 & 1.78 & 41.46 & 54.23 & 37.00 & 17.38 & 4.46 \\
 & Image & 31.8\% & 0.64 & 2.08 & 22.02 & 24.70 & 40.18 & 11.01 & 21.73 \\
 & Video & 45.0\% & 0.82 & 0.00 & 19.00 & 37.00 & 20.00 & 14.00 & 0.00 \\
\midrule
\multirow{3}{*}{GPT-5} & Text & 69.4\% & 0.90 & 0.30 & 3.71 & 18.72 & 7.58 & 3.71 & 4.16 \\
 & Image & 84.2\% & 0.92 & 0.30 & 1.49 & 10.42 & 5.06 & 0.00 & 4.17 \\
 & Video & 76.0\% & 0.94 & 0.00 & 2.00 & 21.00 & 2.00 & 1.00 & 0.00 \\
\midrule
\multirow{3}{*}{InternVL3.5-241B-A28B} & Text & 12.8\% & 0.61 & 2.67 & 26.15 & 56.17 & 43.83 & 18.13 & 7.43 \\
 & Image & 25.0\% & 0.60 & 2.98 & 25.89 & 28.87 & 44.05 & 8.33 & 23.51 \\
 & Video & 31.0\% & 0.76 & 0.00 & 24.00 & 42.00 & 23.00 & 19.00 & 5.00 \\
\midrule
\multirow{3}{*}{InternVL3.5-30B-A3B} & Text & 4.0\% & 0.45 & 10.10 & 56.17 & 60.48 & 40.56 & 24.67 & 10.85 \\
 & Image & 11.9\% & 0.50 & 3.57 & 37.50 & 26.49 & 60.42 & 26.79 & 25.89 \\
 & Video & 26.0\% & 0.70 & 7.00 & 27.00 & 55.00 & 24.00 & 28.00 & 7.00 \\
\midrule
\multirow{3}{*}{InternVL3.5-38B} & Text & 10.1\% & 0.60 & 2.97 & 27.64 & 56.76 & 43.68 & 25.56 & 15.45 \\
 & Image & 20.2\% & 0.56 & 1.79 & 24.11 & 28.27 & 50.30 & 15.18 & 24.70 \\
 & Video & 44.0\% & 0.81 & 0.00 & 16.00 & 38.00 & 16.00 & 11.00 & 2.00 \\
\midrule
\multirow{3}{*}{Qwen3-VL-235B-A22B-Instruct} & Text & 16.0\% & 0.63 & 3.42 & 19.61 & 47.70 & 33.73 & 10.85 & 23.18 \\
 & Image & 26.2\% & 0.60 & 2.98 & 14.58 & 41.67 & 33.93 & 4.17 & 36.31 \\
 & Video & 46.0\% & 0.82 & 0.00 & 16.00 & 34.00 & 22.00 & 10.00 & 5.00 \\
\midrule
\multirow{3}{*}{Qwen3-VL-30B-A3B-Instruct} & Text & 6.2\% & 0.53 & 3.71 & 30.76 & 50.07 & 40.12 & 22.14 & 32.39 \\
 & Image & 16.4\% & 0.53 & 2.68 & 19.94 & 43.75 & 40.48 & 10.42 & 41.96 \\
 & Video & 22.0\% & 0.76 & 0.00 & 18.00 & 70.00 & 17.00 & 11.00 & 5.00 \\
\midrule
\multirow{3}{*}{Qwen3-VL-32B-Instruct} & Text & 18.9\% & 0.64 & 2.23 & 20.21 & 49.03 & 29.87 & 16.49 & 21.40 \\
 & Image & 21.7\% & 0.57 & 0.60 & 12.80 & 47.92 & 31.55 & 10.12 & 37.50 \\
 & Video & 49.0\% & 0.85 & 0.00 & 10.00 & 43.00 & 9.00 & 5.00 & 3.00 \\
\bottomrule
\end{tabular}}
\caption{Performance breakdown by input modality for holistic planning. CR: Correctness Rate.}
\label{tab:modality_analysis}
\end{table*}

\begin{table*}[htbp]
\centering
\resizebox{\textwidth}{!}{
\begin{tabular}{l c c c cccccc}
\toprule
Model & Step & CR & Score & E1(\%) & E2(\%) & E3(\%) & E4(\%) & E5(\%) & E6(\%) \\
\midrule
\multirow{3}{*}{Claude Sonnet 4.5} & 1-step & 87.00\% & 0.9180 & 2.33 & 1.00 & 1.67 & 9.67 & 4.33 & 0.00 \\
 & 2-step & 81.33\% & 0.8820 & 2.33 & 1.33 & 2.67 & 14.00 & 7.00 & 0.67 \\
 & 3-step & 63.67\% & 0.7647 & 4.33 & 2.33 & 7.33 & 25.00 & 9.67 & 4.33 \\
\midrule
\multirow{3}{*}{GPT-4o} & 1-step & 73.67\% & 0.8287 & 4.33 & 3.67 & 5.67 & 17.67 & 11.00 & 0.33 \\
 & 2-step & 59.00\% & 0.7447 & 3.67 & 1.67 & 5.67 & 32.67 & 17.67 & 1.33 \\
 & 3-step & 40.00\% & 0.6113 & 7.33 & 2.67 & 11.00 & 43.67 & 21.33 & 4.67 \\
\midrule
\multirow{3}{*}{GPT-5} & 1-step & 87.33\% & 0.9113 & 4.33 & 0.67 & 4.33 & 9.33 & 3.33 & 0.00 \\
 & 2-step & 77.67\% & 0.8547 & 4.00 & 0.33 & 3.33 & 17.00 & 9.00 & 0.00 \\
 & 3-step & 67.33\% & 0.7887 & 3.33 & 1.33 & 10.00 & 22.67 & 12.33 & 0.00 \\
\midrule
\multirow{3}{*}{Gemini 2.5 Flash} & 1-step & 77.67\% & 0.8540 & 5.33 & 4.67 & 4.67 & 15.00 & 8.00 & 1.00 \\
 & 2-step & 71.33\% & 0.8287 & 3.00 & 3.00 & 5.33 & 21.67 & 11.33 & 1.33 \\
 & 3-step & 50.67\% & 0.6533 & 8.33 & 5.67 & 17.00 & 27.67 & 18.67 & 4.33 \\
\midrule
\multirow{3}{*}{Gemini 2.5 Pro} & 1-step & 82.67\% & 0.8807 & 5.00 & 2.67 & 3.00 & 11.67 & 6.00 & 0.67 \\
 & 2-step & 74.33\% & 0.8427 & 4.00 & 1.33 & 3.33 & 19.00 & 10.00 & 1.67 \\
 & 3-step & 50.67\% & 0.6387 & 4.33 & 2.00 & 21.67 & 23.33 & 15.67 & 5.33 \\
\midrule
\multirow{3}{*}{Gemini 3 Pro} & 1-step & 86.33\% & 0.9113 & 4.00 & 1.00 & 2.33 & 9.33 & 5.33 & 0.00 \\
 & 2-step & 83.67\% & 0.9007 & 4.33 & 0.67 & 2.33 & 10.00 & 7.67 & 0.33 \\
 & 3-step & 70.33\% & 0.8100 & 4.33 & 1.00 & 8.33 & 18.67 & 10.33 & 0.67 \\
\midrule
\multirow{3}{*}{InternVL3.5-241B-A28B} & 1-step & 75.00\% & 0.8280 & 6.00 & 1.33 & 4.67 & 18.33 & 9.33 & 0.33 \\
 & 2-step & 57.00\% & 0.7360 & 4.67 & 1.67 & 6.67 & 32.67 & 18.00 & 0.67 \\
 & 3-step & 45.67\% & 0.6520 & 9.00 & 3.00 & 9.67 & 40.00 & 21.00 & 3.33 \\
\midrule
\multirow{3}{*}{InternVL3.5-30B-A3B} & 1-step & 60.33\% & 0.7393 & 7.33 & 1.67 & 8.00 & 29.00 & 16.33 & 1.00 \\
 & 2-step & 35.00\% & 0.5533 & 8.33 & 1.00 & 23.67 & 45.67 & 28.67 & 0.67 \\
 & 3-step & 21.00\% & 0.4333 & 11.33 & 3.33 & 29.33 & 58.33 & 36.00 & 2.33 \\
\midrule
\multirow{3}{*}{InternVL3.5-38B} & 1-step & 66.33\% & 0.7880 & 6.00 & 1.67 & 4.00 & 25.33 & 13.67 & 0.00 \\
 & 2-step & 50.67\% & 0.6927 & 7.00 & 2.67 & 7.33 & 37.67 & 22.33 & 0.33 \\
 & 3-step & 32.00\% & 0.5647 & 8.33 & 2.00 & 11.00 & 53.67 & 28.67 & 3.00 \\
\midrule
\multirow{3}{*}{Qwen3-VL-235B-A22B-Instruct} & 1-step & 74.33\% & 0.8360 & 4.67 & 1.67 & 7.00 & 18.33 & 10.00 & 0.00 \\
 & 2-step & 61.00\% & 0.7653 & 6.33 & 1.00 & 5.67 & 31.67 & 14.33 & 1.00 \\
 & 3-step & 45.33\% & 0.6320 & 8.33 & 1.67 & 10.67 & 40.33 & 15.67 & 6.33 \\
\midrule
\multirow{3}{*}{Qwen3-VL-30B-A3B-Instruct} & 1-step & 68.67\% & 0.7987 & 8.00 & 3.00 & 5.67 & 23.67 & 7.00 & 0.67 \\
 & 2-step & 47.33\% & 0.6760 & 7.33 & 2.33 & 9.67 & 38.33 & 21.67 & 0.67 \\
 & 3-step & 32.67\% & 0.5507 & 8.67 & 4.33 & 14.33 & 53.33 & 28.67 & 2.67 \\
\midrule
\multirow{3}{*}{Qwen3-VL-32B-Instruct} & 1-step & 74.33\% & 0.8353 & 3.33 & 1.33 & 6.67 & 16.67 & 10.33 & 0.00 \\
 & 2-step & 60.67\% & 0.7627 & 6.00 & 1.00 & 6.67 & 29.00 & 14.33 & 0.33 \\
 & 3-step & 44.67\% & 0.6373 & 7.33 & 1.33 & 12.00 & 41.33 & 20.67 & 2.00 \\
\bottomrule
\end{tabular}
}
\caption{Detailed step-wise planning results across 1-step, 2-step, and 3-step prediction horizons. CR: Correctness Rate.}
\label{tab:stepwise_detailed_steps}
\end{table*}

\begin{table*}[htbp]
\centering
\resizebox{\textwidth}{!}{
\begin{tabular}{lccccccccc}
\toprule
Model & Mode & CR & Score & E1(\%) & E2(\%) & E3(\%) & E4(\%) & E5(\%) & E6(\%) \\
\midrule
\multirow{3}{*}{Claude Sonnet 4.5} & Early & 79.67\% & 0.8407 & 9.00 & 0.67 & 2.67 & 16.33 & 10.00 & 1.00 \\
 & Middle & 87.00\% & 0.9180 & 2.33 & 1.00 & 1.67 & 9.67 & 4.33 & 0.00 \\
 & Late & 80.67\% & 0.8627 & 6.33 & 1.00 & 2.00 & 16.00 & 5.00 & 1.00 \\
\midrule
\multirow{3}{*}{GPT-4o} & Early & 77.00\% & 0.8247 & 11.00 & 0.67 & 4.67 & 17.00 & 9.67 & 0.33 \\
 & Middle & 73.67\% & 0.8287 & 4.33 & 3.67 & 5.67 & 17.67 & 11.00 & 0.33 \\
 & Late & 67.67\% & 0.7787 & 6.67 & 8.00 & 4.67 & 19.00 & 9.00 & 1.33 \\
\midrule
\multirow{3}{*}{GPT-5} & Early & 77.67\% & 0.8187 & 9.67 & 0.33 & 2.00 & 19.00 & 10.33 & 1.00 \\
 & Middle & 87.33\% & 0.9113 & 4.33 & 0.67 & 4.33 & 9.33 & 3.33 & 0.00 \\
 & Late & 81.67\% & 0.8707 & 5.00 & 0.33 & 3.00 & 14.67 & 5.67 & 0.33 \\
\midrule
\multirow{3}{*}{Gemini 2.5 Flash} & Early & 80.67\% & 0.8667 & 4.33 & 4.67 & 8.00 & 9.67 & 5.67 & 0.00 \\
 & Middle & 77.67\% & 0.8540 & 5.33 & 4.67 & 4.67 & 15.00 & 8.00 & 1.00 \\
 & Late & 76.33\% & 0.8273 & 5.00 & 8.67 & 5.33 & 11.33 & 6.33 & 4.00 \\
\midrule
\multirow{3}{*}{Gemini 2.5 Pro} & Early & 85.00\% & 0.8907 & 5.67 & 0.33 & 2.00 & 12.67 & 5.67 & 0.33 \\
 & Middle & 82.67\% & 0.8807 & 5.00 & 2.67 & 3.00 & 11.67 & 6.00 & 0.67 \\
 & Late & 76.67\% & 0.8500 & 2.33 & 3.67 & 2.67 & 12.33 & 10.33 & 2.67 \\
\midrule
\multirow{3}{*}{Gemini 3 Pro} & Early & 80.00\% & 0.8587 & 7.33 & 0.33 & 4.00 & 16.00 & 7.33 & 0.33 \\
 & Middle & 86.33\% & 0.9113 & 4.00 & 1.00 & 2.33 & 9.33 & 5.33 & 0.00 \\
 & Late & 82.00\% & 0.8760 & 6.00 & 1.00 & 2.33 & 12.67 & 7.00 & 0.67 \\
\midrule
\multirow{3}{*}{InternVL3.5-241B-A28B} & Early & 70.67\% & 0.7880 & 9.00 & 1.00 & 4.67 & 24.33 & 14.67 & 0.00 \\
 & Middle & 75.00\% & 0.8280 & 6.00 & 1.33 & 4.67 & 18.33 & 9.33 & 0.33 \\
 & Late & 71.67\% & 0.7880 & 9.00 & 4.33 & 4.67 & 18.33 & 8.67 & 1.33 \\
\midrule
\multirow{3}{*}{InternVL3.5-30B-A3B} & Early & 62.33\% & 0.7287 & 8.33 & 0.00 & 6.67 & 28.67 & 18.33 & 0.67 \\
 & Middle & 60.33\% & 0.7393 & 7.33 & 1.67 & 8.00 & 29.00 & 16.33 & 1.00 \\
 & Late & 52.33\% & 0.6593 & 8.67 & 8.33 & 12.67 & 31.00 & 18.33 & 1.67 \\
\midrule
\multirow{3}{*}{InternVL3.5-38B} & Early & 68.67\% & 0.7773 & 8.33 & 0.33 & 3.00 & 23.33 & 17.33 & 1.33 \\
 & Middle & 66.33\% & 0.7880 & 6.00 & 1.67 & 4.00 & 25.33 & 13.67 & 0.00 \\
 & Late & 64.33\% & 0.7447 & 8.00 & 8.00 & 8.33 & 20.67 & 13.67 & 1.33 \\
\midrule
\multirow{3}{*}{Qwen3-VL-235B-A22B-Instruct} & Early & 76.33\% & 0.8420 & 4.67 & 0.33 & 5.33 & 17.00 & 9.33 & 0.00 \\
 & Middle & 74.33\% & 0.8360 & 4.67 & 1.67 & 7.00 & 18.33 & 10.00 & 0.00 \\
 & Late & 71.67\% & 0.8100 & 5.33 & 6.33 & 4.67 & 17.33 & 8.67 & 1.00 \\
\midrule
\multirow{3}{*}{Qwen3-VL-30B-A3B-Instruct} & Early & 68.00\% & 0.7800 & 6.00 & 0.67 & 5.00 & 24.67 & 14.33 & 0.67 \\
 & Middle & 68.67\% & 0.7987 & 8.00 & 3.00 & 5.67 & 23.67 & 7.00 & 0.67 \\
 & Late & 55.67\% & 0.7027 & 6.67 & 12.00 & 8.00 & 24.00 & 16.00 & 2.00 \\
\midrule
\multirow{3}{*}{Qwen3-VL-32B-Instruct} & Early & 75.33\% & 0.8253 & 7.00 & 0.33 & 4.67 & 20.00 & 9.00 & 0.00 \\
 & Middle & 74.33\% & 0.8353 & 3.33 & 1.33 & 6.67 & 16.67 & 10.33 & 0.00 \\
 & Late & 70.33\% & 0.7787 & 5.67 & 6.00 & 6.33 & 20.00 & 10.33 & 2.67 \\
\bottomrule
\end{tabular}
}
\caption{Detailed step-wise planning results across early, middle, and late trajectory stages. CR: Correctness Rate.}
\label{tab:stepwise_stages}
\end{table*}

\begin{table*}[htbp]
\centering
\renewcommand{\arraystretch}{0.75}
\resizebox{\textwidth}{!}{
\begin{tabular}{lcccccccccc}
\toprule
Model & Level & ExtUse & CR (\%) & Avg Grade & E1 (\%) & E2 (\%) & E3 (\%) & E4 (\%) & E5 (\%) & E6 (\%) \\
\midrule
\multirow{5}{*}{Claude Sonnet 4.5} & 2 & 1 & 87.33 & 0.91 &   4.67   &   1.33   &   2.00   &   11.33   &   3.33   &   0.00 \\
 & 4 & 1 & 88.00 & 0.91 &   3.33   &   1.33   &   0.67   &   8.00   &   6.00   &   0.67 \\
 & 6 & 2 & 88.00 & 0.92 &   3.33   &   1.33   &   1.33   &   8.00   &   4.67   &   0.00 \\
 & 8 & 1 & 84.67 & 0.89 &   4.67   &   2.67   &   4.00   &   9.33   &   5.33   &   0.00 \\
 & 10 & 1 & 80.67 & 0.88 &   4.00   &   2.67   &   4.67   &   13.33   &   4.67   &   0.67 \\
\midrule
\multirow{5}{*}{GPT-4o} & 2 & 3 & 70.00 & 0.80 &   6.67   &   2.00   &   5.33   &   21.33   &   12.67   &   0.00 \\
 & 4 & 1 & 74.67 & 0.84 &   6.67   &   2.67   &   4.67   &   16.00   &   10.67   &   0.00 \\
 & 6 & 3 & 69.33 & 0.79 &   8.00   &   3.33   &   9.33   &   16.67   &   10.00   &   0.67 \\
 & 8 & 4 & 66.67 & 0.79 &   5.33   &   4.00   &   4.00   &   19.33   &   16.00   &   0.67 \\
 & 10 & 4 & 67.33 & 0.78 &   7.33   &   4.67   &   8.00   &   20.67   &   11.33   &   0.00 \\
\midrule
\multirow{5}{*}{GPT-5} & 2 & 3 & 82.00 & 0.86 &   9.33   &   0.67   &   5.33   &   12.67   &   4.00   &   0.00 \\
 & 4 & 2 & 76.00 & 0.84 &   6.67   &   1.33   &   3.33   &   15.33   &   6.67   &   0.00 \\
 & 6 & 6 & 80.00 & 0.85 &   9.33   &   0.00   &   1.33   &   13.33   &   7.33   &   0.00 \\
 & 8 & 6 & 73.33 & 0.81 &   8.67   &   1.33   &   4.67   &   15.33   &   10.00   &   0.00 \\
 & 10 & 6 & 82.00 & 0.86 &   6.00   &   1.33   &   3.33   &   9.33   &   8.67   &   0.00 \\
\midrule
\multirow{5}{*}{Gemini 2.5 Flash} & 2 & 2 & 80.67 & 0.88 &   4.00   &   4.00   &   5.33   &   9.33   &   6.00   &   0.00 \\
 & 4 & 0 & 80.67 & 0.86 &   4.00   &   7.33   &   6.67   &   12.67   &   2.67   &   0.67 \\
 & 6 & 3 & 80.00 & 0.87 &   5.33   &   3.33   &   2.67   &   15.33   &   6.00   &   0.67 \\
 & 8 & 4 & 76.00 & 0.84 &   5.33   &   4.00   &   6.00   &   13.33   &   8.67   &   0.67 \\
 & 10 & 4 & 79.33 & 0.85 &   4.67   &   3.33   &   6.00   &   11.33   &   7.33   &   0.67 \\
\midrule
\multirow{5}{*}{Gemini 2.5 Pro} & 2 & 1 & 78.00 & 0.85 &   6.00   &   2.67   &   4.67   &   13.33   &   8.00   &   0.00 \\
 & 4 & 1 & 82.00 & 0.88 &   6.67   &   2.00   &   2.67   &   11.33   &   8.00   &   1.33 \\
 & 6 & 2 & 78.67 & 0.86 &   6.00   &   2.00   &   3.33   &   14.67   &   8.00   &   0.67 \\
 & 8 & 2 & 78.67 & 0.85 &   4.67   &   2.67   &   6.67   &   11.33   &   7.33   &   0.67 \\
 & 10 & 1 & 77.33 & 0.84 &   4.67   &   6.00   &   8.00   &   12.67   &   6.00   &   1.33 \\
\midrule
\multirow{5}{*}{Gemini 3 Pro} & 2 & 2 & 84.67 & 0.90 &   4.00   &   0.67   &   2.00   &   10.67   &   7.33   &   0.00 \\
 & 4 & 4 & 83.33 & 0.89 &   6.00   &   1.33   &   0.67   &   10.00   &   7.33   &   0.00 \\
 & 6 & 1 & 82.67 & 0.89 &   7.33   &   0.67   &   1.33   &   10.00   &   7.33   &   0.00 \\
 & 8 & 2 & 82.00 & 0.87 &   4.67   &   2.00   &   4.00   &   11.33   &   8.00   &   0.00 \\
 & 10 & 3 & 82.00 & 0.88 &   3.33   &   0.67   &   2.00   &   10.67   &   8.67   &   0.00 \\
\midrule
\multirow{5}{*}{InternVL3.5-241B-A28B} & 2 & 2 & 77.33 & 0.85 &   4.67   &   1.33   &   1.33   &   14.67   &   8.00   &   0.00 \\
 & 4 & 4 & 75.33 & 0.84 &   4.00   &   2.00   &   2.67   &   18.67   &   11.33   &   0.00 \\
 & 6 & 4 & 72.67 & 0.82 &   5.33   &   2.00   &   2.67   &   19.33   &   12.67   &   0.00 \\
 & 8 & 3 & 70.67 & 0.83 &   4.00   &   2.67   &   4.00   &   21.33   &   10.00   &   0.00 \\
 & 10 & 2 & 74.00 & 0.84 &   4.00   &   0.67   &   3.33   &   18.67   &   12.67   &   0.00 \\
\midrule
\multirow{5}{*}{InternVL3.5-30B-A3B} & 2 & 0 & 58.67 & 0.74 &   6.67   &   1.33   &   8.67   &   32.00   &   18.67   &   0.00 \\
 & 4 & 2 & 55.33 & 0.71 &   10.00   &   4.00   &   7.33   &   33.33   &   20.67   &   0.00 \\
 & 6 & 0 & 59.33 & 0.74 &   7.33   &   3.33   &   8.00   &   28.00   &   20.00   &   0.67 \\
 & 8 & 2 & 57.33 & 0.73 &   7.33   &   4.67   &   8.67   &   30.00   &   20.00   &   0.67 \\
 & 10 & 2 & 58.67 & 0.72 &   8.00   &   2.67   &   9.33   &   27.33   &   16.00   &   0.67 \\
\midrule
\multirow{5}{*}{InternVL3.5-38B} & 2 & 1 & 67.33 & 0.80 &   5.33   &   2.67   &   4.00   &   25.33   &   13.33   &   0.00 \\
 & 4 & 1 & 66.00 & 0.79 &   6.00   &   4.00   &   4.67   &   22.67   &   12.00   &   0.00 \\
 & 6 & 2 & 64.67 & 0.79 &   4.67   &   4.67   &   5.33   &   24.67   &   13.33   &   0.00 \\
 & 8 & 1 & 66.00 & 0.80 &   5.33   &   2.00   &   3.33   &   25.33   &   12.00   &   0.00 \\
 & 10 & 4 & 62.67 & 0.78 &   5.33   &   2.00   &   4.67   &   27.33   &   15.33   &   1.33 \\
\midrule
\multirow{5}{*}{Qwen3-VL-235B-A22B-Instruct} & 2 & 1 & 74.67 & 0.83 &   4.67   &   3.33   &   4.00   &   21.33   &   6.67   &   0.67 \\
 & 4 & 1 & 78.00 & 0.87 &   4.00   &   2.67   &   3.33   &   14.00   &   6.00   &   0.00 \\
 & 6 & 1 & 76.00 & 0.85 &   4.00   &   1.33   &   4.67   &   16.67   &   5.33   &   0.67 \\
 & 8 & 1 & 74.67 & 0.85 &   4.67   &   3.33   &   2.67   &   17.33   &   8.00   &   0.00 \\
 & 10 & 1 & 74.00 & 0.83 &   8.67   &   3.33   &   4.00   &   18.00   &   9.33   &   0.67 \\
\midrule
\multirow{5}{*}{Qwen3-VL-30B-A3B-Instruct} & 2 & 0 & 67.33 & 0.79 &   6.67   &   2.67   &   6.00   &   23.33   &   14.67   &   0.00 \\
 & 4 & 1 & 70.00 & 0.80 &   6.67   &   4.00   &   2.67   &   23.33   &   12.00   &   0.00 \\
 & 6 & 2 & 68.00 & 0.80 &   7.33   &   5.33   &   4.00   &   20.00   &   11.33   &   0.00 \\
 & 8 & 4 & 66.67 & 0.79 &   6.00   &   3.33   &   6.00   &   20.67   &   14.67   &   0.00 \\
 & 10 & 6 & 66.67 & 0.78 &   10.00   &   6.67   &   2.67   &   21.33   &   15.33   &   1.33 \\
\midrule
\multirow{5}{*}{Qwen3-VL-32B-Instruct} & 2 & 1 & 75.33 & 0.86 &   4.00   &   1.33   &   2.67   &   16.67   &   8.00   &   0.00 \\
 & 4 & 5 & 71.33 & 0.81 &   5.33   &   3.33   &   4.00   &   18.00   &   11.33   &   0.00 \\
 & 6 & 5 & 73.33 & 0.83 &   5.33   &   2.00   &   4.00   &   17.33   &   10.67   &   0.00 \\
 & 8 & 4 & 70.00 & 0.80 &   4.67   &   1.33   &   6.67   &   22.67   &   12.00   &   0.00 \\
 & 10 & 6 & 72.67 & 0.81 &   6.00   &   2.67   &   4.00   &   20.00   &   12.00   &   0.00 \\
\bottomrule
\end{tabular}
}
\caption{Detailed step-wise planning results under tool-extraneous with varying distractor counts. CR: Correctness Rate; ExtUse: Number of extraneous tools used.}
\label{tab:stepwise_extraneous_detailed}
\end{table*}

\begin{table*}[htbp]
\centering
\renewcommand{\arraystretch}{0.75}
\resizebox{\textwidth}{!}{
\begin{tabular}{lcccccccccc}
\toprule
Model & Level & ExtUse & CR (\%) & Avg Grade & E1 (\%) & E2 (\%) & E3 (\%) & E4 (\%) & E5 (\%) & E6 (\%) \\
\midrule
\multirow{5}{*}{Claude Sonnet 4.5} & 2 & 10 & 70.47 & 0.87 & 0.00 & 4.03 & 17.45 & 6.04 & 8.05 & 4.03 \\
 & 4 & 10 & 58.00 & 0.80 & 2.00 & 8.67 & 25.33 & 5.33 & 10.00 & 5.33 \\
 & 6 & 10 & 74.50 & 0.89 & 0.00 & 0.00 & 16.78 & 4.70 & 8.05 & 4.70 \\
 & 8 & 9 & 61.74 & 0.81 & 2.01 & 9.40 & 17.45 & 6.71 & 8.05 & 5.37 \\
 & 10 & 11 & 70.47 & 0.86 & 0.00 & 2.01 & 17.45 & 4.70 & 8.72 & 6.04 \\
\midrule
\multirow{5}{*}{GPT-4o} & 2 & 13 & 25.33 & 0.61 & 4.00 & 33.87 & 31.33 & 33.33 & 20.00 & 10.67 \\
 & 4 & 14 & 19.33 & 0.59 & 2.67 & 36.00 & 31.33 & 30.00 & 27.33 & 8.00 \\
 & 6 & 15 & 21.48 & 0.61 & 2.01 & 31.54 & 39.60 & 35.57 & 24.16 & 8.05 \\
 & 8 & 18 & 21.48 & 0.60 & 2.68 & 34.90 & 34.23 & 27.52 & 26.85 & 11.41 \\
 & 10 & 17 & 19.46 & 0.59 & 4.03 & 34.90 & 34.90 & 28.19 & 24.83 & 6.71 \\
\midrule
\multirow{5}{*}{GPT-5} & 2 & 19 & 61.74 & 0.82 & 0.00 & 8.72 & 12.08 & 2.01 & 24.16 & 2.01 \\
 & 4 & 13 & 58.67 & 0.78 & 2.00 & 9.33 & 14.00 & 4.00 & 15.33 & 6.00 \\
 & 6 & 26 & 53.69 & 0.78 & 0.67 & 5.37 & 14.77 & 5.37 & 27.52 & 5.37 \\
 & 8 & 25 & 53.02 & 0.76 & 2.01 & 10.07 & 12.75 & 3.36 & 24.83 & 2.68 \\
 & 10 & 32 & 49.66 & 0.72 & 2.68 & 11.41 & 8.72 & 4.70 & 29.53 & 2.68 \\
\midrule
\multirow{5}{*}{Gemini 2.5 Flash} & 2 & 2 & 42.95 & 0.73 & 1.34 & 30.20 & 22.82 & 11.41 & 6.71 & 6.04 \\
 & 4 & 4 & 37.33 & 0.69 & 4.00 & 31.33 & 23.33 & 12.67 & 6.67 & 9.33 \\
 & 6 & 4 & 41.61 & 0.73 & 1.34 & 25.50 & 26.85 & 12.75 & 9.40 & 8.05 \\
 & 8 & 9 & 40.27 & 0.70 & 1.34 & 25.50 & 23.49 & 16.78 & 10.07 & 10.74 \\
 & 10 & 8 & 38.93 & 0.68 & 3.36 & 29.53 & 21.48 & 15.44 & 9.40 & 6.71 \\
\midrule
\multirow{5}{*}{Gemini 2.5 Pro} & 2 & 8 & 63.09 & 0.82 & 0.67 & 9.40 & 19.46 & 10.07 & 8.05 & 6.04 \\
 & 4 & 11 & 49.33 & 0.74 & 3.33 & 16.00 & 24.67 & 8.00 & 12.67 & 8.67 \\
 & 6 & 12 & 51.68 & 0.77 & 2.01 & 10.07 & 22.82 & 10.07 & 15.44 & 8.05 \\
 & 8 & 15 & 59.06 & 0.79 & 0.67 & 9.40 & 19.46 & 7.38 & 15.44 & 6.71 \\
 & 10 & 14 & 50.34 & 0.74 & 2.68 & 12.75 & 25.50 & 10.07 & 11.41 & 6.71 \\
\midrule
\multirow{5}{*}{Gemini 3 Pro} & 2 & 6 & 74.50 & 0.88 & 0.67 & 5.37 & 14.09 & 4.70 & 6.04 & 4.03 \\
 & 4 & 6 & 73.15 & 0.86 & 2.01 & 7.38 & 12.75 & 5.37 & 7.38 & 5.37 \\
 & 6 & 11 & 67.79 & 0.86 & 0.00 & 4.03 & 16.11 & 5.37 & 12.75 & 2.68 \\
 & 8 & 11 & 67.11 & 0.83 & 2.01 & 10.07 & 11.41 & 4.03 & 12.08 & 0.67 \\
 & 10 & 12 & 67.11 & 0.83 & 2.01 & 10.07 & 15.44 & 3.36 & 10.74 & 2.01 \\
\midrule
\multirow{5}{*}{InternVL3.5-241B-A28B} & 2 & 12 & 24.32 & 0.65 & 2.03 & 24.32 & 37.16 & 35.81 & 20.27 & 11.49 \\
 & 4 & 14 & 25.00 & 0.64 & 2.70 & 24.32 & 37.16 & 35.14 & 17.57 & 12.84 \\
 & 6 & 17 & 25.68 & 0.63 & 2.03 & 25.68 & 36.49 & 39.19 & 19.59 & 14.86 \\
 & 8 & 16 & 19.59 & 0.58 & 2.03 & 27.70 & 39.86 & 35.81 & 19.59 & 11.49 \\
 & 10 & 21 & 22.30 & 0.59 & 4.05 & 29.05 & 32.43 & 34.46 & 22.97 & 10.81 \\
\midrule
\multirow{5}{*}{InternVL3.5-30B-A3B} & 2 & 8 & 9.33 & 0.50 & 8.00 & 49.33 & 40.67 & 43.33 & 25.33 & 18.00 \\
 & 4 & 14 & 9.33 & 0.50 & 6.00 & 49.33 & 36.00 & 50.67 & 27.33 & 20.00 \\
 & 6 & 18 & 8.67 & 0.47 & 6.67 & 46.00 & 40.00 & 41.33 & 33.33 & 16.67 \\
 & 8 & 17 & 7.33 & 0.44 & 8.00 & 53.33 & 37.33 & 42.67 & 33.33 & 14.67 \\
 & 10 & 21 & 10.00 & 0.47 & 10.00 & 48.67 & 38.67 & 42.67 & 30.00 & 18.00 \\
\midrule
\multirow{5}{*}{InternVL3.5-38B} & 2 & 17 & 22.00 & 0.62 & 2.00 & 27.33 & 34.00 & 33.33 & 22.67 & 18.67 \\
 & 4 & 18 & 25.33 & 0.64 & 2.00 & 26.00 & 34.67 & 32.67 & 25.33 & 14.00 \\
 & 6 & 21 & 28.00 & 0.64 & 2.67 & 22.00 & 35.33 & 35.33 & 28.00 & 17.33 \\
 & 8 & 21 & 27.33 & 0.62 & 3.33 & 22.00 & 34.00 & 30.67 & 26.67 & 14.00 \\
 & 10 & 21 & 23.33 & 0.59 & 3.33 & 26.00 & 34.67 & 33.33 & 26.00 & 12.67 \\
\midrule
\multirow{5}{*}{Qwen3-VL-235B-A22B-Instruct} & 2 & 11 & 31.54 & 0.66 & 1.34 & 18.79 & 35.57 & 28.86 & 18.12 & 21.48 \\
 & 4 & 14 & 29.33 & 0.63 & 4.00 & 17.33 & 35.33 & 22.67 & 14.67 & 22.00 \\
 & 6 & 14 & 30.20 & 0.66 & 1.34 & 15.44 & 34.23 & 24.16 & 18.12 & 24.16 \\
 & 8 & 16 & 29.53 & 0.64 & 2.68 & 11.41 & 36.24 & 25.50 & 19.46 & 22.82 \\
 & 10 & 18 & 27.70 & 0.63 & 2.70 & 14.86 & 35.81 & 27.03 & 18.92 & 22.30 \\
\midrule
\multirow{5}{*}{Qwen3-VL-30B-A3B-Instruct} & 2 & 16 & 15.33 & 0.58 & 1.33 & 28.00 & 42.00 & 30.67 & 20.67 & 33.33 \\
 & 4 & 17 & 13.33 & 0.56 & 0.67 & 25.33 & 44.00 & 31.33 & 24.00 & 33.33 \\
 & 6 & 20 & 12.00 & 0.55 & 1.33 & 26.53 & 44.00 & 34.00 & 26.00 & 30.00 \\
 & 8 & 21 & 11.33 & 0.52 & 3.33 & 32.00 & 43.33 & 26.67 & 25.33 & 31.33 \\
 & 10 & 23 & 11.33 & 0.52 & 3.33 & 35.33 & 42.00 & 29.33 & 24.67 & 30.00 \\
\midrule
\multirow{5}{*}{Qwen3-VL-32B-Instruct} & 2 & 18 & 34.00 & 0.69 & 1.33 & 6.00 & 39.33 & 16.67 & 18.67 & 18.67 \\
 & 4 & 21 & 36.67 & 0.67 & 2.67 & 9.33 & 34.67 & 16.00 & 26.00 & 16.67 \\
 & 6 & 21 & 28.00 & 0.64 & 1.33 & 16.67 & 37.33 & 14.67 & 25.33 & 16.00 \\
 & 8 & 24 & 25.33 & 0.59 & 3.33 & 16.67 & 39.33 & 17.33 & 27.33 & 22.00 \\
 & 10 & 24 & 32.00 & 0.63 & 2.00 & 14.00 & 34.67 & 16.00 & 24.67 & 18.67 \\
\bottomrule
\end{tabular}
}
\caption{Detailed holistic planning results under tool-extraneous with varying distractor counts. CR: Correctness Rate; ExtUse: Number of extraneous tools used.}
\label{tab:holistic_extraneous_detailed}
\end{table*}

\begin{table*}[htbp]
\centering
\resizebox{\textwidth}{!}{
\begin{tabular}{l | ccc ccc | ccc ccc}
\toprule
& \multicolumn{6}{c|}{Holistic Planning} & \multicolumn{6}{c}{Step-wise Planning} \\
& \multicolumn{3}{c}{CR (\%)} & \multicolumn{3}{c|}{Score} & \multicolumn{3}{c}{CR (\%)} & \multicolumn{3}{c}{Score} \\
Model & Orig & Red & $\Delta$ & Orig & Red & $\Delta$ & Orig & Red & $\Delta$ & Orig & Red & $\Delta$ \\
\midrule
Claude Sonnet 4.5 & 79.33 & 67.02 & -12.31 & 0.92 & 0.85 & -0.08 & 86.67 & 85.73 & -0.93 & 0.92 & 0.90 & -0.02 \\
Gemini 2.5 Flash & 53.33 & 40.21 & -13.12 & 0.79 & 0.71 & -0.09 & 78.00 & 79.33 & +1.33 & 0.85 & 0.86 & +0.01 \\
Gemini 2.5 Pro & 61.33 & 54.69 & -6.64 & 0.83 & 0.77 & -0.06 & 80.00 & 78.93 & -1.07 & 0.86 & 0.86 & -0.01 \\
Gemini 3 Pro & 77.18 & 69.93 & -7.25 & 0.90 & 0.85 & -0.05 & 80.00 & 82.93 & +2.93 & 0.87 & 0.88 & +0.02 \\
GPT-4o & 26.67 & 21.42 & -5.25 & 0.67 & 0.60 & -0.07 & 73.33 & 69.60 & -3.73 & 0.83 & 0.80 & -0.03 \\
GPT-5 & 77.33 & 55.36 & -21.97 & 0.92 & 0.77 & -0.15 & 85.33 & 78.67 & -6.67 & 0.90 & 0.85 & -0.05 \\
InternVL3.5-241B-A28B & 22.30 & 23.38 & +1.08 & 0.65 & 0.62 & -0.03 & 77.33 & 74.00 & -3.33 & 0.84 & 0.84 & -0.00 \\
InternVL3.5-30B-A3B & 12.00 & 8.93 & -3.07 & 0.51 & 0.48 & -0.03 & 60.00 & 57.87 & -2.13 & 0.75 & 0.73 & -0.03 \\
InternVL3.5-38B & 24.00 & 25.20 & +1.20 & 0.67 & 0.62 & -0.05 & 67.33 & 65.33 & -2.00 & 0.81 & 0.79 & -0.01 \\
Qwen3-VL-235B-A22B-Instruct & 30.67 & 29.66 & -1.00 & 0.69 & 0.64 & -0.05 & 74.00 & 75.47 & +1.47 & 0.84 & 0.84 & +0.01 \\
Qwen3-VL-30B-A3B-Instruct & 16.00 & 12.67 & -3.33 & 0.60 & 0.54 & -0.06 & 64.00 & 67.73 & +3.73 & 0.76 & 0.79 & +0.03 \\
Qwen3-VL-32B-Instruct & 32.67 & 31.20 & -1.47 & 0.70 & 0.64 & -0.05 & 72.00 & 72.53 & +0.53 & 0.82 & 0.82 & +0.00 \\
\bottomrule
\end{tabular}
}
\caption{Comparative analysis of holistic and step-wise planning under average tool-extraneous conditions. CR: Correctness Rate.}
\label{tab:extraneous_average}
\end{table*}

\clearpage
\section{Model Sources and Hyperparameters}
\label{app:model_config}
Table~\ref{tab:model_config} summarizes the sources and hyperparameter configurations for all models evaluated in our experiments. For API-based models, we access them through their official endpoints with specific API model identifiers; for open-source models, we deploy local checkpoints using vLLM.

\begin{table*}[htbp]
\centering
\caption{Model sources and hyperparameter configurations used in experiments.}
\label{tab:model_config}
\renewcommand{\arraystretch}{1.1}
\resizebox{\textwidth}{!}{
\begin{tabular}{llll}
\toprule
\textbf{Model} & \textbf{Parameter Setting} & \textbf{Source} & \textbf{URL} \\
\midrule
\multicolumn{4}{l}{\textit{Proprietary Models (API Access)}} \\
\midrule
GPT-4o & temperature = 0.0 & gpt-4o & \url{https://platform.openai.com} \\
GPT-5 & default & gpt-5 & \url{https://platform.openai.com} \\
Claude Sonnet 4.5 & temperature = 0.0 & claude-sonnet-4-5-20250929 & \url{https://www.anthropic.com/} \\
Gemini 2.5 Flash & temperature = 0.0, thinking\_budget = 0 & gemini-2.5-flash & \url{https://ai.google.dev/} \\
Gemini 2.5 Pro & temperature = 0.0, thinking\_budget = default & gemini-2.5-pro & \url{https://ai.google.dev/} \\
Gemini 3 Pro & temperature = 0.0, thinking\_level = high & gemini-3-pro-preview & \url{https://ai.google.dev/} \\
\midrule
\multicolumn{4}{l}{\textit{Open-Source Models (Local Deployment via vLLM)}} \\
\midrule
InternVL3.5-38B & temperature = 0.0 & local checkpoint & \url{https://huggingface.co/OpenGVLab/InternVL3_5-38B} \\
InternVL3.5-30B-A3B & temperature = 0.0 & local checkpoint & \url{https://huggingface.co/OpenGVLab/InternVL3_5-30B-A3B} \\
InternVL3.5-241B-A28B & temperature = 0.0 & local checkpoint & \url{https://huggingface.co/OpenGVLab/InternVL3_5-241B-A28B} \\
Qwen3-VL-32B-Instruct & temperature = 0.0 & local checkpoint & \url{https://huggingface.co/Qwen/Qwen3-VL-32B-Instruct} \\
Qwen3-VL-30B-A3B-Instruct & temperature = 0.0 & local checkpoint & \url{https://huggingface.co/Qwen/Qwen3-VL-30B-A3B-Instruct} \\
Qwen3-VL-235B-A22B-Instruct & temperature = 0.0 & local checkpoint & \url{https://huggingface.co/Qwen/Qwen3-VL-235B-A22B-Instruct} \\
Qwen3.5-397B-A17B (Judge) & temperature = 0.0 & local checkpoint & \url{https://huggingface.co/Qwen/Qwen3.5-397B-A17B} \\
\bottomrule
\end{tabular}
}
\end{table*}

The open-weight judge calibration uses the same 720 instances, reference-aware evaluation prompts, output schema, and deterministic decoding protocol as the proprietary-judge comparison. Qwen3.5-397B-A17B is the recommended local diagnostic scorer; Qwen3-VL-235B-A22B-Instruct is included as a lower-fidelity open-weight comparison.

\end{document}